\documentclass{egpubl}
\usepackage{eg2022}
\usepackage{xcolor}

\STAR                   % uncomment for STAR contribution
\usepackage[T1]{fontenc}
\usepackage{dfadobe}  

\usepackage{cite}  % comment out for biblatex with backend=biber

\BibtexOrBiblatex
\electronicVersion
\PrintedOrElectronic
\ifpdf \usepackage[pdftex]{graphicx} \pdfcompresslevel=9
\else \usepackage[dvips]{graphicx} \fi

\usepackage{egweblnk} 
\usepackage{amssymb,amsmath,bm,tikz,pgfplots,subfig,rotating}
\pgfplotsset{compat=1.17}
\usetikzlibrary{calc,patterns,angles,quotes}

\usepackage{microtype}

\usepackage{multirow}
\newcommand{\tabincell}[2]{\begin{tabular}{@{}#1@{}}#2\end{tabular}}
\usepackage{makecell}

\DeclareMathOperator{\sdf}{SDF}
\DeclareMathOperator{\tsdf}{TSDF}

\newcommand{\ppl}{\text{ppl}}
\newcommand{\pos}{\text{pos}}
\newcommand{\smooth}{\text{smooth}}

\newcommand{\rigid}{\text{rigid}}

\DeclareMathOperator*{\argmin}{arg\,min}
\DeclareMathOperator*{\rot}{Rot}
\DeclareMathOperator*{\diagmat}{diag}
\DeclareMathOperator*{\areafunc}{area}
\DeclareMathOperator{\off}{off}
\DeclareMathOperator{\rank}{rank}

\newcommand{\wid}{w_i^{\text{d}}}

\newcommand{\tablefont}[1]{\fontsize{6.4}{7.58}\selectfont #1}

\newcommand{\eg}[0]{\emph{e.g.,~}} % RMD a useful command for consistency throughout document
\newcommand{\ie}[0]{\emph{i.e.,~}} % RMD a useful command for consistency throughout document
\newcommand{\st}[0]{\text{s.t.~}}

\newcommand{\headingbreak}{\\[-0.2em]}

\title[A Survey of Non-Rigid 3D Registration]%
{A Survey of Non-Rigid 3D Registration}

\author[B. Deng et al.]
{\parbox{\textwidth}{\centering Bailin Deng$^{1}$\orcid{0000-0002-0158-7670}
		\quad Yuxin Yao$^{2}$
		\quad Roberto M. Dyke$^{3}$\orcid{0000-0003-0361-422X}
		\quad Juyong Zhang$^{2}$\orcid{0000-0002-1805-1426}\thanks{Corresponding author: juyong@ustc.edu.cn (Juyong Zhang)}
	}
	\\
	{\parbox{\textwidth}{\centering $^1$Cardiff University \quad
			$^2$University of Science and Technology of China \quad
			$^3$Universit\`{a} della Svizzera italiana
		}
	}
}
\begin{document}
	
	% uncomment for using teaser
	% \teaser{
	%  \includegraphics[width=\linewidth]{eg_new}
	%  \centering
	%   \caption{New EG Logo}
	% \label{fig:teaser}
	%}
	
	\maketitle
	
	%-------------------------------------------------------------------------
	\begin{abstract}
	Non-rigid registration computes an alignment between a source surface with a target surface in a non-rigid manner. In the past decade, with the advances in 3D sensing technologies that can measure time-varying surfaces, non-rigid registration has been applied for the acquisition of deformable shapes and has a wide range of applications. This survey presents a comprehensive review of non-rigid registration methods for 3D shapes, focusing on techniques related to dynamic shape acquisition and reconstruction. In particular, we review different approaches for representing the deformation field, and the methods for computing the desired deformation. Both optimization-based and learning-based methods are covered. We also review benchmarks and datasets for evaluating non-rigid registration methods, and discuss potential future research directions.
	

	
	\end{abstract}  
	%-------------------------------------------------------------------------
	\section{Introduction}
\begin{figure}[t]
	\centering
	\includegraphics[width=\columnwidth]{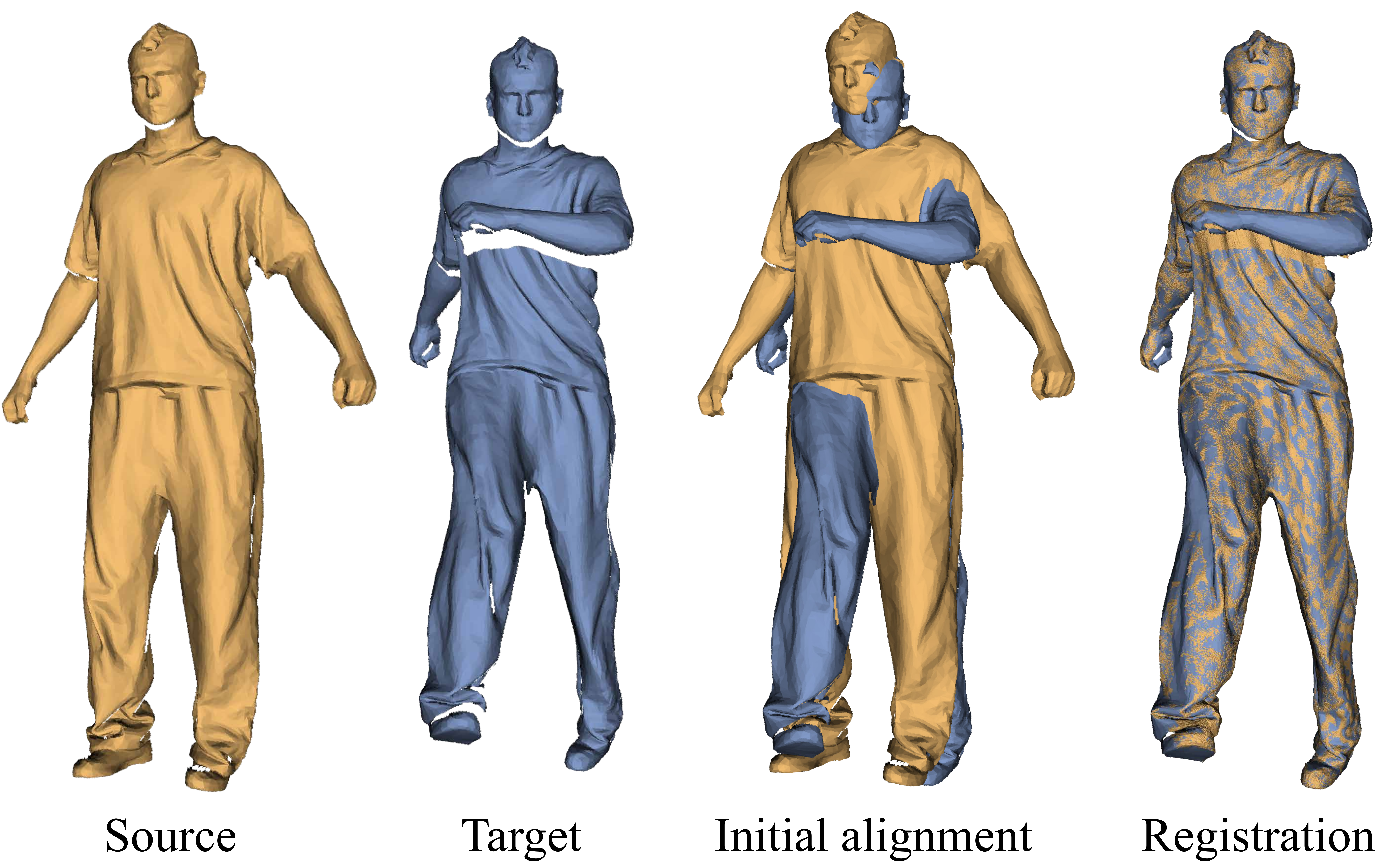}
	\caption{An example of non-rigid registration, where the source surface is deformed in a non-rigid manner to align with the target.}
	\label{fig:NRR}
	\end{figure}

	Surface registration computes a deformation that aligns a source surface with a target surface (see Fig.~\ref{fig:NRR} for an example). It has many applications in computer graphics and computer vision. One common application is 3D surface acquisition, where an object is scanned from different directions and the scans are aligned via registration to derive the 3D measurements in a common coordinate system. Depending on the application, the surfaces can be aligned by a rigid or non-rigid deformation. In the former case, the whole source surface undergoes a single rotation and translation, which is suitable for the registration of static shapes. In the latter case, different parts of the source surface can undergo different deformations, to account for the non-rigid behavior---such as articulation of the underlying shape. Computing such deformations is a fundamental problem for the acquisition and analysis of non-rigidly deformable objects. In the past decade, with the availability of consumer depth sensors that can measure time-varying surfaces, non-rigid registration has been applied to dynamic shape reconstruction problems such as human performance capture, enabling a wide range of applications in VR, AR, and entertainment.
	
	Non-rigid registration is a challenging problem. First, unlike rigid registration which only involves a single rotation and translation, non-rigid registration often needs to determine a deformation field for the source surface. A proper representation of the deformation field needs to be chosen to ensure sufficient expressiveness for an accurate alignment without incurring excessive computational costs. In addition, 3D measure data often contain noise and outliers, and the two surfaces to be aligned can have a low ratio of overlap when they cover different parts of an object. These can make it difficult to derive a reliable correspondence between the two surfaces for their alignment. Moreover, in many practical applications, the registration needs to be performed efficiently, preferably in real-time. This survey provides an overview of significant work that addresses these challenges.

	\subsection{Scope and Related Works}

	This survey considers non-rigid registration methods for surfaces in 3D spaces, focusing on techniques related to dynamic shape acquisition and reconstruction in computer graphics and vision. As such, some shape correspondence methods intended for other applications are not included in this survey. For example, we do not include shape correspondence methods that require the source and target surfaces to have a bijective correspondence, as the inputs to a 3D acquisition system are often partial scans rather than complete shapes.
	We do include methods that compute correspondence between a partial shape and a complete shape, since such scenarios can occur in a reconstruction system where a template mesh is updated according to input partial scans~\cite{guo2015robust,li2017robust,yu2018doublefusion,su2020robustfusion}.
	Our discussion is focused on methods that perform registration using geometry information. Multimodal methods that utilize geometry together with other types of information have been recently reviewed in~\cite{Saiti2020survey} and are outside the scope of this survey.
	
	Methods for 3D registration and correspondence have been reviewed in some existing surveys. Among them, \cite{vankaick2011survey} and~\cite{sahillioglu2020recent} focus on shape correspondence rather than registration. \cite{Tam2013survey} consider both rigid and non-rigid registration, but it was published almost a decade ago and did not include more recent development in non-rigid registration, such as the use of robust norms to handle outliers and partial overlaps, efficient numerical solvers for non-smooth formulations, and deep learning-based methods. A more recent review of non-rigid registration methods and software solutions is presented in~\cite{keszei2017survey}, but it focused on applications in medical imaging only. Our survey provides an up-to-date overview of the literature in graphics and vision.
	
	\subsection{Organization}
	We first review different representations of deformation fields in Section~\ref{sec:representation}. Afterwards, methods for determining the deformation/correspondence are reviewed in detail. We divide such methods into two categories: \emph{extrinsic methods} attempt to directly reduce the distance between the source and target surfaces measured in the ambient 3D space, while \emph{intrinsic methods} utilize intrinsic metrics on the surfaces to compute the alignment.  These two types of methods are reviewed in Section~\ref{sec:extrinsic} and Section~\ref{sec:intrinsic} respectively, each covering both optimization-based and learning-based approaches. Section~\ref{sec:benchmarks} provides an overview of datasets and benchmarks for evaluating non-rigid registration methods. Finally, we discuss research challenges and future directions in Section~\ref{sec:future}.
	
	\subsection{Notations}
	In the following discussion, we use $\mathcal{X}$ and $\mathcal{Y}$ to denote the source surface and the target surface in the 3D space, respectively. The surfaces can be represented either explicitly as point clouds or meshes, or implicitly as level sets of signed distance functions.  For explicit representations, we use $\mathbf{X}=\{\mathbf{x}_1, \ldots, \mathbf{x}_m\}$ and $\mathbf{Y}=\{\mathbf{y}_1, \ldots, \mathbf{y}_n\}$ to denote the point sets on the source surface and the target surface respectively, and use $\mathbf{\widehat{X}}=\{\widehat{\mathbf{x}}_1, \ldots,\widehat{\mathbf{x}}_m\}$ to denote the positions of source surface points after the deformation. 
	
	\section{Representation of the Deformation Field}
	\label{sec:representation}
	Many methods compute a deformation field that transforms the source surface to align it with the target surface.
	The representation of the deformation field often involves a trade-off between its expressiveness and the computational cost. On one hand, a representation with higher degrees of freedom tends to be more expressive and may achieve better alignment. On the other hand, higher degrees of freedom will require a larger number of variables to be determined, which can increase the computational cost. In this section, we review different representations that have been used in the literature. Since our aim  is to categorize different representations, for each type of representation we only cite some representative papers as examples instead of listing all methods that use the representation.

    \paragraph*{Pointwise position variables} 
    One of the simplest ways to define a deformation field is to directly regard the point positions $\widehat{\mathbf{x}}$ on the new shape as the variables and compute them via optimization~\cite{liao2009modeling, huang2011global, yamazaki2013non}. 
    In practice, due to the physical behavior modeled by the deformation, a vertex can rarely move independently of other vertices, and such pointwise position variables can lead to redundant degrees of freedom. Therefore, the variables are often subject to additional regularization such as local shape preservation~\cite{liao2009modeling,huang2011global} and local similarity~\cite{yamazaki2013non}.
	
	\paragraph*{Pointwise affine transformation} Instead of treating the point positions as variables, some methods define an affine transformation $(\mathbf{A}_i, \mathbf{t}_i)$ on each point of the source surface~\cite{allen2003space}:
	\begin{equation}
	\label{Eq:pointwiseAffine}
		\widehat{\mathbf{x}}_i = \mathbf{A}_i\mathbf{x}_i+\mathbf{t}_i,
	\end{equation}
	where $\mathbf{A}_i\in\mathbb{R}^{3\times 3}$ and $\mathbf{t}_i\in\mathbb{R}^3$. 
	Compared with pointwise position variables, this can better model more complex deformations such as local rotations.
	Based on the observation that the deformations in many non-rigid registration problems are locally rigid, affine transformations are often computed with a constraint that they are close to rigid transformations, \ie the matrix $\mathbf{A}_i\in\mathbb{R}^{3\times 3}$ should be close to a rotation matrix~\cite{amberg2007optimal,yoshiyasu2014conformal, yang2015sparse, li2019robust, yang2019global}. 

\begin{figure}[t]
	    \centering
	    \includegraphics[width=0.35\columnwidth]{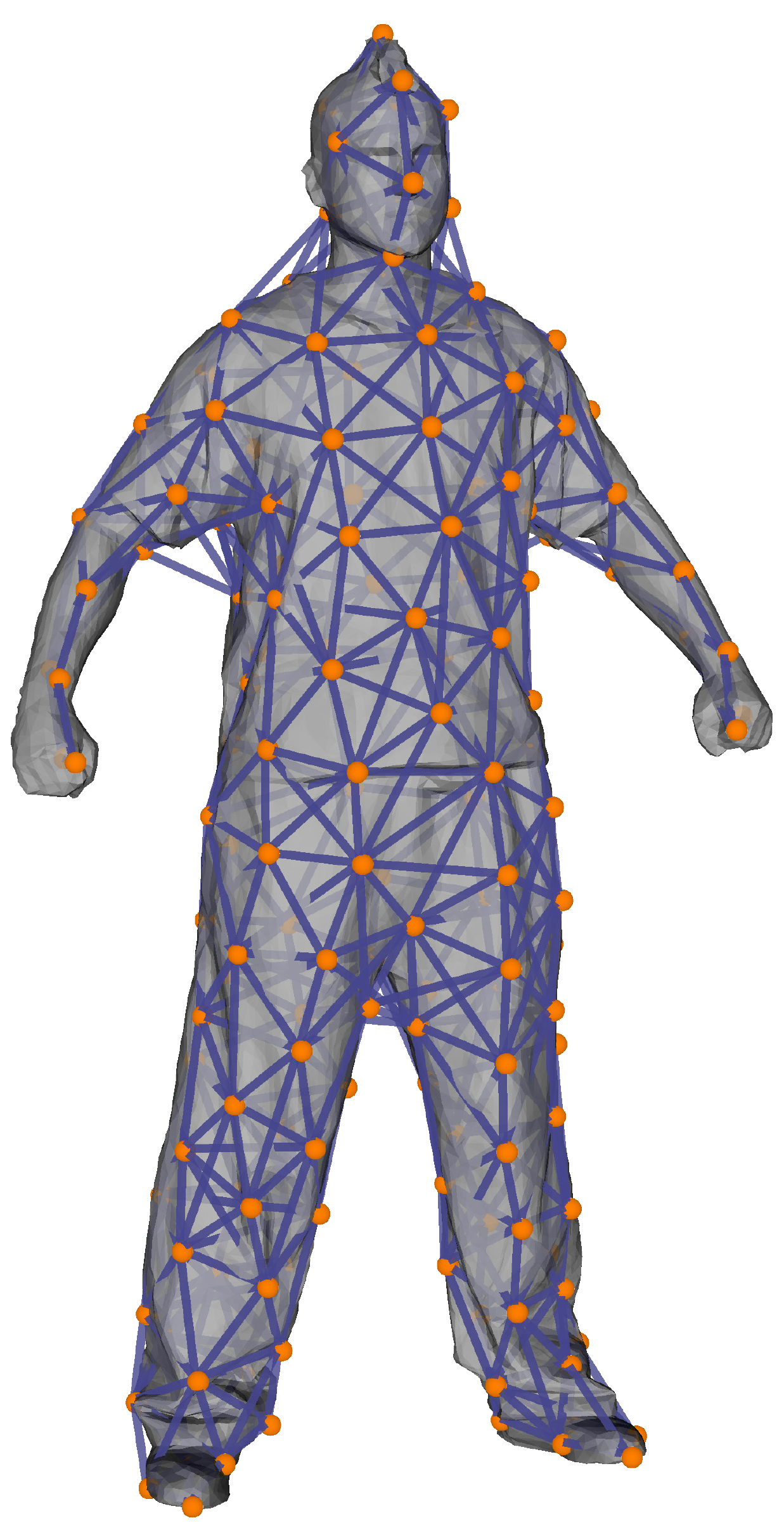}
    	\caption{A deformation graph defined over a mesh surface, with the graph nodes shown in orange.}
	    \label{fig:DeformationGraph}
	\end{figure}

    \paragraph*{Deformation graph-based methods} 
    The methods mentioned previously store the deformation field densely over each sample point on the source surface. This can lead to a large number of variables that incur high computational costs and memory consumption.
    To reduce the degrees of freedom and achieve better efficiency, some methods adopt an embedded deformation graph~\cite{sumner2007embedded} to represent the deformation~\cite{li2008global, li2009robust, bonarrigo2014deformable, cao2015two, lin2016color, li2017robust, li2017robust, yao2020quasi, li2020robust, zampogiannis2019topology}. These methods build a deformation graph with nodes embedded on the surface, and with edges indicating the proximity between the nodes (see Fig.~\ref{fig:DeformationGraph} for an example).
    The nodes are typically a subset of the sample points on the source surface.
    Each node is associated with an affine transformation and influences the deformation in its surrounding region on the surface.
    For a point $\mathbf{x}_i$ on the surface, the deformation is a convex combination of the transformations induced from the deformation graph nodes that influence $\mathbf{x}_i$. Specifically, let $\{\mathbf{p}_1, \ldots, \mathbf{p}_r\}$ be the set of nodes in the deformation graph. Then the new position of $\mathbf{x}_i$ can be computed as~\cite{sumner2007embedded}:
    \begin{equation}
		\label{Eq:ED_graph}
		\widehat{\mathbf{x}}_i = \sum_{\mathbf{p}_j\in\mathcal{I}(\mathbf{x}_i)} \omega_{ij}\cdot(\mathbf{A}_j(\mathbf{x}_i - \mathbf{p}_j) + \mathbf{p}_j + \mathbf{t}_j),
	\end{equation}
	where $\mathbf{A}_j\in\mathbb{R}^{3\times 3}$ and $\mathbf{t}_j\in\mathbb{R}^3$ represent the affine transformation corresponding to the node $\mathbf{p}_j$, $\mathcal{I}(\mathbf{x}_i)$ is the set of nodes that influence $\mathbf{x}_i$, and $\{\omega_{ij} \mid \mathbf{p}_j\in\mathcal{I} \}$ are convex combination weights determined by the distance from $\mathbf{x}_i$ to the nodes in $\mathcal{I}(\mathbf{x}_i)$.
	Similar to pointwise transformation, the affine transformation is often required to be close to a rigid transformation, which can be enforced via a regularization term when computing the deformation.
	Alternatively, some methods directly assign a rigid transformation to each node,  with the rigid transformation represented using Lie algebra, quaternions or dual quaternions~\cite{kavan2008geometric,dou2017motion2fusion, li2020robust,li2020robust2,zampogiannis2019topology}.
	In addition, for smoothness of the deformation, it is a common practice to use a regularization term that enforces consistency between the deformation induced by two neighboring nodes $\mathbf{p}_i$ and $\mathbf{p}_j$ on the same node $\mathbf{p}_j$~\cite{sumner2007embedded,li2008global}:
	\begin{equation}
	    E_{\smooth} = \sum_{i} \sum_{j \in \mathcal{N}_g(i)} \| \mathbf{A} (\mathbf{p}_j - \mathbf{p}_i) + \mathbf{p}_i + \mathbf{t}_i - (\mathbf{p}_j + \mathbf{t}_j) \|_2^2,
	   \label{eq:DefGraphSmoothness}
	\end{equation}
	where $\mathcal{N}_g(i)$ is the index set of neighboring nodes for $\mathbf{p}_i$.
	
	A deformation graph is typically constructed by selecting a subset of the sample points on the surface as nodes via uniform sampling, and connecting nearby nodes with edges~\cite{sumner2007embedded}.
    Since the number of deformation graph nodes is typically much smaller than the number of sample points on the surface, such a representation can significantly reduce the number of variables compared to pointwise transformations.
    Besides efficiency, the deformation graph also provides more flexibility in the algorithm: a user can change the number of nodes to control the balance between expressiveness and efficiency. It is also possible to adopt a non-uniform sampling that adapts to the semantics of the deformation~\cite{yu2018doublefusion}.
    
    \paragraph*{RKHS-based methods} Some methods model the deformation with a displacement field $\mathbf{d}(\cdot)$ over the source surface~\cite{myronenko2007non, myronenko2010point,ma2013robust, ma2015non, ma2017non}:
	\begin{equation}
		\widehat{\mathbf{x}}_i = \mathbf{x}_i + \mathbf{d}(\mathbf{x}_i).
		\label{Eq:pointwise_offset}
	\end{equation}
	The displacement field $\mathbf{d}(\cdot)$ is often constructed within a reproducing kernel Hilbert space (RKHS) using a Gaussian kernel ${\Gamma}: \mathbb{R}^3\times\mathbb{R}^3\mapsto\mathbb{R}^{3\times 3}$ with the form
	\[
	    {\Gamma}(\mathbf{x}, \mathbf{y}) = e^{-\beta\|\mathbf{x}-\mathbf{y}\|^2} \cdot \mathbf{I},
	\]
	where $\beta$ is a parameter and $\mathbf{I}\in\mathbb{R}^{3\times 3}$ is the identity matrix. The displacement at a point $\mathbf{x}_i$ can then be written as
	\begin{equation}
		\mathbf{d}(\mathbf{x}_i) = \sum_{j \in \mathcal{I
		}} \Gamma(\mathbf{x}_i, \mathbf{x}_j) ~\mathbf{c}_j,
		\label{Eq:RKHS_deform}
	\end{equation}
	where $\mathcal{I}$ is the index set for either all the source surface points or a subset of them, and $\mathbf{c}_j \in \mathbb{R}^3$ are the variables to be determined.
	The deformation is usually optimized via a probabilistic model, with regularization on the displacement function such as the coherence of motion of nearby points~\cite{myronenko2007non, myronenko2010point}.

    \paragraph*{Patch-based methods} Some methods deform the surface in a patch-based manner~\cite{huang2008nonrigid,cagniart2010free, kozlov2018patch}. They first segment the surface into a set of patches that are either disjoint or partially overlapping. Each patch is assigned with a separate rigid transformation, which induces deformation for points inside and adjacent to the patch. For globally consistent deformation, the new position of a surface point is computed by combining the deformations induced by the patches, typically using Gaussian weights related to the distance from the point to the patches. Such approaches lead to approximately rigid deformation with each patch. For registration between partially overlapping surfaces, the patch-based representation also helps to extend deformation to regions without correspondence to the target shape~\cite{huang2008nonrigid}.
    
    \cite{xu2014nonrigid} also perform registration with a patch-based deformation. Instead of rigid transformation, each patch is deformed by a displacement field using a linear combination of deformation modes learned from simulated data. When determining the deformation, a regularization term is included that penalizes deviation between the displacement fields of two neighboring patches on their common points, to achieve consistency across patches.

    \cite{liang20143d} align a depth image of a human face with a database of face shapes in a patch-based manner, to derive a high-resolution face mesh. The input face is divided into several patches corresponding to different facial parts. Each patch is first matched with the faces in the database to determine its new point positions and normals. The positions and normals from different patches are then fused into the complete 3D mesh.

    \paragraph*{Spline function-based methods} 
    Some methods transform the surface using a deformation field defined by a spline function over the 3D space. 
	One common approach is based on the thin plate spline (TPS)~\cite{Bookstein1989}, which is used to define a displacement field~\cite{chui2003new,Domokos2012nonlinear,santa2016face,fan2016convex,huang2019automatic}
	\[
		\mathbf{d}(\mathbf{x}) = \mathbf{A} \mathbf{x} + \mathbf{b} + \sum_{i} \mathbf{w}_i Q(\|\mathbf{c}_i-\mathbf{x}\|),
	\] 
	where $\{\mathbf{c}_i\}$ is a set of pre-selected control points, and $\mathbf{A} \in \mathbb{R}^{3 \times 3}$, $\mathbf{b}  \in \mathbb{R}^3$ and $\mathbf{w}_i \in \mathbb{R}^3$ are parameters that control the displacement field, and $Q(\cdot)$ is a radial basis function. The control points can be either placed on a uniform grid~\cite{Domokos2012nonlinear}, or chosen based on the feature correspondence between the two surfaces~\cite{huang2019automatic}.

	Another common approach is free-form deformation (FFD)~\cite{Sederberg1986FFD}, which transforms a point $\mathbf{x} = (x_1,x_2,x_3)$ via 
	\[
		\mathbf{\widehat{X}} = \sum_{i}\sum_{j}\sum_{k} \mathbf{p}_{ijk} B_i(x_1)B_j(x_2)B_k(x_3),
	\]  
	where $B_i, B_j, B_k$ are spline basis functions, and $\{\mathbf{p}_{ijk}\}$ are control points that are to be determined. These methods usually select regular grid points as the control points to deform~\cite{Rueckert1999,rouhani2012non}.

    \paragraph*{Grid-based methods} Some methods define a surface implicitly as a level set of a scalar function~\cite{fujiwara2011locally, newcombe2015dynamicfusion,zhang2015efficient,innmann2016volumedeform,slavcheva2017killingfusion}. 
    Typically, the scalar function is a signed distance function (SDF) or a truncated signed distance function (TSDF). The SDF at a point $\mathbf{x} \in \mathbb{R}^3$ can be written as~\cite{fujiwara2011locally}
    \begin{equation}
		\sdf(\mathbf{x}) = (\mathbf{x} - \mathbf{x}^*) \cdot  \mathbf{n}^*,
		\label{eq:SDF}
	\end{equation}
	where $\mathbf{x}^*$ is the closest point to $\mathbf{x}$ on the surface, and $\mathbf{n}^*$ is the unit normal vector of the surface at $\mathbf{x}^*$. Note that the definition above is applicable also for surfaces represented as point clouds: in this case, $(\mathbf{x} - \mathbf{x}^*) \cdot  \mathbf{n}^*$ is the signed distance from $\mathbf{x}$ to the tangent plane at $\mathbf{x}^*$ which is a first-order approximation of the underlying surface~\cite{hoppe1992surface}.
	Considering that points far away from the surface have little effect on the level set, the TSDF truncates the value of $\sdf(\mathbf{x})$ beyond a certain threshold:
	\[
		\tsdf(\mathbf{x}) = \max\{-1, \min\{1, \sdf(\mathbf{x})/t\}\},
	\]
	where $t$ is a threshold parameter. The SDF or TSDF is often discretized  using its values evaluated at a set of points sampled as a uniform grid within a volume that encloses the surface.
	A deformation field defined over the same volume is then used to transform the implicit surface. The deformation field can be represented discretely either on the same grid as the SDF or TSDF~\cite{zhang2015efficient,innmann2016volumedeform,slavcheva2017killingfusion}, or on a coarser grid for better efficiency~\cite{newcombe2015dynamicfusion}.
	The deformation of a point on the implicit surface is interpolated from nearby grid points in the deformation field.
	Alternatively, some methods use such a deformation field to move a set of FFD control points embedded in the same volume, and transform the surface via FFD~\cite{fujiwara2011locally}.
	To achieve local rigidity, the deformation field is either represented using a rigid transformation at each grid point~\cite{fujiwara2011locally,newcombe2015dynamicfusion,zhang2015efficient,innmann2016volumedeform}, or using a translation at each grid point together with regularization that maintains the distance between points after the deformation~\cite{slavcheva2017killingfusion}.

    \paragraph*{Prior-based representations} 
    Some methods deform the surface in specific ways based on priors about the underlying shape. 
    One example is articulated shapes, which can be deformed using skeleton models~\cite{allen2002articulated, pekelny2008articulated, gall2009motion, zheng2010consensus, yu2017bodyfusion}. For example, in~\cite{gall2009motion} the human body motion is modeled using a kinematic skeleton, which consists of straight edges representing the bones, connected at end points representing body joints. The movement of the skeleton is parameterized by a global rigid motion together with the joint angles. The surface is then deformed via skinning~\cite{Jacobson2014skinning}, which transfers the skeleton motion to the surface points. 
    Compared with deformation graphs, such skeletons encode a stronger prior of the structure of the underlying shape and can often represent the deformation behavior with fewer degrees of freedom.
    Some methods also combine a skeleton with a deformation graph or pointwise displacements to model deformations with more detail~\cite{bogo2015detailed,  yu2017bodyfusion, yu2019simulcap,yu2018doublefusion, su2020robustfusion}.
    Besides skeletons, other representations have been proposed to model articulated structures. For example, the shape of human hands is modeled using a combination of spheres and cylinders in~\cite{tagliasacchi2015robust}, and using sphere-meshes in~\cite{tkach2016sphere}.
    The SMPL~\cite{loper2015smpl}, a learned parameterized human body model, has been used for the non-rigid registration of 3D human shapes~\cite{bogo2017dfaust,yu2018doublefusion,marin2020farm}.
    
    The shape of human faces is another example where non-rigid registration can benefit from shape priors. Some parameterized models that capture the variation of human face shapes have been used successfully for facial reconstruction and tracking~\cite{Bouaziz2016Modern}. For example, the morphable model from~\cite{blanz1999morphable} provides a linear representation for the geometry and texture of the human face, and can be used to reconstruct a neutral face model via alignment to scanned data~\cite{weise2011realtime,Ichim2015Dynamic}. Different facial expressions for the same person can be modeled using a linear blendshape model, which can be used for real-time face tracking via registration~\cite{weise2011realtime}.
    The bilinear face model from~\cite{cao2014facewarehouse} is used in~\cite{Achenbach2015Accurate} to align a template face mesh with a point cloud.

	\paragraph*{Summary and discussion} 
	Each representation of deformation has its benefits and limitations. Pointwise position variables and pointwise affine transformations provide an abundance of degrees of freedom to model the deformation in detail, with pointwise affine transformations having the additional benefit of making it easy to model local rotations.
	However, they require a large number of variables and may induce a high computational cost.
	Representations based on deformation graphs, RKHS and spline functions can reduce the degrees of freedom needed to model a deformation. This can be suitable for deformations that have a low-dimensional structure. For example, deformation graphs are commonly used for registration between shapes of the same human subject with different poses, where the deformation is primarily induced by the articulation of joints.
	Meanwhile, such lower-dimensional representations may be less capable of modeling fine-scale variations in the deformation field, and the control structure may need to be carefully selected to account for the topology of the shape and to ensure expressiveness.
	Patch-based representations can be suitable when the shape consists of multiple regions and the points within the same region undergo similar deformation.
	Additional care needs to be taken to ensure a smooth transition of the deformation across the boundaries between patches, which can increase the complexity of the algorithm.
	Grid-based methods provide implicit representations for the surfaces, which can better handle topological changes between the source and the target surfaces. However, they can incur high memory consumption when a high-resolution grid is used.
	Prior-based representations can effectively utilize the characteristics of specific classes of shapes such as human faces and hands. On the other hand, their performance may degrade for shapes that are not covered by the parametric model.

	\section{Extrinsic Methods}
	\label{sec:extrinsic}
	
	Given the representation of a deformation field, non-rigid registration amounts to searching the space of deformations for one solution that best aligns the two surfaces. Therefore, it is important to design criteria to measure the alignment quality. A large class of methods measures the alignment error using the distance between two surfaces in the ambient 3D space. These methods are called extrinsic methods in this paper, since they involve the extrinsic distance between surfaces. This section reviews extrinsic methods in detail.

	\subsection{Optimization-Based Methods}
	
	Many methods compute the deformation by minimizing a target function $E$ about the deformation, typically formulated as
	\begin{equation}
		E = E_{\text{align}} + \alpha E_{\text{reg}},
	\end{equation}
	where $E_{\text{align}}$ measures the alignment error between the deformed source surface and the target surface, and $E_{\text{reg}}$ is a regularization term that enforces constraints on the deformation such as smoothness. $\alpha > 0$ is a weight that controls the balance between these two terms.
	Various forms of the alignment term and the regularization term have been proposed in the past.
	In the following, we review some representative formulations of the terms.

    \subsubsection{Alignment term}
    \label{sec:alignment}
    
    A simple and popular alignment term is based on the distance from each source surface point to the target surface~\cite{besl1992method}: 
	\begin{equation}
		\label{Eq:align_point_to_point}
		E_{\text{pp}} = \sum_{i = 1}^m \wid \|\widehat{\mathbf{x}}_i - \mathbf{y}_{\rho(i)}\|^2,
	\end{equation}
	where $\rho(i)$ is the index of the corresponding point on the target surface for the point $\widehat{\mathbf{x}}_i$, and $\wid$ is a weight that can be used to control the influence of different points based on the reliability of their correspondence.
	Such point-to-point distance terms have been used in many registration methods~\cite{allen2003space, pauly2005example, amberg2007optimal, li2008global, liao2009modeling, chang2011global, hontani2012robust, rouhani2012non, yoshiyasu2014conformal}.
	The correspondence can be identified by either searching for the closest point on the target surface~\cite{allen2003space,amberg2007optimal}, formulating the correspondence as optimization variables~\cite{li2008global}, or matching features such as scene flows~\cite{xu2017flycap}. 

	Some methods measure the alignment using the following function instead~\cite{chen1992object}:
	\begin{equation}
	    \label{Eq:align_point_to_plane}
		E_{\ppl} = \sum_i \wid \left(\mathbf{n}_{\rho(i)} \cdot (\widehat{\mathbf{x}}_i - \mathbf{y}_{\rho(i)})\right)^2,
	\end{equation}
	where $\mathbf{n}_{\rho(i)}$ is the unit normal of the target surface at the corresponding point $\mathbf{y}_{\rho(i)}$ for $\mathbf{x}_i$. The term $\left(\mathbf{n}_{\rho(i)} \cdot (\widehat{\mathbf{x}}_i - \mathbf{y}_{\rho(i)})\right)^2$ measures the squared distance from $\widehat{\mathbf{x}}_i$ to the tangent plane at $\mathbf{y}_{\rho(i)}$, and is often called the point-to-plane distance. In rigid registration, it is well-known that the ICP algorithm based on the point-to-plane distance can converge faster than the point-to-point distance, as the tangent plane provides a first-order approximation of the local surface shape~\cite{pottmann2006geometry}.
	The point-to-plane distance has been used in many non-rigid registration algorithms~\cite{li2009robust,xu2017flycap,li2020robust2}. It can be used either as the alignment term alone~\cite{dou2016fusion4d, yu2017bodyfusion,dou2017motion2fusion,li2018articulatedfusion,yu2018doublefusion,zhang2019interactionfusion,li2020robust2,li2020robust}, or in combination with the point-to-point distance in~\eqref{Eq:align_point_to_point}~\cite{li2009robust,chang2011global,yamazaki2013non, Achenbach2015Accurate, guo2015robust,  thomas2016augmented, innmann2016volumedeform,li2017robust, wang2017templateless,xu2017flycap,li2020topology,su2020robustfusion,zampogiannis2019topology}.

	If the target surface is represented implicitly with an SDF or TSDF and the source surface is represented explicitly, then the distance from a source point $\mathbf{x}_i$ to the target surface  can be simply evaluated as $|D (\mathbf{x}_i)|$, where $D(\cdot)$ is the SDF or TSDF. The alignment term can then be defined as
	\begin{equation}
	    E_{\text{align}} = \sum_i \wid \phi(|D(\mathbf{x}_i)|),
	\end{equation}
	where $\phi$ is an increasing function on $[0, +\infty)$. This definition has been used in~\cite{dou2017motion2fusion,xu2019unstructuredfusion,su2020robustfusion,li2020robust2} for aligning a template mesh with a TSDF.

    If both surfaces are represented using SDFs, the alignment term can be defined as the sum of the squared difference between the SDFs of the deformed source surface and the target surface, evaluated over some sample points in an enclosing volume~\cite{fujiwara2011locally, zhang2015efficient, slavcheva2017killingfusion}.
    The underlying idea is that if the two surfaces are entirely aligned, their SDFs should be the same everywhere.

	Some methods consider the alignment from a probabilistic perspective, typically using mixtures of Gaussians~\cite{jian2005robust,myronenko2007non,myronenko2010point,Jian2011robust,ma2013robust,ge2014non,ge2015non,ma2015robust,ma2015non}. 
	Their alignment terms are formulated as a probabilistic measure, such as the negative log posterior probability~\cite{myronenko2007non,myronenko2010point} and the $\ell_2$ difference between probability density functions~\cite{jian2005robust,Jian2011robust}. Such probabilistic approaches are discussed in detail in Section~\ref{sec:probabilistic}.
    
    \subsubsection{Regularization term}
    \label{sec:regularization}
    
    The regularization term can be a weighted sum of multiple terms, each enforcing a different requirement on the deformation field. Below we review the main types of regularization that have been used in the literature.
    
    \paragraph*{Smoothness}
    To avoid unnatural deformed shapes, it is a common requirement that the deformation field should be smooth. This can be achieved using different types of regularization terms according to the representation of the deformation. For deformation fields represented as pointwise transformations, the smoothness can be enforced by penalizing the difference between transformations on neighboring points. For example, in~\cite{allen2003space} the surface is deformed by pointwise affine transformations, and the regularization term for smoothness is defined as
    \begin{equation}
        E_{\smooth} = \sum_{i}\sum_{j\in\mathcal{N}(i)}\|\mathbf{T}_i - \mathbf{T}_j\|_F^2, \label{Eq:smooth_adjacent_trans}
    \end{equation}
    where the matrix $\mathbf{T}_i$ represents the affine transformation on $\mathbf{x}_i$, $\|\cdot\|_F$ is the Frobenius norm, and $\mathcal{N}(i)$ is the index set of the neighboring vertices for $\mathbf{x}_i$. Similar smoothness terms have also been used in~\cite{pauly2005example, amberg2007optimal,fan2018dense,zampogiannis2019topology}.
    
    For deformations defined using a deformation graph, the smoothness can be enforced via consistency between the deformations on a graph node that is induced by a neighboring node and the node itself, resulting in a smoothness term as shown in~\eqref{eq:DefGraphSmoothness}. This regularization is commonly used in registration methods using deformation graphs~\cite{li2008global,li2009robust,guo2015robust,cao2015two,wang2017templateless,xu2017flycap,yao2020quasi}. 
    
    For RKHS-based deformation fields, the smoothness is often induced by penalizing a norm of the deformation field within the RKHS. For example, the deformation field in~\cite{myronenko2007non} is defined using Gaussian kernels as shown in~\eqref{Eq:RKHS_deform}, and the regularization is defined as
    \begin{equation}
        E_{\smooth} = \sum_{i \in \mathcal{I}} \sum_{j \in \mathcal{I}} 
		\Gamma(\mathbf{x}_i, \mathbf{x}_j) ~\mathbf{c}_i \cdot \mathbf{c}_j.
		\label{Eq:smooth_RKHS}
    \end{equation}
    Similar regularizations are also used in~\cite{jian2005robust, myronenko2010point,ma2013robust,ge2014non,ge2015non,ma2017non}.

    \paragraph*{Positional constraints}
    Depending on the application, some points on the deformed surface are required to be close to some  target positions. For example, when aligning a template mesh for human bodies or heads to scanned data, the landmark vertices on the deformed template mesh are required to be close to the landmark locations on the data~\cite{allen2003space,sorkine2004least,amberg2007optimal,yeh2010template,thomas2016augmented,li2020topology,zampogiannis2019topology}. Such positional constraints can be enforced with the following term:
    \begin{equation}
        E_{\pos} = \sum_{\mathbf{x}_i\in \mathcal{C}} \|\widehat{\mathbf{x}}_i - \mathbf{m}_i\|^2, \label{Eq:landmark_term}
    \end{equation}
    where $\mathcal{C}$ is the set of  source surface points with positional constraints, and $\mathbf{m}_i$ is the target position for $\mathbf{x}_i$.

    \paragraph*{Local shape preservation}
    When deforming the source surface, it is desirable to prevent distortion of the local surface. Depending on the assumption about the deformation, different types of regularizations can be introduced to achieve this goal.
    
    Many methods assume the deformation to be locally rigid, \ie each local region undergoes an approximately rigid transformation while the whole surface is deformed in a non-rigid way to align with the target. 
    Various forms of regularization have been proposed to achieve such local rigidity.
    For many methods that assign affine transformations to either the source points or the deformation graph nodes, regularization terms are introduced to require each affine transformation to be close to a rigid transformation. Some methods use the following term~\cite{li2008global,li2009robust, guo2015robust, cao2015two, wang2017templateless,xu2017flycap}:
	\begin{equation}
		E_{\rigid} = \sum_{i} \rot{} (\mathbf{A}_i),
	\end{equation} 
	where $\mathbf{A}_i \in \mathbb{R}^3$ denotes the transformation matrix for the affine transformation at a graph node, and  
		\begin{align*}
		\rot{}(\mathbf{A}_i) = &
		(\mathbf{a}_{i,1} \cdot \mathbf{a}_{i,2})^2 +
		(\mathbf{a}_{i,1} \cdot \mathbf{a}_{i,3})^2 +
		(\mathbf{a}_{i,2} \cdot \mathbf{a}_{i,3})^2\\	
		& + (1 - \mathbf{a}_{i,1} \cdot \mathbf{a}_{i,1})^2
		+ (1 - \mathbf{a}_{i,2} \cdot \mathbf{a}_{i,2})^2
		(1 - \mathbf{a}_{i,3} \cdot \mathbf{a}_{i,3})^2
		\end{align*}
	with $\mathbf{a}_{i,1}, \mathbf{a}_{i,2}, \mathbf{a}_{i,3}$ being the three column vectors of $\mathbf{A}_i$. The function $\rot{} (\mathbf{A}_i)$ enforces the condition $\mathbf{A}_i^\top \mathbf{A}_i = \mathbf{I}$ by penalizing the deviation between the two matrices using their upper triangular elements.
	This condition alone does not guarantee $\mathbf{A}_i$ to be a rotation matrix, since a reflection matrix $\mathbf{A}_i$ with orthonormal columns and $\det(\mathbf{A}_i) = -1$ can still satisfy the condition. In~\cite{dou2016fusion4d}, an additional term $(\det(\mathbf{A}_i)-1))^2$ is introduced to avoid such reflections. 
	Alternatively, some methods enforce local rigidity by penalizing the deviation between the matrix $\mathbf{A}_i$ and its closest rotation matrix~\cite{yang2015sparse,guo2017global,li2019robust,yang2019global,yao2020quasi}.
	A similar approach is adopted in~\cite{Sahillioglu2015ashape} for the registration of tetrahedral meshes, by penalizing the deviation between the deformation gradient of each tetrahedron in the source mesh and its closest rotation matrix; an additional Tikhonov regularizer is also introduced to penalize the change of point positions between compared with the previous iteration, which helps to prevent element inversions and handle large deformations.
	
	Since local rigidity implies that the distance metric is also preserved locally, some methods enforce the condition by penalizing the change of distance between each point and its neighboring points~\cite{wand2007reconstruction, sussmuth2008reconstructing, innmann2016volumedeform,yang2019global,li2020topology}:
	\begin{equation}
		E_{\text{ARAP}} = \sum_{i=1}^m \gamma_{i}\sum_{j\in\mathcal{N}(i)} \gamma_{ij}\|\mathbf{R}_i(\mathbf{x}_j - \mathbf{x}_i) - (\widehat{\mathbf{x}}_j - \widehat{\mathbf{x}}_i)\|^2_2,
		\label{eq:ARAP}
	\end{equation}
	where $\mathbf{R}_i$ is an auxiliary variable for a local rotation, and $\gamma_{i}, \gamma_{ij}$ are weights. Unlike previous formulations of local rigidity, this regularization term can be applied to general deformation fields, not just those represented with affine transformations.

	The local rigidity conditions require the deformation to be \emph{near-isometric}. Other regularizations based on weaker conditions have also been proposed to allow for \emph{non-isometric} deformations. 
	Since the Laplacian at each point encodes the local geometry~\cite{alexa2003differential}, some methods penalize the change of the Laplacians after the deformation (potentially up to rotation and scaling)~\cite{liao2009modeling,huang2011global,Achenbach2015Accurate,ge2015non}.
	Meanwhile, in~\cite{yamazaki2013non,aigerman2014lifted,jiang2017consistent,jiang2019huber} the deformation is required to be locally close to a similarity transformation. 
	In other works, \cite{yoshiyasu2014conformal,Wu2019global} require the deformation to be locally as conformal as possible, while 
	\cite{wand2009efficient} introduce a regularization for local volume preservation.

    \paragraph*{Parameter regularization}
    In some methods that rely on a parameterized shape model for the deformation, regularization terms for the shape parameters are introduced to obtain more natural results. In~\cite{Bouaziz2013online}, the registration for facial tracking uses a linear blendshape model to represent facial expressions, and the target function includes a regularization term that penalizes the $\ell_2$-norm of the blendshape coefficients.
    In~\cite{tagliasacchi2015robust}, a parameterized hand model is aligned with the scanned data, using a regularization term that penalizes the deviation between the hand parameters and a PCA subspace that indicates feasible poses.

    \paragraph*{Other regularizations}
    Other types of regularization can be used according to the specific need of an application. For example, the registration between human body motion frames in~\cite{li2017robust} introduces a regularization for the temporal coherence of deformation between adjacent frames. For the registration of hand shapes in~\cite{tagliasacchi2015robust}, regularization terms are introduced to prevent self-collision and avoid impossible poses.

	\subsubsection{Improving robustness}
	Many alignment terms and regularization terms are formulated in a sum-of-squares form, which can be the squared $\ell_2$ norm of an error vector. 
	The minimization of such an $\ell_2$-norm will attempt to reduce the magnitude of all the error components,  which may not be suitable for some cases. 
	For example, when the two point sets contain outliers or overlap partially, some source points may not correspond to any point on the target surface.
	If we minimize the point-to-point distance~\eqref{Eq:align_point_to_point} with the same weights for all points, then reducing the distance for the points without true correspondence may result in an erroneous alignment. 
	One way to address this issue is to choose a proper weight $\wid$ to reduce the influence of source points without reliable correspondence on the target surface. This can be done either by setting the weights according to specific quality criteria of the correspondence, or by jointly optimizing the weights such that they are adjusted automatically~\cite{amberg2007optimal, li2008global, chang2011global,zhang2015efficient,lin2016color,innmann2016volumedeform,kozlov2018patch}. 
 	
 	Another popular approach is to perform sparsity optimization on the distance values. The idea is to minimize a robust norm that can reduce the magnitudes of many error components while allowing some components to gain large values, thus promoting the sparsity of the error vector~\cite{Bach2012Optimization}.
 	This can be applied to not only the alignment term, but also a regularization term such as the smoothness of the transformation. Sparsity optimization for the regularization error is beneficial in some applications, because this would allow a deformation that is less smooth in some local regions such as the joints for an articulated motion. 
 	
	 A popular choice of sparsity-inducing norms is the $\ell_p$-norm with $0 \leq p \leq 1$.
	 In~\cite{guo2015robust}, the $\ell_0$-norm is applied to the smoothness term for the affine transformations defined over a deformation graph. \cite{yang2015sparse} apply the $\ell_1$-norm to the smoothness term for the affine transformations on each point. In~\cite{jiang2019huber}, the $\ell_1$-norm is applied to the alignment term instead. \cite{li2019robust} and~\cite{yang2019global} apply the $\ell_1$-norm to both the smoothness term and the alignment term. 
	 
	 Other robust norms have also been used in the literature. The Tukey function is applied to the point-to-plane alignment term in~\cite{newcombe2015dynamicfusion}. The Geman-McClure function is used in the alignment terms in~\cite{yu2018doublefusion}, \cite{xu2019unstructuredfusion} and~\cite{su2020robustfusion}. The Huber-$L_1$ loss is applied to the smoothness term in~\cite{jiang2019huber} and~\cite{zampogiannis2019topology}. The Welsch function is used for both the alignment term and the smoothness term in~\cite{yao2020quasi}.

	\subsubsection{Probabilistic models}
	\label{sec:probabilistic}
	
	The optimization-based methods above penalize the geometric distance between the two surfaces. Another class of methods models the alignment accuracy from a probabilistic perspective instead.
	
	A popular probabilistic approach is coherent point drift (CPD) and its variants.
	The original CPD method~\cite{myronenko2007non} considers the source point set as the centroids of equally-weighted Gaussians with equal isotropic covariance matrices in a Gaussian Mixture Model (GMM), and considers the target point set as the data points. 
	Using an RKHS-based displacement field \eqref{Eq:RKHS_deform} to deform the source surface, they align the two surfaces by minimizing the posterior probability, which is equivalent to minimizing the following alignment term
	\begin{equation}
	 	E_{\text{align}} = -\sum_{j=1}^{n} \log \sum_{i=1}^{m} \exp({-\frac{1}{2} \left\| \frac{\mathbf{y}_{j}-\widehat{\mathbf{x}}_{i}}{\sigma} \right\|^{2}}),
	 	\label{eq:CPD_align}
	\end{equation}
	where $\sigma$ is the standard deviation parameter for the Gaussians.
	The above measure is combined with a regularization term for the smoothness of the displacement to derive the target function.
	In~\cite{myronenko2010point}, an additional uniform distribution is introduced to the GMM to account for noise and outliers.
	In~\cite{ge2014non} and~\cite{ge2015non}, the CPD framework is used with additional regularization terms that preserve local structures.
	
    Local structures are also utilized in~\cite{ma2015non} for registration. Similar to CPD, they use a GMM to model the relationship between the two point sets. Unlike the CPD that uses equal membership probabilities in the mixture model, they use local shape features to find correspondence between point sets and determine the membership probabilities. 
    The correspondence and the probabilities are continuously updated during the optimization according to the deformation.
    \cite{ma2013robust} and~\cite{ma2015robust} adopt a similar strategy and use  feature-based correspondence to facilitate registration. Instead of maximum likelihood estimation, they update the deformation according to the correspondence via $L_2 E$~\cite{scott2001parametric}, a robust estimator related to minimizing the $L_2$ distance to the ground-truth density function.

	\cite{hirose2020bayesian} formulates the CPD problem in a Bayesian setting and proposes a solver with guaranteed convergence as well as an acceleration scheme. The Bayesian formulation was also used in~\cite{hirose2020acceleration} to accelerate non-rigid registration, by first performing registration on a downsampled source point set and then interpolating the deformation using Gaussian process regression.

	Different from CPD,  in~\cite{jian2005robust} and~\cite{Jian2011robust} each of the two point sets is modeled as a mixture of Gaussians, and the registration is formulated as a problem of aligning the two mixtures. 
	They measure the alignment error using the $L_2$ distance between the probability density functions of the two Gaussian mixtures.
	Using either a TPS or RKHS representation for the deformation field, the above measure is combined with a regularization term for the deformation field, which is then minimized to derive the deformation.

	\subsubsection{Initialization}
	Typically, numerical optimization only finds a local minimum near the initial solution. Therefore, it is often necessary to find a proper initialization in order to achieve the desired result.
	A simple and popular approach is to first compute a rigid transformation and (optionally) a uniform scaling that roughly align the two surfaces~\cite{li2008global,liao2009modeling,yamazaki2013non,fan2018dense,yao2020quasi,sahilliouglu2021scale}, using the ICP algorithm or its variants.
	
	When there is a significant non-rigid deformation between the two surfaces, more sophisticated initialization may be needed. Many methods initialize the optimization using sparse correspondence that is computed by matching features on the two surfaces. These features are often identified using shape descriptors.
	
	\begin{figure}
	    \centering
	    \includegraphics[width=\columnwidth]{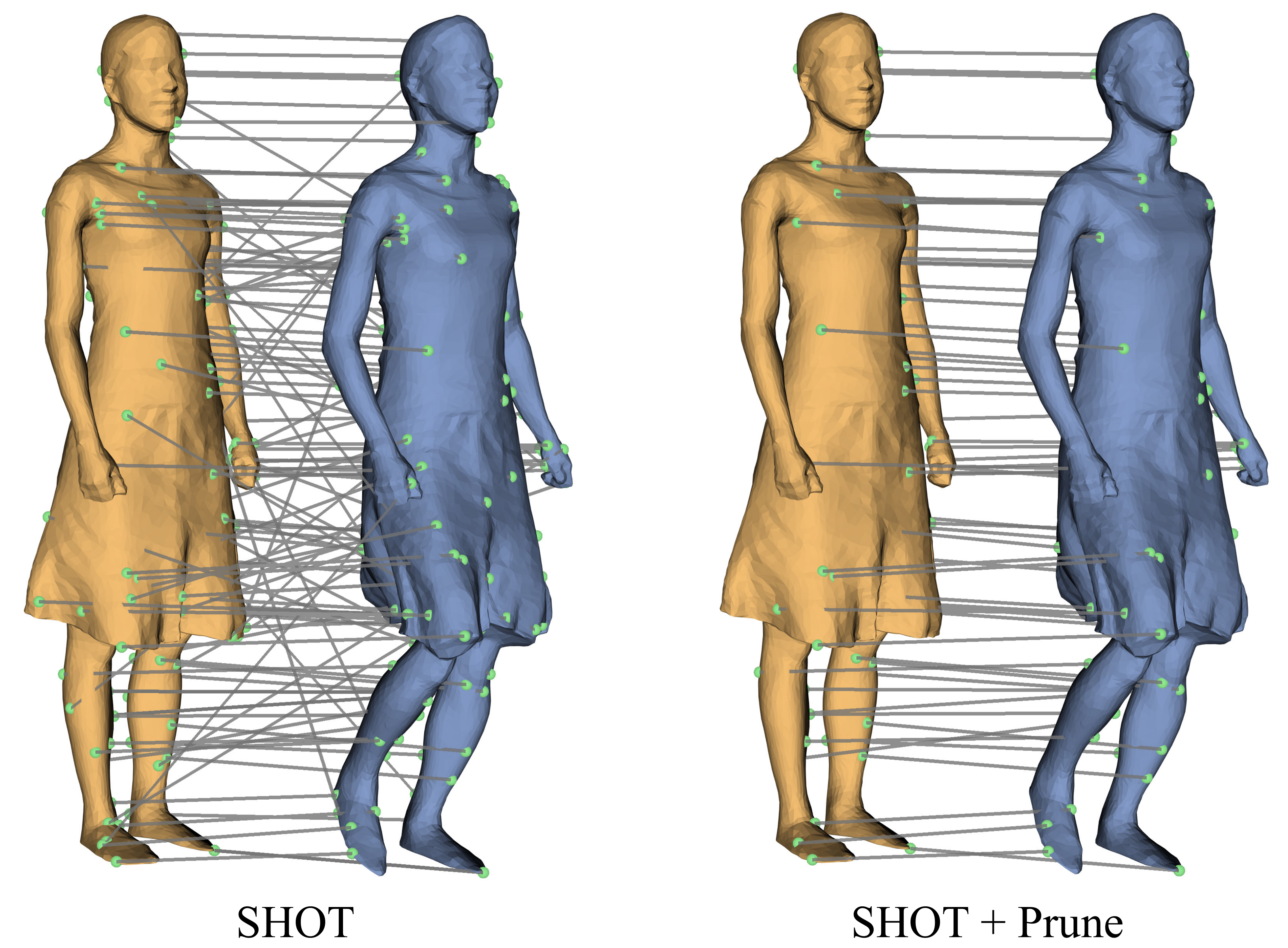}
	    \caption{Sparse initial correspondence between two models, computed by matching SHOT descriptors~{\cite{salti2014shot}} followed by pruning the correspondence~{\cite{tam2014diffusion}}.}
	    \label{fig:initialization}
	\end{figure}
	
	\cite{chang2008automatic} construct correspondence using spin-images~\cite{johnson1997spin}, a descriptor for local geometry. They randomly subsample the points on the two surfaces to compute their spin-images, and find the point pairs whose spin-images are significantly more similar than others.

    \cite{golyanik2016extended} extract 3D keypoints on the point clouds with an Intrinsic Shape Signature descriptor~\cite{zhong2009intrinsic}, \cite{rusu2008aligning} compute Persistent Feature Histograms (PFH) at the keypoints, and then compare the PFHs to determine initial correspondence.
	\cite{ma2015non, ma2017non} adopt Fast Point Feature Histograms (FPFH)~\cite{rusu2009fast} descriptors, which simplify the calculation of PFH and achieve comparable discriminative power.
	
	\cite{yang2015sparse, li2019robust, yang2019global, yao2020quasi} find the initial correspondence using the SHOT descriptor~\cite{salti2014shot}, which is more robust to noisy data. They first match the closest SHOT descriptors, and then use a diffusion pruning method~\cite{tam2014diffusion} to prune the correspondences with inconsistent geodesic distances on the source surface and the target surface. Further discussion of pruning techniques may be found in Section~\ref{sec:corr-pruning}.
	
	Besides the 3D geometry itself, auxiliary information can also be used to identify initial correspondence.
	For example, for input surfaces in RGB-D format, some methods first use compute SIFT features~\cite{lowe2004distinctive,liao2009modeling, innmann2016volumedeform, li2020topology, zampogiannis2019topology} on the RGB images to identify corresponding pixels; the correspondence is then transferred to the 3D surfaces using the association between the RGB and the depth data~\cite{zheng2010consensus, ma2015robust, innmann2016volumedeform}.

	\subsubsection{Numerical solvers}
	
	The optimization problem for non-rigid registration is typically non-linear and non-convex, and the solution is often computed using an iterative numerical solver. 
	Problems with a differentiable target function can be solved using gradient-based solvers. Each iteration of such solvers typically amounts to using the gradient information to determine a search direction, followed by a line search along the direction to find new variable values that lower the target function.
	Examples of such solvers include quasi-Newton methods such as BFGS or L-BFGS~\cite{jorge2006numerical}, which are popular for unconstrained optimization problems~\cite{allen2002articulated, allen2003space, wand2009efficient, ma2013robust, yin2016full, yao2020quasi}.
	If the target function has a sum-of-squares form, then the problem can also be solved using specialized non-linear least squares (NLLS) solvers such as Gauss-Newton or Levenberg-Marquardt~\cite{wand2007reconstruction,li2008global,li2009robust, chang2011global,santa2013correspondence,yoshiyasu2014conformal,bonarrigo2014deformable, zhang2015efficient, chebrolu2020spatio, zampogiannis2019topology}. 
	Such quasi-Newton and NLLS solvers can achieve a super-linear convergence rate~\cite{jorge2006numerical}, and there are established open-source implementations such as Ceres Solver for Levenberg-Marquardt~\cite{ceres-solver}. For a complicated target function, the derivation and computation of its gradient can be non-trivial and error-prone. This issue is mitigated by automatic differentiation tools provided by software libraries such as Ceres Solver, which only require the user to implement the target function and can automatically compute its derivatives.
	
	Despite the popularity of gradient-based solvers, they are not applicable to all problems. For example, robust norms can lead to a non-smooth target function that cannot be handled using gradient-based solvers. Such non-smooth problems are often solved using a first-order method~\cite{beck2017first} instead. The main idea is to introduce auxiliary variables to reformulate an equivalent optimization problem, whose solution can be computed using a method that involves simple sub-problems that are easy to solve. For example, block coordinate descent~\cite{tseng2001convergence} is used to solve the reformulated problem in~\cite{guo2015robust}, while ADMM~\cite{Boyd2011ADMM} is used in~\cite{yang2015sparse,li2019robust,yang2019global}.
	One benefit of first-order solvers is that their sub-problems often have closed-form solutions that can be computed in parallel, allowing them to benefit from multi-core systems to improve efficiency.
	
	The idea of decomposing a problem into sub-problems with closed-form solutions is not limited to non-smoooth problems.
	In~\cite{amberg2007optimal}, it is noted that the target function can be minimized in closed-form by solving a linear system if the point correspondence is fixed. Thus they alternate between point correspondence search and linear solve, resulting in a procedure similar to the ICP algorithm for rigid registration. Such ICP-like solvers are also used in~\cite{yamazaki2013non,yoshiyasu2014conformal,innmann2016volumedeform}.
	For methods based on probabilistic models, the problem is often solved using an EM algorithm consisting of alternating updates with closed-form steps~\cite{myronenko2007non, myronenko2010point, ge2015non, ma2017non, kozlov2018patch, gao2019filterreg, hirose2020bayesian}.

	\subsection{Learning-Based Methods}
	With a suitable target function, an optimization-based registration method tends to work well when the two surfaces are close. For problems that involve large deformations, optimization methods may converge to an undesirable local minimum if a reliable initial correspondence is not provided. Such a correspondence is not always easy to compute: closest point correspondence is sensitive to the relative position between the surfaces, while shape descriptors may still be ambiguous. For such challenging cases, prior knowledge learned from large-scale data can be beneficial. With the fast development of deep learning in recent years, various learning-based methods have been proposed for non-rigid registration. 
	These methods train a model{\textemdash}typically a neural network{\textemdash}using a set of example data,   such that the trained model can reliably predict the deformation or correspondence for unseen data. The training is done by searching for model parameters to minimize a loss function that is evaluated using the training data. 
	In the following, we will review some representative exiting works.

	\subsubsection{Supervised methods}
	A supervised method trains the model using data with ground-truth labels, such as the correspondence or the deformation field. The aim is to train a model that can predict the unknown ground truth for unseen data with reasonably high accuracy.
	
    \paragraph*{Shape correspondence} Some methods focus on finding correspondence between shapes. 
	\cite{wei2016dense} propose a method to compute dense correspondence between human body shapes. They use a \emph{convolutional neural network} (CNN) to extract the feature descriptors of the depth image and establish the correspondences by nearest neighbor search in the space of descriptors. 
	\cite{trappolini2021shape} propose a transformer-based architecture to register 3D point clouds and obtain their correspondence, using a surface attention mechanism that adapts to the point cloud density and the underlying geometry.
	
	\paragraph*{Learnable correspondence with deformation graph} Some methods integrate the learned correspondence and combine it with a deformation graph, to deform the source surface and align with the target surface by optimization. 
    \cite{bozic2020deepdeform} adopt an encoder-decoder architecture to determine the correspondence between two RGB-D frames, by predicting for each source point a heatmap of correspondence probability on the target frame. The correspondence, identified with the highest-probability point on the target, is then used to optimize a deformation graph-based transformation to align the surfaces. They also use a semi-supervised process to collect a large amount of data for training and testing. 
    Concurrently, \cite{li2020learning} propose an end-to-end method to perform non-rigid RGB-D tracking. They train a fully convolutional network to extract feature maps for both the source and the target frames, which are used to define a feature term in an optimization problem for a deformation graph-based transformation that aligns the frames. They solve the optimization using Gauss-Newton, and train an encoder-decoder network to produce PCG preconditioners that accelerate the solver's convergence.
    Further, \cite{bovzivc2020neural} propose a dense correspondence prediction module for the source RGB image and the target RGB image. Through this module, they obtain the dense correspondences and weights to measure the accuracy of these correspondences. Under these correspondences, they use the deformation graph with a differentiable solver to align the input point clouds, which are obtained by depth images. They can obtain better correspondences and reliable weights by end-to-end training. 

	\paragraph*{Learnable displacement field} 
	In~\cite{shimada2019dispvoxnets}, two input point clouds are first converted to voxel grid representations, and then aligned using a voxel displacement field predicted by a network pipeline. The pipeline consists of a displacement estimation stage that predicts large global deformation and a refinement stage that recovers small displacements, each implemented with a UNet-style 3D convolutional network. The model is trained on the FLAME~\cite{li2017learning}, DFAUST~\cite{bogo2017dfaust}, thin plate~\cite{golyanik2018hdm}, and cloth~\cite{bednarik2018learning} datasets. 

	\paragraph*{Registration with parametric representation} Recently, many methods based on parametric shape models have emerged. Some methods use the SMPL model~\cite{loper2015smpl} as the template for the human shape, then optimize SMPL parameters to align with any human body point clouds, including the human body in clothes~\cite{groueix2018coded,wang2020sequential,wang2021locally}. In~\cite{groueix2018coded}, an autoencoder network is used to predict the correspondence between a template shape and an input human shape, which is then used to perform matching between two human shapes with the template mesh as the proxy. 
	The network is trained using synthetic data generating from SMPL with estimated parameters from the SURREAL dataset~\cite{varol2017learning}.
	\cite{wang2020sequential} take a sequence of point clouds as the input, and propose a spatial-temporal attention convolution to directly regress the vertex coordinates of a fitted mesh model in a coarse-to-fine manner. The model is trained using the scanned data sequences and corresponding SMPL models from the DFAUST dataset~\cite{bogo2017dfaust}.
	In~\cite{wang2021locally}, the SMPL model is fitted to the input point cloud of a dressed human. They use the occupancy functions to represent the canonical space, and propose a piecewise transformation field to deform the source point into the canonical space.  The CAPE dataset~\cite{ma2020learning} is used for training.

	\paragraph*{Loss function design}
	Different loss function terms have been adopted in supervised methods to learn the correspondence or deformation. When ground-truth correspondence is available in the training data, some methods introduce function terms that require the predicted correspondence to be close to the ground truth.
	For example, \cite{bozic2020deepdeform} use a probability heatmap to represent point correspondence, and compute the binary cross-entropy loss and negative log-likelihood loss between the predicted and ground-truth heatmaps to penalize their deviation.
	Other methods require some points on the predicted deformed source surface to be close to their ground-truth positions, using a loss function term that penalizes the difference~\cite{groueix2018coded, bovzivc2020neural, wang2020sequential,li2020learning,trappolini2021shape}.
	In~\cite{shimada2019dispvoxnets}, a coarse alignment prediction module is trained with a loss function that penalizes the deviation from the ground-truth displacement, while a refinement module is trained using an alignment loss similar to~\eqref{Eq:align_point_to_point} to refine the registration.

	\subsubsection{Unsupervised methods} 
	Supervised methods require training data with ground-truth correspondence or deformation, which are not easy to obtain. Thus some methods train the model in an unsupervised manner, using data without ground-truth labels.
	In this case, the loss function plays an important role in guiding the training of the model. Below we review some loss functions that have been used in the literature.
	
	\paragraph*{GMM loss}  
	\cite{wang2019non} use a CNN to predict a TPS-based deformation field that aligns two point clouds. To train the model, they use a GMM loss to measure the alignment quality, which has the same form as~\eqref{eq:CPD_align}, with the Gaussian standard deviation being a hyper-parameter.
	
	\paragraph*{Chamfer distance}
	\cite{wang2019coherent} use an MLP to predict a displacement field for aligning two point clouds. They train the model using the Chamfer distance to measure the alignment~\cite{Fan2017}:
	\[
		L_{\text{Chamfer}} = \sum_{\mathbf{x}\in\mathbf{X}} \min_{\mathbf{y}\in\mathbf{Y}}\|\widehat{\mathbf{x}}-\mathbf{y}\|_2^2 + \sum_{\mathbf{y}\in\mathbf{Y}}\min_{\mathbf{x}\in\mathbf{X}}\|\widehat{\mathbf{x}}-\mathbf{y}\|_2^2.
	\]
	For point clouds with outliers and missing data, they also propose the following clipped Chamfer distance loss:
	\[
		L_{\text{C-Chamfer}} = \sum_{\mathbf{x}\in\mathbf{X}} \max(\min_{\mathbf{y}\in\mathbf{Y}}\|\widehat{\mathbf{x}}-\mathbf{y}\|_2^2, c) + \sum_{\mathbf{y}\in\mathbf{Y}}\max(\min_{\mathbf{x}\in\mathbf{X}}\|\widehat{\mathbf{x}}-\mathbf{y}\|_2^2,c),
	\]
	where $c$ is a hyper-parameter. 

    %full to full 
    \paragraph*{Point-to-point reconstruction loss} \cite{zeng2021corrnet3d} propose a feature embedding module to extract the deep features of the input point clouds, a correspondence indicator module to learn the correspondences, and a symmetric deformer module to deform the source point cloud to align with the target point cloud and deform the target point cloud to align with the source point cloud. Then they use a point-to-point reconstruction loss to measure the quality of the deformation and the alignment:
     \[
      L_{\text{pp-recon}} = \|\mathbf{X} - \widehat{\mathbf{Y}}\|_F^2 + \|\mathbf{Y} - \widehat{\mathbf{X}}\|_F^2,
     \]
     where $\widehat{\mathbf{X}}$ and $\widehat{\mathbf{Y}}$ are the deformed source point cloud and the target point cloud, respectively. 
     %In their setting, the number of points in the two input point clouds is the same and a one-to-one correspondence can be established by reordering the points. 
     This point-to-point reconstruction loss is often easier to train than the Chamfer distance.

     \paragraph*{Earth mover's distance}
     \cite{Wang20193DN} use an encoder-decoder network to learn a displacement field to align a source mesh with a target mesh or point cloud. They use the earth mover's distance~\cite{Fan2017} as part of their alignment measure between the source and target:
     \[
        L_{\text{EMD}} = \min_{\phi: \mathcal{X} \mapsto  \mathcal{Y}} \sum_{ {\mathbf{x}} \in \mathcal{X} } \| \widehat{\mathbf{x}} - \phi(\widehat{\mathbf{x}}) \|_2,
     \]
     where $\phi: \mathcal{X} \mapsto  \mathcal{Y}$ is a bijection. In~\cite{Wang20193DN}, this term is used in conjunction with the Chamfer distance to measure the alignment error.
     
     %full to full
	\paragraph*{Multi-view projection loss} \cite{feng2021recurrent} propose a multi-view projection loss function for aligning point clouds and achieve better results than the Chamfer distance. 
	The idea is to project the 3D shapes onto multiple 2D planes with different orientations, and measure the differences of the depth and mask images between the source shape and the target shapes as an indication of the alignment error. 
	The main difference compared with Chamfer distance and Earth mover's distance is that it provides more directional information.
	In addition, they propose a recurrent update framework that defines several rigid transformations $\{\mathbf{R}_r\}$ without specific locations to represent the deformation field. To represent the deformation on each source point, they define
	\[
		\widehat{\mathbf{x}}_i = \sum_{r=1}^K {w}_i^r\cdot \mathbf{R}_r,
	\]
	where ${w}_i^r$ is the skinning weight for $\mathbf{x}_i$ with the constraint:
	\[
		\sum_{r=1}^k w_i^r = 1, \quad \forall i = 1, 2, \ldots, m.
	\]  
     Their model is trained on the HumanMotion~\cite{vlasic2008articulated}, TOSCA~\cite{bronstein2008numerical} and SURREAL~\cite{varol2017learning} datasets, and tested on the FaceWareHouse~\cite{cao2014facewarehouse} and DFAUST~\cite{bogo2017dfaust} datasets.
     
     \paragraph*{Regularization}
     Similar to optimization-based methods, it is often insufficient to use just an alignment term for the training loss. Some additional regularization terms may be needed to prevent trivial solutions and to produce natural results. For example, in~\cite{Wang20193DN}, the training loss includes a symmetry term to promote symmetry for man-made objects, a Laplacian term to preserve local geometric detail, and a local permutation invariant term to prevent self-intersections. In~\cite{feng2021recurrent}, they include an as-rigid-as-possible term to preserve local shapes, and regularization terms for the translation and the skinning weights, respectively.
     Some methods include other regularization terms, such as the regularization of the learnable parameters, to make the training easier and to prevent overfitting~\cite{feng2021recurrent,BozicPZTDN21}.

    \subsection{Applications and Systems}

    With the development of the RGB-D data acquisition equipment and technologies, there has been a growing number of works on dynamic reconstruction and tracking based on non-rigid registration. 

    \subsubsection{Template-less methods}  DynamicFusion~\cite{newcombe2015dynamicfusion} adopts the TSDF to represent the canonical model and support the fusion operation over multiple frames. Without any prior, they simultaneously fuse geometry information and estimate the deformation field. 
    The system can run in real-time and obtain relatively complete reconstruction results with details. However, it is not robust to fast movements and occluded areas. 
    Further, VolumeDeform~\cite{innmann2016volumedeform} uses a volumetric representation to define the deformation graph and encode the scene's geometry, the accuracy of correspondences is improved by introducing the SIFT feature. 
    \cite{dou2016fusion4d} use multi-view capture information and propose a key volume strategy to handle failures in tracking. To improve the accuracy of the correspondences obtained, they use a learning-based method, GPC~\cite{wang2016global}, and extend it to make use of both the RGB image and the depth image to obtain more robust results within an acceptable time.
    \cite{guo2017real} use a single RGB-D camera to reconstruct geometry, albedo, and motion in real-time. 
    Observing that the appearance is beneficial to the accuracy of the reconstruction, they combine the shading information to estimate the motion more accurately and also obtain a lighting estimation. 
    
    \cite{wang2017templateless} split the input RGB-D sequences into several parts. In each part, they perform the local non-rigid bundle adjustment to register all scans. Then they merge all parts to eliminate the drift problem by global optimization. A final 3D model can be generated by using Poisson surface reconstruction or volumetric fusion.  
    \cite{slavcheva2018sobolevfusion} represent each RGB-D frame as a TSDF and deform it to align with the canonical TSDF. They use a Sobolev gradient flow to achieve faster computation and better detail. They additionally compare different regularization strategies, and utilize signature matching to handle rapid motions~\cite{slavcheva2020variational}.  
    \cite{xu2015deformable} propose a two-stage method. First, they perform CPD to register the previous frame and the current frame. Second, they use a mean shape and a deformation on a low-dimensional space to fuse the information of the new frame. 
    \cite{xu2018online} uses non-rigid registration to reconstruct nearly rigid objects. Due to the noise and distortion caused by camera acquisition, they use a deformed graph with few graph nodes as a deformation representation to reduce the error and improve the reconstruction quality.

    \subsubsection{Methods based on priors} 
    Considering the prior information of the human body, \cite{bogo2015detailed} use the human parametric model to reconstruct people in motion from monocular RGB-D sequences. Besides estimating the low-dimensional representation, they define a variable-detail shape model and use a low polygon count mesh with a high-resolution texture map and displacement map to represent the details. BodyFusion~\cite{yu2017bodyfusion} embeds the body skeleton model in the human model to enhance the robustness to large motions. They propose a binding term to establish the relationship between the skeleton model and the deformation graph. They improve the performance and robustness of the non-rigid reconstruction and tracking. However, a skeleton model that is too sparse is not robust to handle fast motion or incomplete surfaces. \cite{yu2018doublefusion} propose a double-layer model including a body template model and a dense geometry model to address this problem. It makes the system more robust to the fast movement.
    \cite{zheng2018hybridfusion} add sparse inertial measurement units (IMUs) to improve the tracking performance.
    UnstructuredFusion~\cite{xu2019unstructuredfusion} uses three deep cameras without any careful pre-calibration  to cover the whole human body and achieve real-time high-quality human capture. In addition, they propose a skeleton warping based non-rigid tracking scheme to calibrate the camera and track at the same time. 
    \cite{thomas2016augmented} use the first RGB-D frame with facial landmarks and neutral expression to obtain an initial head model, then they use it to track changes in facial expression. \cite{yu2018headfusion} combine 3D Morphable Model (3DMM)~\cite{blanz1999morphable} and a head reconstruction framework without prior for accurate head tracking. \cite{tagliasacchi2015robust,  tkach2016sphere,tkach2017online} use the prior of a hand to model and track dynamic hand motion in real-time. 
    In~\cite{weise2011realtime} and~\cite{Bouaziz2013online}, parametric face models are used for real-time face tracking on commodity RGB-D sensors, allowing the user to control the facial expressions of a digital avatar.
    Besides these class-specific priors, some methods establish the template model by the first frame or use a pre-scanned model~\cite{tong2012scanning, xu2017flycap, li2020robust}.  

    \subsubsection{Learning-based methods}  
    \cite{dou2017motion2fusion, bozic2020deepdeform} use the learning-based technique to obtain a more accurate correspondence. \cite{li2020learning} improve the registration quality by using the learned feature as an alignment term. \cite{bovzivc2020neural} combine correspondence learning and differentiable solving, and form an end-to-end network.
    \cite{BozicPZTDN21} propose an implicit neural deformation graph with learnable node positions and influence radii, which can be applied to a sequence of depth camera observations to perform globally-consistent registration and reconstruct a non-rigidly moving object.
    \cite{su2020robustfusion} adopts an occupancy network, a pose and shape estimation and semantics networks are used to improve the robustness of the reconstruction. \cite{yu2021function4d} combine a temporal volumetric fusion technique and deep implicit functions. They first use the multi-view RGB-D input to perform dynamic fusion and obtain a volumetric fusion. They then re-render multi-view RGB-D images from this volume. Finally, they obtain the detailed model through a deep implicit surface reconstruction step. \cite{zhang2019interactionfusion} perform real-time reconstruction of a hand and deformable objects held by the hand. They first segment the hand, the deformable object and the background, and then use an LSTM-based network to predict the hand pose according to temporal information. Finally, they utilize the predicted pose to perform the frame-by-frame optimized tracking. 

    \subsubsection{Methods combining optimization and learning}
    Some methods combine optimization-based and learning-based approaches to improve non-rigid reconstruction performance. \cite{bozic2020deepdeform} first use a learning-based approach to obtain sparse correspondences represented by probability heatmaps.
    The correspondences are then utilized to formulate an optimization problem for non-rigid registration.  \cite{zhang2019interactionfusion, su2020robustfusion} first use a learning-based method to predict the parameters of a prior-based shape model to achieve an initial coarse alignment. Afterwards, the registration is further improved by optimizing a detailed deformation field. 
    \cite{bovzivc2020neural} first use a CNN to predict dense correspondences between the two surfaces; the correspondences are then used as constraints to perform an optimization-based registration.  
    \cite{li2020learning} propose an optimization formulation with an alignment term that is based on learned features and helps the Gauss-Newton solver to avoid local minima; additionally, a learned preconditioner is utilized to improve the convergence of the PCG solver used by the Gauss-Newton method.
    
    \subsection{Summary}
    Tables~\ref{tab:extrinsic-optimization-methods} and \ref{tab:extrinsic-learning-methods} summarize the optimization-based and learning-based extrinsic methods discussed in this section, respectively.
    Papers that involve both optimization-based and learning-based methods are listed in both tables according to their relevant components.
    
	\begin{table*}[!t]
	\caption{A summary of optimization-based extrinsic methods.}
	\label{tab:extrinsic-optimization-methods}
	\setlength{\tabcolsep}{1.2pt}
	\centering
	\tablefont{
		\begin{tabular}{ >{\centering}p{92.8pt} | c c c c |c c c c c | c c | c c c c c c c c}
			\Xhline{1pt}
			\multicolumn{1}{ c |}{~}
			&\multicolumn{4}{ c |}{Alignment}
			&\multicolumn{5}{ c |}{Regularization}
            &\multicolumn{2}{ c |}{Robustness}
			&\multicolumn{8}{ c }{Deformation field} \\\cline{2-20}
			\multicolumn{1}{ c |}{}
			&\multicolumn{1}{c }{\tabincell{c}{Point-to-\headingbreak{}point}}
            &\multicolumn{1}{ c }{\tabincell{c}{Point-to-\headingbreak{}plane}}
            &\multicolumn{1}{ c }{\tabincell{c}{SDF/\headingbreak{}TSDF}}
            &\multicolumn{1}{ c | }{\tabincell{c}{Probabi-\headingbreak{}listic}}
			&\multicolumn{1}{ c }{\tabincell{c}{Smooth}}
            &\multicolumn{1}{ c }{\tabincell{c}{Position}}
            &\multicolumn{1}{ c }{\tabincell{c}{Near-\headingbreak{}isometry}}
            &\multicolumn{1}{ c }{\tabincell{c}{Non-\headingbreak{}isometry}}
            &\multicolumn{1}{ c| }{Other}
            &\multicolumn{1}{c }{\tabincell{c}{Adapt.\headingbreak{}Weight}}
            &\multicolumn{1}{ c| }{\tabincell{c}{Robust\headingbreak{}Norm}}
            &\multicolumn{1}{ c }{\tabincell{c}{Pointwise\headingbreak{}Pos. Var.}}
            &\multicolumn{1}{ c }{\tabincell{c}{Pointwise\headingbreak{}Aff. Trans.}}
            &\multicolumn{1}{ c }{\tabincell{c}{Deform.\headingbreak{}Graph}}
            &\multicolumn{1}{ c }{\tabincell{c}{Patch}}
            &\multicolumn{1}{ c }{\tabincell{c}{RKHS}}
            &\multicolumn{1}{ c }{\tabincell{c}{Spline\headingbreak{}Func.}}
            &\multicolumn{1}{ c }{\tabincell{c}{Grid}}
            &\multicolumn{1}{ c }{\tabincell{c}{Prior}} \\\hline
\cite{Rueckert1999}  &  &  &  &  & \checkmark &  &  &  & \checkmark &  &  &  &  &  &  &  &  \checkmark  &  & \\
\cite{allen2002articulated}  &  &  &  &  &  &  \checkmark  &  &  &  &  &  &  &  &  &  &  &  &  &  \checkmark \\
\cite{allen2003space}  &  \checkmark  &  &  &  &  \checkmark  &  \checkmark  &  &  &  &  &  &  &  \checkmark  &  &  &  &  &  & \\
\cite{chui2003new,rouhani2012non}  &  \checkmark  &  &  &  &  \checkmark  &  &  &  &  &  &  &  &  &  &  &  &  \checkmark  &  & \\
\cite{jian2005robust,huang2019automatic}  &  &  &  &  \checkmark  &  \checkmark  &  &  &  &  &  &  &  &  &  &  &  &  \checkmark  &  & \\
%\cite{jian2005robust}  &  &  &  &  \checkmark  &  \checkmark  &  &  &  &  &  &  &  &  &  &  &  &  \checkmark  &  & \\
\cite{pauly2005example,fan2018dense}  &  \checkmark  &  &  &  &  \checkmark  &  &  &  &  &  &  &  \checkmark  &  &  &  &  &  &  & \\
%\cite{pauly2005example}  &  \checkmark  &  &  &  &  \checkmark  &  &  &  &  &  &  &  \checkmark  &  &  &  &  &  &  & \\
\cite{amberg2007optimal}  &  \checkmark  &  &  &  &  \checkmark  &  \checkmark  &  &  &  &  \checkmark  &  &  &  \checkmark  &  &  &  &  &  & \\
%\tabincell{c}{\cite{myronenko2007non,ma2013robust,ma2015non} \\ \cite{myronenko2010point,ma2015robust} \\ \cite{ma2017non,hirose2020acceleration}}  
\begin{minipage}{92pt}
\centering
\cite{myronenko2007non,myronenko2010point,ma2013robust,ma2015robust,ma2015non,ma2017non,hirose2020acceleration}
\end{minipage}
&  &  &  &  \checkmark  &  \checkmark  &  &  &  &  &  &  &  &  &  &  &  \checkmark  &  &  & \\
%\cite{myronenko2007non}  &  &  &  &  \checkmark  &  \checkmark  &  &  &  &  &  &  &  &  &  &  &  \checkmark  &  &  & \\
\cite{wand2007reconstruction}  &  \checkmark  &  &  &  &  \checkmark  &  &  \checkmark  &  &  &  &  &  \checkmark  &  &  &  &  &  &  & \\
%\cite{huang2008nonrigid}  &  \checkmark  &  \checkmark  &  &  &  &  &  \checkmark  &  &  &  &  &  &  &  &  \checkmark  &  &  &  & \\
\cite{li2008global,lin2016color}  &  \checkmark  &  &  &  &  \checkmark  &  &  \checkmark  &  &  &  \checkmark  &  &  &  &  \checkmark  &  &  &  &  & \\
%\cite{li2008global}  &  \checkmark  &  &  &  &  \checkmark  &  &  \checkmark  &  &  &  \checkmark  &  &  &  &  \checkmark  &  &  &  &  & \\
\cite{pekelny2008articulated,gall2009motion,jiang2017consistent}  &  \checkmark  &  &  &  &  \checkmark  &  &  &  &  &  &  &  &  &  &  &  &  &  &  \checkmark \\
%\cite{pekelny2008articulated}  &  \checkmark  &  &  &  &  \checkmark  &  &  &  &  &  &  &  &  &  &  &  &  &  &  \checkmark \\
\cite{li2009robust}  &  \checkmark  &  \checkmark  &  &  &  \checkmark  &  &  \checkmark  &  &  &  &  &  &  &  \checkmark  &  &  &  &  & \\
\cite{liao2009modeling}  &  \checkmark  &  &  &  &  &  &  &  \checkmark  &  &  &  &  \checkmark  &  &  &  &  &  &  & \\
\cite{wand2009efficient}  &  \checkmark  &  \checkmark  &  &  &  \checkmark  &  &  \checkmark  &  \checkmark  &  &  &  &  \checkmark  &  &  &  &  &  &  & \\
\cite{cagniart2010free}  &  \checkmark  &  &  &  &  &  &  \checkmark  &  &  &  &  &  &  &  &  \checkmark  &  &  &  & \\
%\cite{myronenko2010point}  &  &  &  &  \checkmark  &  \checkmark  &  &  &  &  &  &  &  &  &  &  &  \checkmark  &  &  & \\
\cite{yeh2010template}  &  &  &  &  &  \checkmark  &  \checkmark  &  &  &  &  &  &  \checkmark  &  &  &  &  &  &  & \\
\cite{chang2011global}  &  \checkmark  &  \checkmark  &  &  &  &  &  &  &  \checkmark  &  \checkmark  &  &  &  &  &  &  &  &  &  \checkmark \\
\cite{fujiwara2011locally}  &  &  &  \checkmark  &  &  &  &  &  &  &  &  &  &  &  &  &  &  &  \checkmark  & \\
\cite{huang2011global}  &  \checkmark  &  &  &  &  &  &  \checkmark  &  &  &  &  &  \checkmark  &  &  &  &  &  &  & \\
\cite{Jian2011robust}  &  &  &  &  \checkmark  &  &  &  &  &  &  &  &  &  &  &  &  &  \checkmark  &  & \\
\cite{weise2011realtime}  &  &  \checkmark  &  &  \checkmark  &  &  &  &  &  \checkmark  &  &  &  \checkmark  &  &  &  &  &  &  &  \checkmark \\
\cite{hontani2012robust}  &  \checkmark  &  &  &  &  \checkmark  &  &  &  &  &  &  \checkmark  &  &  \checkmark  &  &  &  &  &  & \\
%\cite{rouhani2012non}  &  \checkmark  &  &  &  &  \checkmark  &  &  &  &  &  &  &  &  &  &  &  &  \checkmark  &  & \\
\cite{tong2012scanning}  &  &  &  &  &  \checkmark  &  \checkmark  &  \checkmark  &  &  &  &  &  &  &  \checkmark  &  &  &  &  & \\
\cite{Bouaziz2013online}  &  \checkmark  &  &  &  &  \checkmark  &  &  &  &  &  &  \checkmark  &  &  &  &  &  &  &  &  \checkmark \\
%\cite{ma2013robust}  &  &  &  &  \checkmark  &  \checkmark  &  &  &  &  &  &  &  &  &  &  &  \checkmark  &  &  & \\
\cite{yamazaki2013non}  &  \checkmark  &  \checkmark  &  &  &  &  &  &  \checkmark  &  &  &  &  \checkmark  &  &  &  &  &  &  & \\
\cite{bonarrigo2014deformable}  &  \checkmark  &  &  &  &  \checkmark  &  &  &  &  &  &  &  &  &  \checkmark  &  &  &  &  & \\
\cite{cao2014facewarehouse}  &  \checkmark  &  &  &  &  \checkmark  &  \checkmark  &  &  &  &  &  &  &  &  &  &  &  &  &  \checkmark \\
\cite{ge2014non}  &  &  &  &  \checkmark  &  \checkmark  &  &  &  & \checkmark  &  &  &  &  &  &  & \checkmark &  &  & \\
\cite{liang20143d}  &   &  &  &  &  &  \checkmark  &  &  & \checkmark &  &  &  &  &  &  \checkmark  &  &  &  & \\
\cite{xu2014nonrigid}  &  \checkmark  &  &  &  &  \checkmark  &  &  &  &  &  &  &  &  &  &  \checkmark  &  &  &  & \\
\cite{yoshiyasu2014conformal}  &  \checkmark  &  &  &  &  \checkmark  &  &  &  \checkmark  &  &  &  &  &  \checkmark  &  &  &  &  &  & \\
\cite{Achenbach2015Accurate}  &  \checkmark  &  \checkmark  &  &  &  &  &  &  \checkmark  &  &  &  \checkmark  &  \checkmark  &  &  &  &  &  &  & \\
\cite{bogo2015detailed}  &  \checkmark  &  \checkmark  &  &  &  \checkmark  &  &  \checkmark  &  &  &  &  &  &  &  &  &  &  &  &  \checkmark \\
\cite{cao2015two}  &  \checkmark  &  &  &  &  \checkmark  &  &  \checkmark  &  &  &  &  &  &  &  \checkmark  &  &  &  &  & \\
\cite{ge2015non}  &  &  &  &  \checkmark  &  \checkmark  &  &  &  \checkmark  &  &  &  &  &  &  &  &  \checkmark  &  &  & \\
\cite{guo2015robust}  &  \checkmark  &  \checkmark  &  &  &  \checkmark  &  &  \checkmark  &  &  &  &  \checkmark  &  &  &  \checkmark  &  &  &  &  & \\
%\cite{ma2015non}  &  &  &  &  \checkmark  &  \checkmark  &  &  &  &  &  &  &  &  &  &  &  \checkmark  &  &  & \\
\cite{newcombe2015dynamicfusion}  &  &  \checkmark  &  \checkmark  &  &  \checkmark  &  &  &  &  &  &  \checkmark  &  &  &  &  &  &  &  \checkmark  & \\
\cite{Sahillioglu2015ashape}  & \checkmark &  &  &  &  &  & \checkmark &  & \checkmark  &  &  & \checkmark &  &  &  &  &  &  &  \\
\cite{tagliasacchi2015robust,yu2018headfusion}  &  \checkmark  &  &  &  &  &  &  &  &  \checkmark  &  &  &  &  &  &  &  &  &  &  \checkmark \\
\cite{xu2015deformable} &  &  &  & \checkmark &  &  &  &  & \checkmark &  &  &  \checkmark &  &  &  &  &  &  & \\
\cite{yang2015sparse}  &  \checkmark  &  &  &  &  \checkmark  &  &  &  &  &  &  \checkmark  &  &  \checkmark  &  &  &  &  &  & \\
\cite{zhang2015efficient}  &  &  &  \checkmark  &  &  &  &  &  &  &  \checkmark  &  &  &  &  &  &  &  &  \checkmark  & \\
\cite{dou2016fusion4d}  &  &  \checkmark  &  &  &  \checkmark  &  &  \checkmark  &  &  &  &  \checkmark  &  &  &  \checkmark  &  &  &  &  & \\
\cite{innmann2016volumedeform}  &  \checkmark  &  \checkmark  &  &  &  &  &  &  &  &  \checkmark  &  &  &  &  &  &  &  &  \checkmark  & \\
\cite{santa2016face,fan2016convex}  &  \checkmark  &  &  &  &  &  &  &  &  &  &  &  &  &  &  &  &  \checkmark  &  & \\
%\cite{santa2016face}  &  \checkmark  &  &  &  &  &  &  &  &  &  &  &  &  &  &  &  &  \checkmark  &  & \\
\cite{tkach2016sphere}  &  \checkmark  &  &  &  &  &  &  \checkmark  &  &  &  &  &  &  &  &  &  &  &  &  \checkmark \\
\cite{thomas2016augmented}  &   &  \checkmark  &  &  &  &  \checkmark  &  &  &  &  &  &  &  &  &  &  &  &  &  \checkmark \\
\cite{bogo2017dfaust}  &  \checkmark  &  &  &  &  \checkmark  &  &  &  &  &  &  &  &  &  &  &  &  &  &  \checkmark \\
\cite{dou2017motion2fusion}  &  \checkmark  &  \checkmark  &  \checkmark  &  &  \checkmark  &  &  &  &  &  &  &  &  &  \checkmark  &  &  &  &  & \\
\cite{guo2017global}  &  \checkmark  &  &  &  &  \checkmark  &  &  \checkmark  &  &  &  &  \checkmark  &  & \checkmark  &  &  &  &  &  & \\
\cite{guo2017real}  &  &  \checkmark  &  &  &  \checkmark  &  &  \checkmark  &  &  &  &  &  &  &  &  &  &  &  \checkmark  & \\
%\cite{jiang2017consistent}  &  \checkmark  &  &  &  &  &  \checkmark  &  &  \checkmark  &  &  &  &  \checkmark  &  &  &  &  &  &  & \\
\cite{li2017robust}  &  \checkmark  &  \checkmark  &  &  &  \checkmark  &  &  \checkmark  &  &  \checkmark  &  &  &  &  &  \checkmark  &  &  &  &  & \\
%\cite{ma2017non}  &  &  &  &  \checkmark  &  \checkmark  &  &  &  &  &  &  &  &  &  &  &  \checkmark  &  &  & \\
\cite{slavcheva2017killingfusion,slavcheva2018sobolevfusion,slavcheva2020variational}  &  &  &  \checkmark  &  &  \checkmark  &  &  &  &  &  &  &  &  &  &  &  &  &  \checkmark  & \\
\cite{tkach2017online}  &  \checkmark  &  &  &  &  \checkmark  &  &  &  &  \checkmark  &  &  &  &  &  &  &  &  &  &  \checkmark \\
\cite{wang2017templateless, xu2017flycap}  &  \checkmark  &  \checkmark  &  &  &  \checkmark  &  &  \checkmark  &  &  &  &  &  &  &  &  \checkmark  &  &  &  & \\
\cite{yu2017bodyfusion}  &  &  \checkmark  &  &  &  \checkmark  &  &  &  &  \checkmark  &  &  \checkmark  &  &  &  \checkmark  &  &  &  &  &  \checkmark \\
%\cite{xu2017flycap}  &  \checkmark  &  \checkmark  &  &  &  \checkmark  &  &  \checkmark  &  &  &  &  &  &  &  &  \checkmark  &  &  &  & \\
\cite{kozlov2018patch}  &  \checkmark  &  &  &  &  \checkmark  &  &  &  &  &  \checkmark  &  &  &  &  &  \checkmark  &  &  &  & \\
\cite{li2018articulatedfusion}  &  &  \checkmark  &  &  &  \checkmark  &  &  &  &  &  &  &  &  &  &  \checkmark  &  &  &  & \\
%\cite{slavcheva2018sobolevfusion}  &  &  &  \checkmark  &  &  \checkmark  &  &  &  &  &  &  &  &  &  &  &  &  &  \checkmark  & \\
\cite{xu2018online}  &  \checkmark  &  \checkmark  &  &  &  \checkmark  &  &  &  &  &  &  &  &  &  \checkmark  &  &  &  &  & \\
\cite{yu2018doublefusion}  &  &  \checkmark  &  &  &  \checkmark  &   &  &  &  &  &  \checkmark  &  &  &  \checkmark  &  &  &  &  &  \checkmark \\
%\cite{yu2018headfusion}  &  \checkmark  &  &  &  &  &  &  &  &  \checkmark  &  &  &  &  &  &  &  &  &  &  \checkmark \\
\cite{zheng2018hybridfusion}  &  \checkmark  &  &  &  &  &  &  \checkmark  &  &  \checkmark  &  &  &  &  &  \checkmark  &  &  &  &  &  \checkmark \\
%\cite{dyke2019non}  &  \checkmark  &  \checkmark  &  &  &  &  &  \checkmark  &  &  &  &  &  &  &  &  &  &  &  & \\
\cite{jiang2019huber}  &  &  &  &  &  \checkmark  &  \checkmark  &  &  \checkmark  &  &  &  &  &  \checkmark  &  &  &  &  &  & \\
\cite{li2019robust,yang2019global}  &  \checkmark  &  &  &  &  \checkmark  &  &  \checkmark  &  &  &  &  \checkmark  &  &  \checkmark  &  &  &  &  &  & \\
\cite{Wu2019global}  &  \checkmark  & \checkmark &  &  &  \checkmark  &   &  \checkmark  &  \checkmark  &  &  &  \checkmark  &  &  \checkmark  &  &  &  &  &  & \\
%\cite{yang2019global}  &  \checkmark  &  &  &  &  \checkmark  &  &  \checkmark  &  &  &  &  \checkmark  &  &  \checkmark  &  &  &  &  &  & \\
\cite{yu2019simulcap}  &  &  \checkmark  &  &  &  &  &  &  &  \checkmark  &  &  &  &  &  &  &  &  &  &  \checkmark \\
\cite{zhang2019interactionfusion}  &  &  \checkmark  &  &  &  \checkmark  &  &  &  &  \checkmark  &  &  &  &  &  &  &  &  &  \checkmark  & \\
\cite{bovzivc2020neural}  & \checkmark &  &  &  & \checkmark &  &  &  & \checkmark  & \checkmark &  &  &  & \checkmark &  &  &  &  &  \\
\cite{bozic2020deepdeform}  & \checkmark & \checkmark &  &  &  &  & \checkmark &  &  &  &  &  &  & \checkmark &  &  &  &  &  \\
%\cite{hirose2020acceleration}  &  &  &  &  \checkmark  &  \checkmark  &  &  &  &  &  &  &  &  &  &  &  \checkmark  &  &  & \\
\cite{li2020learning}  & \checkmark &  &  &  &  &  & \checkmark &  & \checkmark  &  &  &  &  & \checkmark &  &  &  &  &  \\
\cite{li2020topology}  &   &  \checkmark  &  &  &  &  \checkmark  &  \checkmark  &  &  &  &  &  &  &  \checkmark  &  &  &  &  & \\
\cite{li2020robust2}  &  &  \checkmark  &  \checkmark  &  &  \checkmark  &  &  &  &  \checkmark  &  &  &  &  &  \checkmark  &  &  &  &  & \\
\cite{li2020robust}  &  &  \checkmark  &  &  &  \checkmark  &  &  \checkmark  &  &  &  &  &  &  &  \checkmark  &  &  &  &  & \\
%\cite{slavcheva2020variational}  &  &  &  &  &  &  &  &  &  &  &  &  &  &  &  &  &  &  & \\
\cite{su2020robustfusion}  &  &  \checkmark  &  \checkmark  &  &  &   &  &  &  \checkmark  &  &  \checkmark  &  &  &  &  &  &  &  &  \checkmark \\
\cite{xu2019unstructuredfusion}  &  &  \checkmark  &    &  &  &   &  \checkmark  &  &  \checkmark  &  &  \checkmark  &  &  &  &  \checkmark  &  &  &  & \\
\cite{yao2020quasi}  &  \checkmark  &  &  &  &  \checkmark  &  &  \checkmark  &  &  &  &  \checkmark  &  &  &  \checkmark  &  &  &  &  & \\
\cite{yu2021function4d}  &  &  \checkmark  &  &  &  &  &  \checkmark  &  &  &  &  &  &  &  \checkmark  &  &  &  &  & \\
\cite{zampogiannis2019topology}  &   &  \checkmark  &  &  &  \checkmark  &  \checkmark  &  &  &  &  &  \checkmark  &  &  &  \checkmark  &  &  &  &  & \\
			\Xhline{1pt}
		\end{tabular}
	}
\end{table*}
 
\begin{table*}[!t]
	\caption{A summary of learning-based extrinsic methods.}
	\label{tab:extrinsic-learning-methods}
	\setlength{\tabcolsep}{2pt}
	\centering
	\tablefont{
		\begin{tabular}{ c | c c | c c c | c c c c c c c}
			\Xhline{1pt}
			\multicolumn{1}{ c | }{\multirow{2}{*}{~}}
            &\multicolumn{2}{ c |}{Objective}
            &\multicolumn{3}{ c |}{Training type}
			%&\multicolumn{4}{| c }{Deformation field}
            &\multicolumn{6}{ c }{Loss}\\\cline{2-12}
           % &\multicolumn{1}{| c }{\multirow{2}{*}{Probalisic model}}
           % &\multicolumn{1}{| c }{\multirow{2}{*}{Parametric models}}\\\cline{2-16}
			%\midrule
			\multicolumn{1}{ c| }{}
			&\multicolumn{1}{c }{\tabincell{c}{Learnable \headingbreak{}correspondences}}
            &\multicolumn{1}{ c| }{\tabincell{c}{Learnable \headingbreak{}deformation}}
            &\multicolumn{1}{ c }{Supervised}
            &\multicolumn{1}{ c }{\tabincell{c}{Semi-\headingbreak{}supervised}}
            &\multicolumn{1}{ c| }{Unsupervised}
            &\multicolumn{1}{ c }{\tabincell{c}{Ground-truth\headingbreak{} correspondences}}
            &\multicolumn{1}{ c }{\tabincell{c}{Ground-truth\headingbreak{} positions}}
            &\multicolumn{1}{ c }{Alignment}
           % &\multicolumn{1}{ c }{Smoothness}
            &\multicolumn{1}{ c }{\tabincell{c}{Near-isometry}}
            &\multicolumn{1}{ c }{\tabincell{c}{Non-isometry}}
            &\multicolumn{1}{ c }{Other}\\\hline
\cite{wei2016dense}  &  \checkmark  &  &  \checkmark  &  &  &  \checkmark  &  &  &  &  & \\
\cite{groueix2018coded}  &  \checkmark  &  \checkmark  &  \checkmark  &  &  \checkmark  & \checkmark &  \checkmark  &  &  \checkmark  &  \checkmark  & \\
\cite{shimada2019dispvoxnets}  &  &  \checkmark  &  \checkmark  &  &  &  &  \checkmark  &  \checkmark  &  &  & \\
\cite{wang2019non}  &  \checkmark  &  &  &  &  \checkmark  &  &  &  &  &  &  \\
\cite{Wang20193DN}  &  &  \checkmark  &  &  &  \checkmark  &  &  &  \checkmark  &  &    &  \\
\cite{wang2019coherent}  &  &  \checkmark  &  &  &  \checkmark  &  &  &  \checkmark  &  &  & \\
\cite{zhang2019interactionfusion,su2020robustfusion}  &  &  \checkmark  &  \checkmark  &  &  &  &  &   &  &  & \checkmark \\
\cite{bovzivc2020neural}  &  \checkmark  &  &  \checkmark  &  &  &  &  \checkmark  &  &  &  & \\
\cite{bozic2020deepdeform}  &  \checkmark  &  & \checkmark &  &  &  \checkmark  &  \checkmark  &  &   &  &  \checkmark \\
\cite{li2020learning}  &  &  \checkmark  &  &  \checkmark  &  &  &  \checkmark  &   &    &  & \\
\cite{wang2020sequential}  &  \checkmark  & \checkmark &  \checkmark  &  &  &  &  \checkmark  &   &  \checkmark  &  \checkmark  & \\
\cite{BozicPZTDN21}  &  &  \checkmark  &  &  &  \checkmark  &  &  &  \checkmark  &  \checkmark  &  &  \checkmark \\
\cite{feng2021recurrent}  &  &  \checkmark  &  &  &  \checkmark  &  &  &  \checkmark  &  \checkmark  &  &  \checkmark \\
\cite{trappolini2021shape}  &  \checkmark  & \checkmark &  \checkmark  &  &  &  &  \checkmark  &  &  &  & \\
\cite{zeng2021corrnet3d}  &  \checkmark  &  \checkmark  &  &  &  \checkmark  &  &  &  \checkmark  &  &  &  \checkmark \\
\Xhline{1pt}
\end{tabular}
    }
\end{table*}

	\section{Intrinsic Methods}
	\label{sec:intrinsic}
	Unlike extrinsic methods that align surfaces in the original $\mathbb{R}^3$ coordinate space, by the most general definition, intrinsic methods transform a set of surfaces into an alternative representation in another coordinate space---or domain---in which they are aligned. 
	With this in mind, intrinsic methods can be categorized by the characteristics of the domain they are embedded in, \ie (1) parametric; (2) minimum-distortion; and (3) spectral/functional. Parametrization-based methods compute low-dimensional embeddings, typically in $\mathbb{R}^2$. Minimum-distortion refers to methods that minimize some form of intrinsic correspondence measure, the dimensions of which depend on the number of keypoints between surfaces. Spectral/functional methods embed surfaces in a high-dimensional space where the correspondence problem may be solved using established mathematical tools. Furthermore, by avoiding working directly on the graph, the influence of mesh connectivity is lessened greatly.

    Much of the related literature only establishes correspondence. Few methods explicitly deform the source shape to the target shape. However, it is possible to incorporate correspondences as hard/soft constraints in a non-rigid registration framework to help align surfaces.
    
	A variety of techniques can be extended to the problem of non-rigid registration. A key deficiency of many approaches is their sensitivity to partial data---a common scenario in registration. This vulnerability is caused by the use of intrinsic measures that are sensitive to geometric and topological noise. Despite this challenge, a variety of methods have been proposed in the literature to address this problem, \eg minimum-distortion~\cite{sahillioglu2018genetic}; spectral reconstruction~\cite{cosmo2019isospectralization,hamidian2020surface}; learning-based~\cite{eisenberger2021neuromorph}; hybrid (intrinsic initialization then extrinsic refinement) methods~\cite{eisenberger2020smooth,marin2020farm}.
	
	\cite{sahillioglu2020recent} recently reviewed a broad range of literature that explores the problem of shape correspondence---with a predominant emphasis on intrinsic methods. For this survey, a keen focus is made on literature that seeks to develop robust partial matching techniques. These works contribute towards solving the partial matching problem, which is especially relevant for applications, such as reconstruction, where data is missing or incomplete. One may consider these methods to address similar problems to registration methods through the use of intrinsic data.
	It is also possible to combine intrinsic and extrinsic techniques to form robust registration pipelines. After a systematic review of material relating to parameterization-based methods, it has been deemed appropriate to limit the scope of this survey to exclude such methods. The reader is directed to~\cite{sheffer2006mesh,hormann2007mesh} for a broad overview of parameterization techniques and discussions on early works on cross-parameterization.
	
	\subsection{Optimization-Based methods}
	The general objective of most intrinsic matching algorithms is to compute a bijective linear map $T : \mathcal{X} \mapsto \mathcal{Y}$ that maps corresponding points between two shapes ($\mathcal{X}$ and $\mathcal{Y}$). For convenience of notation, in this section, let all shapes comprise of $n$ points (\ie $m=n$).
    
    \paragraph*{Point-to-point mapping}
    Many intrinsic methods seek to approximate $T$ by computing a point-to-point mapping of a permutation matrix $\mathbf{P} \in \mathcal{P}_n$ where $\mathcal{P}_n = \{ \mathbf{P} \in \left \{ 0,1 \right\}^{n \times n} \mid \mathbf{P}^{\top}\mathbf{P}=\mathbf{I} \}$ is the set of permutations. The problem of finding $\mathbf{P}$ is cast as an optimization problem by a given energy function $E$:
    \begin{equation}\label{eq:bijective-perms}
    \argmin_{\mathbf{P} \in \mathcal{P}_n} E(\mathbf{P}).
    \end{equation}
    In this form, $\mathbf{P}$ is an orthogonal matrix (\st $\mathbf{P}^{\top} \mathbf{P} = \mathbf{I}$), which is naturally bijective. The permutation set is quite large $\left|\mathcal{P}_n\right|=n!$. It is therefore impractical to evaluate all permutations, so further heuristics must be employed.
	
	Worse yet, in the case of partial shape matching, the optimal solution is no longer a bijection but either injective or surjective. The solution space is therefore no longer contained within $\mathcal{P}$ but expanded to an even larger set. If the overlap between $\mathcal{X}$ and $\mathcal{Y}$ was known, then it would be possible to simply seek an optimal bijection between the overlapping regions. However, this leads to a chicken-or-the-egg problem, as without some form of mapping it is not possible to determine the overlap.
	
	\subsubsection{Minimum-distortion methods}
	
	Minimum-distortion methods are a subset of intrinsic techniques that operate principally in the spatial domain and penalize the distortion of intrinsic measures. Intrinsic measures include: \eg geodesics~\cite{bronstein2006generalized}; conformal maps~\cite{lipman2009mobius,kim2010mobius,kim2011blended}; heat kernel maps~\cite{ovsjanikov2010one}; and local feature descriptors~\cite{sun2009concise,bronstein2010scale,tombari2010unique,aubry2011wave}. The objective is to then optimize a latent correspondence such that the distortion induced is minimized. There is a substantial amount of literature surrounding this class of approach. The discussions in this survey are therefore limited to methods that consider the problem of partial matching, as these have greater applicability to the problems of registration and reconstruction.

	\paragraph*{Partial matching}
	The early work~\cite{Anguelov2004} proposes a complete registration pipeline for part-to-full shape matching. To aid the non-rigid ICP method, sparse correspondences are established using a belief propagation algorithm. The algorithm seeks to promote matches between similar keypoint signatures probabilistically. The solution space is constrained by pairwise geodesics, which discard matches that lead to a large metric distortion. This helps to ensure correspondences are locally consistent. The probability-based optimization computes a general mapping, which allows the method to handle partial correspondence. Finally, a non-rigid registration algorithm is applied to align the surfaces. The proposed approach is only suitable for piecewise rigid shape deformation, while most works in the past decade facilitate isometric or near-isometric deformation.
	\cite{bronstein2006generalized} extend multi-dimensional scaling to isometrically embed one surface into another. The method is capable of handling partial matching, but is limited to isometric deformation. The method requires points on the input meshes to be sampled coarsely due to the use of geodesics and a quadratic optimization.
    Given a set of sparse keypoints between a partial source shape and a full target shape, \cite{sahillioglu2012scale} select five of the farthest keypoints on the partial shape, and then perform an exhaustive search to find an optimal permutation of correspondences on the target shape that minimizes a geodesic-based distortion measure. The proposed optimization procedure becomes expensive as the number of keypoints on the full shape is increased.
    \cite{rodola2012game,sahillioglu2018genetic} formulate distortion minimization in evolutionary optimization frameworks.
    \cite{sahillioglu2018genetic} employ a genetic algorithm in which permutations of correspondences are represented by chromosomes. Modifying a part of the current correspondence is equivalent to a mutation, and error is measured by geodesic distance. An important advantage of this approach is that it is robust to poor initialization. Similar to the previously discussed works,~\cite{sahillioglu2018genetic} relies on establishing keypoints consistently between shapes. Uniquely, after each optimization iteration, \cite{sahillioglu2018genetic} refines the keypoint locations on the target shape with respect to the pairwise geodesics of candidate correspondences on the source shape.
    \cite{rodola2012game} use an evolutionary game theory-based approach. Rather than determining a bijective correspondence by solving a traditional quadratic assignment problem, they relax this requirement such that not all points must be in correspondence by solving a quadratic semi-assignment problem.
    \cite{ovsjanikov2010one} propose a method that relies on constructing heat kernel maps for each shape. To handle partial correspondence, the heat diffusion parameter is decreased, which effectively reduces the size of the neighborhood represented in the map. However, the approach is limited to isometric deformation.
    \cite{vestner2017efficient} establish a correspondence by solving a linear assignment problem that is refined in an iterative manner. To handle partial correspondence the assignment problem is relaxed by introducing slack variables, which are intended to absorb erroneous candidate correspondences from parts of a full shape that do not overlap.
    \cite{vankaick2013bilateral} develop a geodesic-based pairwise descriptor that is designed to be robust for partial matching. When incorporated with distortion-minimizing techniques, the method is demonstrated to be robust to a degree of topological noise. The descriptor requires large geodesics, therefore it is relatively expensive to compute, and is superseded by recent descriptor learning methods.

    A common problem amongst most of the described approaches is that many rely on solving complex assignment problems that become very expensive to solve at larger problem sizes. This leads to the use of sparse keypoint sampling techniques, which introduce further complexity to consistently sample shapes. Also, methods that use global measures (\eg geodesics over the entire shape) are sensitive to topological change in which geodesic paths may change. Furthermore, for partial matching, these methods are only suited to (near-)isometric deformation due to their reliance on consistent geodesics between overlapping regions. \cite{cosmo2016partial} demonstrate that state-of-the-art minimum-distortion methods tend to perform relatively poorly for partial correspondence problems when compared to functional and learning-based approaches.
	
	\paragraph*{Sampling}
	Broadly, the partial matching methods, discussed previously, employ computationally demanding techniques to solve general assignment problems. To alleviate some of the computational complexity, sparse sets of keypoints are used. Sampling may be necessary in cases where a method requires the same number of points on each shape. A variety of strategies are employed. \cite{tevs2011intrinsic} use a probabilistic approach to construct an order to incrementally introduce points that discriminate highly from previously added points, minimizing an entropic measure. \cite{sahillioglu2011coarse, sahillioglu2012minimum} propose a coarse-to-fine strategy, where points that are at shape extremities and high curvature areas are sampled first. Ideally, both shapes will exhibit similar geometric curvature properties. After applying an initial sampling procedure, \cite{sahillioglu2018genetic} optimize the location of the sampled points on the target shape---helping to reduce geodesic distortion.  \cite{vestner2017efficient} propose a multi-scale/coarse-to-fine approach that uses farthest point sampling~\cite{eldar1997farthest}. A key benefit of this method is that a dense correspondence is retrieved. This is not so critical when used to initialize a registration method; however, if the initial correspondence is too sparse for a particular scenario, such a refinement technique may be necessary. \cite{liu2019spectral} propose a coarsening strategy that seeks to preserve the original shape's spectral representation. \cite{vestner2017efficient} is adapted to use this technique---producing superior results. \cite{corsini2012efficient,nasikun2018fast,lescoat2020spectral} explore other sampling strategies for surfaces that are able to effectively coarsen and accurately approximate spectral properties of the original surface. When sampling partial shapes, the greatest challenge is selecting consistently localized sample points on the surfaces. Methods that support partial matching generally must incorporate machinery for this into the matching portion of the correspondence pipeline~\cite{sahillioglu2012scale,vestner2017efficient,sahillioglu2018genetic}.

	    \begin{figure*}[htp]
        \centering
        \captionsetup[subfigure]{labelformat=empty} % Default: parens
        \subfloat[$\mathcal{Y}$]{\includegraphics[width=0.1\linewidth]{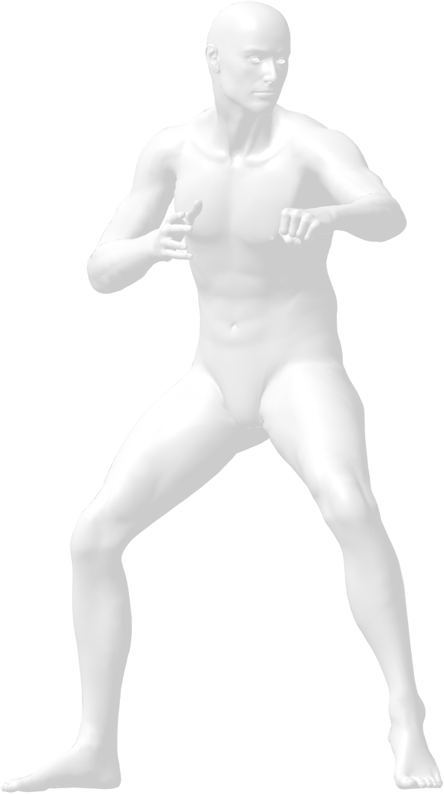}}\hfill
        \subfloat[$\bm{\phi}_2^\mathcal{Y}$]{\includegraphics[width=0.1\linewidth]{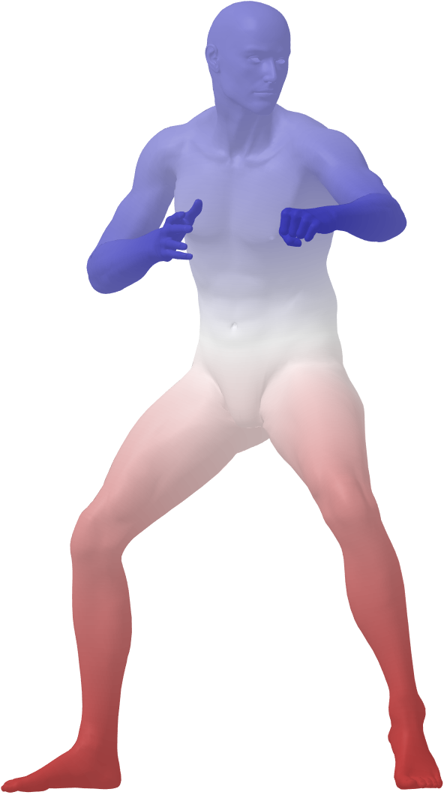}}\hfill
        \subfloat[$\bm{\phi}_3^\mathcal{Y}$]{\includegraphics[width=0.1\linewidth]{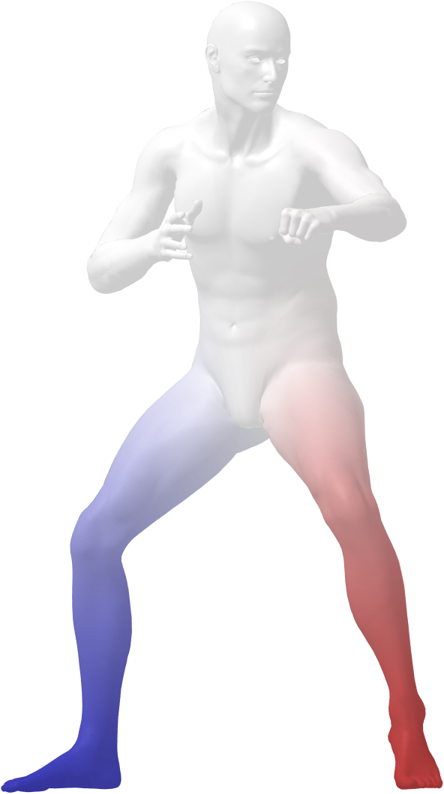}}\hfill
        \subfloat[$\bm{\phi}_4^\mathcal{Y}$]{\includegraphics[width=0.1\linewidth]{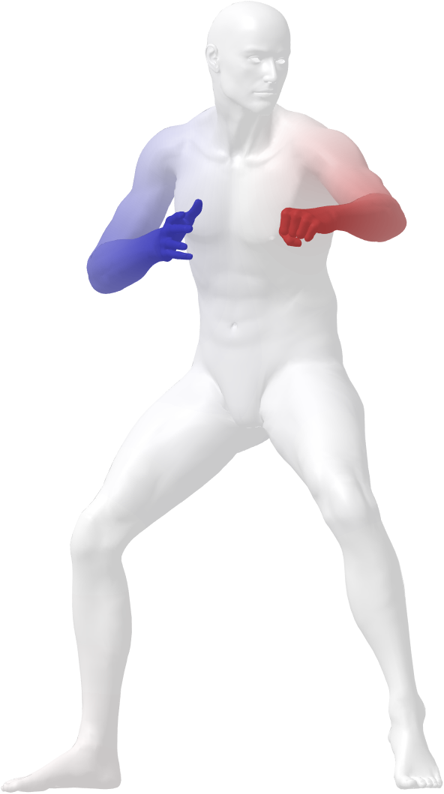}}\hfill
        \subfloat[$\bm{\phi}_5^\mathcal{Y}$]{\includegraphics[width=0.1\linewidth]{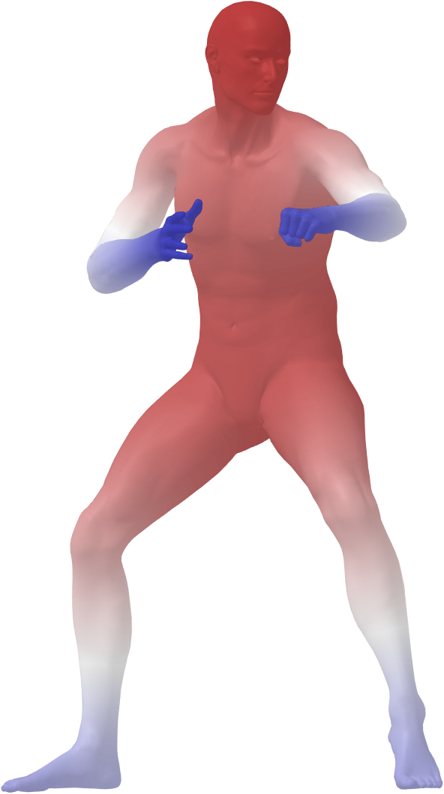}}\hfill
        \subfloat[$\bm{\phi}_6^\mathcal{Y}$]{\includegraphics[width=0.1\linewidth]{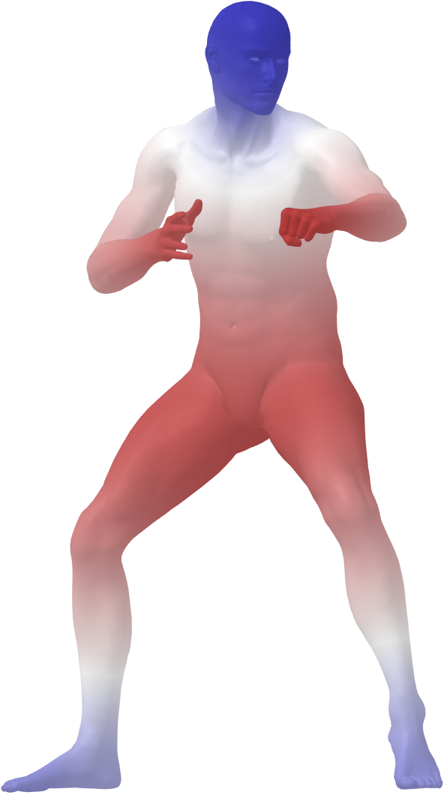}}\hfill
        \subfloat[$\bm{\phi}_7^\mathcal{Y}$]{\includegraphics[width=0.1\linewidth]{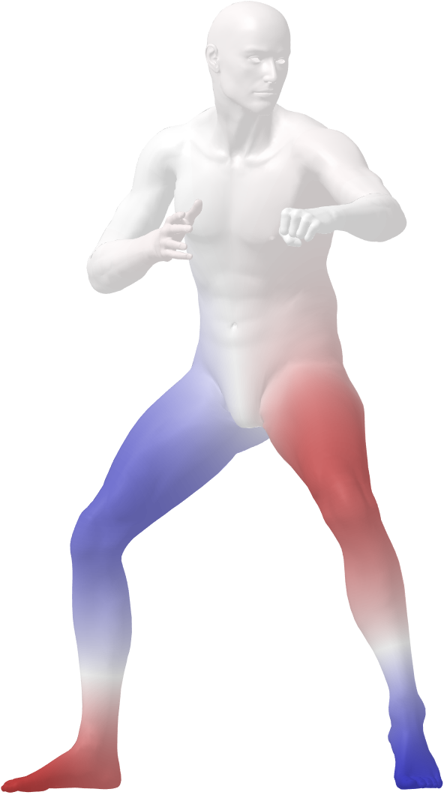}}\hfill
        \subfloat[$\bm{\phi}_8^\mathcal{Y}$]{\includegraphics[width=0.1\linewidth]{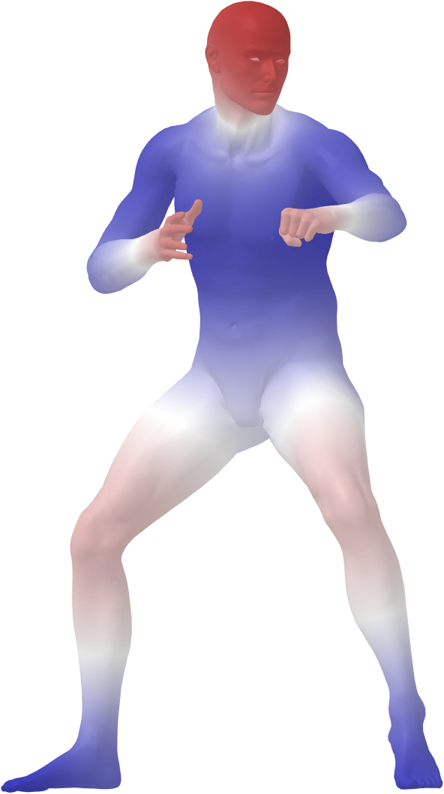}}\hfill
        \subfloat[$\bm{\phi}_9^\mathcal{Y}$]{\includegraphics[width=0.1\linewidth]{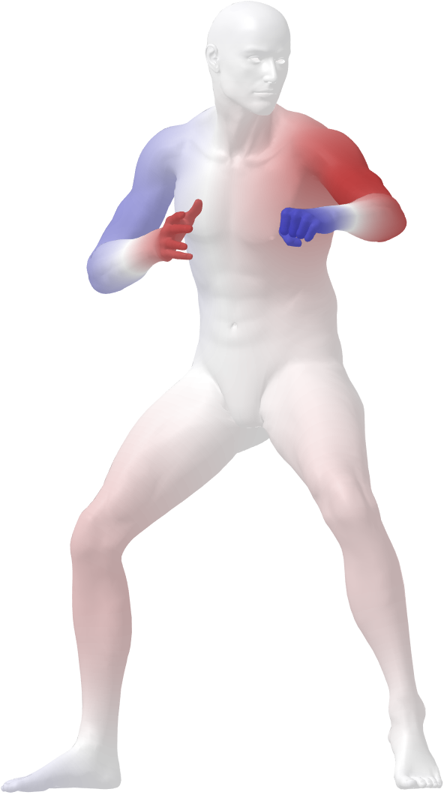}}\hfill\\
        \subfloat[$\mathcal{X}$]{\includegraphics[width=0.07\linewidth]{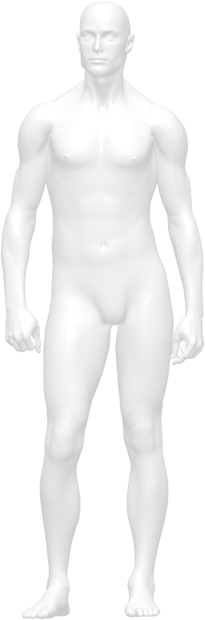}}\hfill
        \subfloat[$\bm{\phi}_2^\mathcal{X}$]{\includegraphics[width=0.07\linewidth]{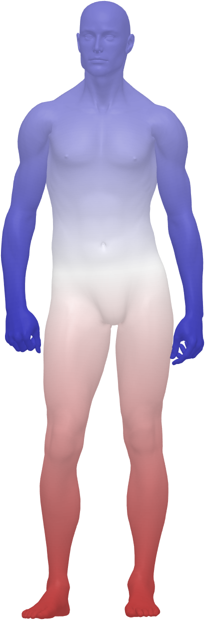}}\hfill
        \subfloat[$\bm{\phi}_3^\mathcal{X}$]{\includegraphics[width=0.07\linewidth]{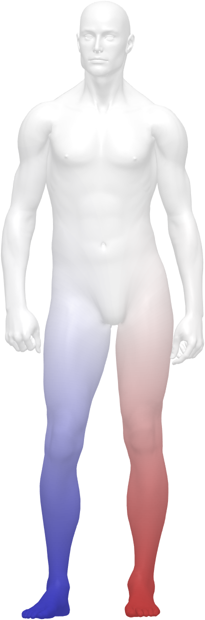}}\hfill
        \subfloat[$\bm{\phi}_4^\mathcal{X}$]{\includegraphics[width=0.07\linewidth]{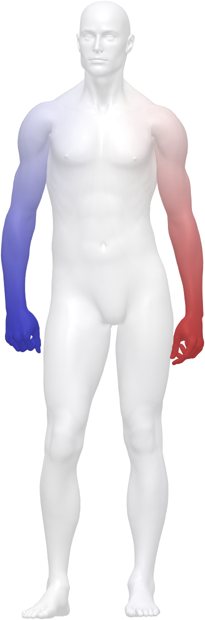}}\hfill
        \subfloat[$\bm{\phi}_5^\mathcal{X}$]{\includegraphics[width=0.07\linewidth]{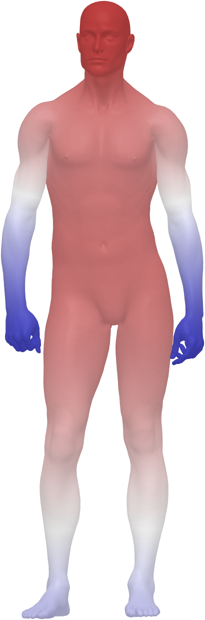}}\hfill
        \subfloat[$\bm{\phi}_6^\mathcal{X}$]{\includegraphics[width=0.07\linewidth]{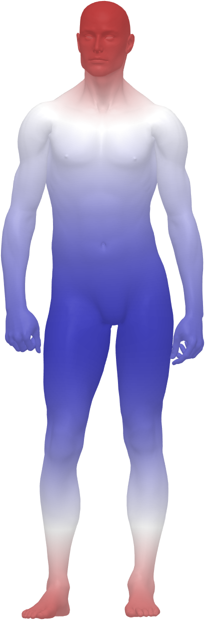}}\hfill
        \subfloat[$\bm{\phi}_7^\mathcal{X}$]{\includegraphics[width=0.07\linewidth]{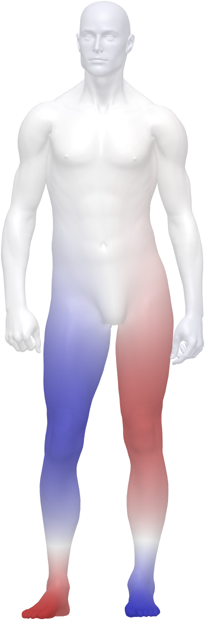}}\hfill
        \subfloat[$\bm{\phi}_8^\mathcal{X}$]{\includegraphics[width=0.07\linewidth]{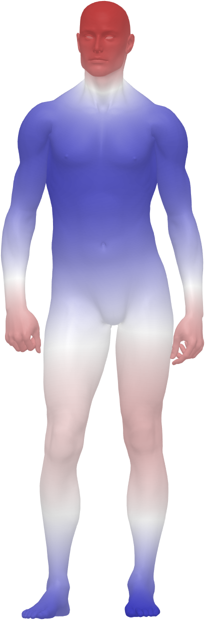}}\hfill
        \subfloat[$\bm{\phi}_9^\mathcal{X}$]{\includegraphics[width=0.07\linewidth]{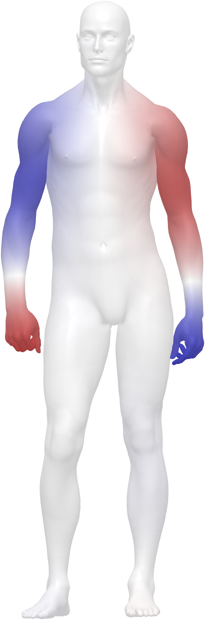}}\\
        \caption{The first eight eigenvectors on $\mathcal{X}$ and $\mathcal{Y}$ in the ascending order of their corresponding eigenvalues. Red indicates where the eigenvector is positive, blue negative, and white is the reference with a value of zero. N.B. the first eigenvalue $\bm{\lambda}_1^\mathcal{X}$ is approximately zero and is therefore omitted.}
        \label{fig:full-eigs-example}
    \end{figure*}

	\paragraph*{Correspondence pruning}\label{sec:corr-pruning}
	Many non-rigid registration methods require a set of accurate sparse correspondences for initialization~\cite{Anguelov2004,huang2008nonrigid,yang2015sparse,li2019robust,yang2019global,yao2020quasi}. A simple approach to compute an initial correspondence is to match feature descriptors between shapes by their similarity. In practice, it is challenging to engineer feature descriptors that are suitably discriminative while also being invariant to different types of deformation. It is therefore necessary to remove poor correspondences. A successful intrinsic-based approach has been to prune correspondences that cause a large amount of geodesic distortion with respect to other candidate correspondences (see Fig.~\ref{fig:initialization}). An early work, \cite{huang2008nonrigid}, measures consistency between geodesics $c_{ij}$ as a ratio of their distances. This formulation has the advantage that distortions between distant geodesics do not disproportionately influence the measure.
    \begin{align}\label{eq:consistency-criterion}
    c_{ij} = \min \left \{ \frac{d_\mathcal{X}(\mathbf{x}_i,\mathbf{x}_j)}{d_\mathcal{Y}(f(\mathbf{x}_i),f(\mathbf{x}_j))},\frac{d_\mathcal{Y}(f(\mathbf{x}_i),f(\mathbf{x}_j))}{d_\mathcal{X}(\mathbf{x}_i,\mathbf{x}_j)} \right \}.
    \end{align}
    A threshold $\tau$ is set experimentally to 0.7. Correspondence pairs are removed if $c_{ij} \leq \tau$.
    Using this measure for determining isometry globally is slow due to the cost associated with computing geodesics over large distances. The proposed method is incorporated into a registration pipeline that is suitable for piecewise rigid deformation.
    
    \cite{tam2014diffusion,tam2014efficient} adapt \eqref{eq:consistency-criterion} to only be computed within a small geodesic disc. The authors note that geodesics to distant points become unreliable due to the presence noise, and under non-isometric deformation or geometric changes such as holes. For queue-based path finding algorithms that are similar to Dijkstra, this is simple to implement: (1) remove nodes from the queue that are beyond the distance threshold; and (2) terminate execution once there are no more unvisited nodes neighboring visited nodes in the queue. \cite{dyke2019non} further adapt this algorithm for non-isometric deformation by replacing the conventional geodesic algorithm with an anisotropic geodesic distance algorithm in which the model for anisotropy is updated iteratively. The described methods are sensitive to their initialization, requiring shape descriptors that are invariant to the exhibited deformation.
    
    \paragraph*{Optimal transport}
    A variety of works have formulated shape correspondence as an optimal transport problem~\cite{solomon2012soft,rodola2012game,solomon2015convolutional,mandad2017varianceminimizing,pai2021fast}. These works consider the Gromov-Wasserstein distance, which is a quadratic problem. This can be computed more efficiently using the Sinkhorn algorithm~\cite{cuturi2013sinkhorn}, which alternately optimizes the transport of the source and the target.
    Relevantly, \cite{bonneel2016wasserstein} demonstrate that optimal transport can be used for shape completion of voxel-based shapes. However, for intrinsic shape correspondence, these techniques are yet to be successfully applied to partial correspondence.

    \paragraph*{Non-isometry} Predominantly, intrinsic methods in the literature typically consider isometric and near-isometric deformation. The challenge of handling non-isometries has been considered in a select range of works. The types of non-isometric deformation addressed by different techniques varies greatly. \cite{lipman2009mobius} search for an optimal conformal map. The method is sensitive to deformations that distort the map, and, therefore, the method performs poorly on partial matching problems. \cite{kim2011blended} combine multiple conformal maps to produce a ``blended map'' that minimizes this distortion.  \cite{tam2014diffusion} seek locally isometric correspondences between shapes, using a geodesic-based global consistency check to prune inconsistent correspondences. This global check can be relaxed to increase the degree of global non-isometry permitted. \cite{dyke2019non} extend~\cite{tam2014diffusion} to handle local anisotropy. \cite{solomon2016entropic} solve an optimal transportation problem, that for full-to-full matching, can handle some non-isometric distortion; however, the approach is not guaranteed to obtain the globally optimal solution. \cite{vestner2017efficient} implement a probabilistic framework that is suitable for finding smooth maps. \cite{sahillioglu2020recent} provides further discussions on the development of shape correspondence approaches that handle non-isometries.

	\paragraph*{Shape recovery}
    Given a template, \cite{devir2009reconstruction} consider the problem of reconstructing a deformed triangle mesh by using an optimization-based technique. An objective function comprising of a regularization term and a data term is proposed. The regularization term uses a pairwise geodesic-based distance between the original template and the deformed template, which seeks to reduce distortion. The data term measures the distance between corresponding points on the deformed shape and the target shape.

	To summarize, metrics used by minimum-distortion methods, such as geodesics, are highly sensitive to geometric and topological noise. For these methods to cope with partial correspondence, an assumption of \mbox{(near-)isometric} deformation must be made. The sampling strategies discussed exhibit inconsistent results when sampling on partial or full surfaces.

	\begin{figure*}[htp]
        \centering
        \captionsetup[subfigure]{labelformat=empty} % Default: parens
        \subfloat[$\mathcal{V}$]{\includegraphics[width=0.07\linewidth]{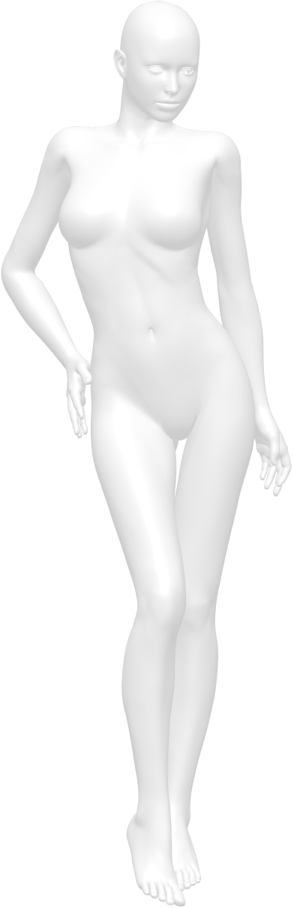}}\hfill
        \subfloat[$\bm{\phi}_2^\mathcal{V}$]{\includegraphics[width=0.07\linewidth]{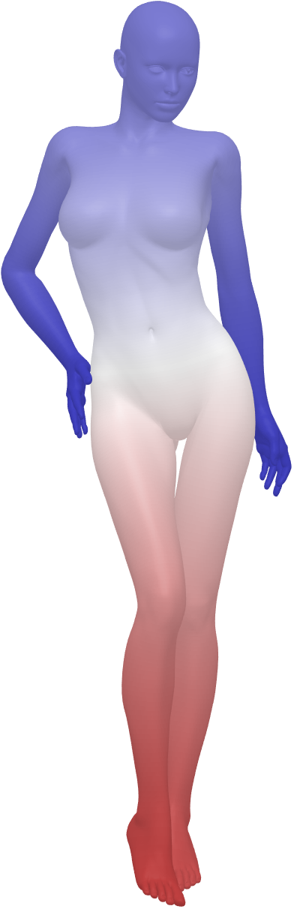}}\hfill
        \subfloat[$\bm{\phi}_3^\mathcal{V}$]{\includegraphics[width=0.07\linewidth]{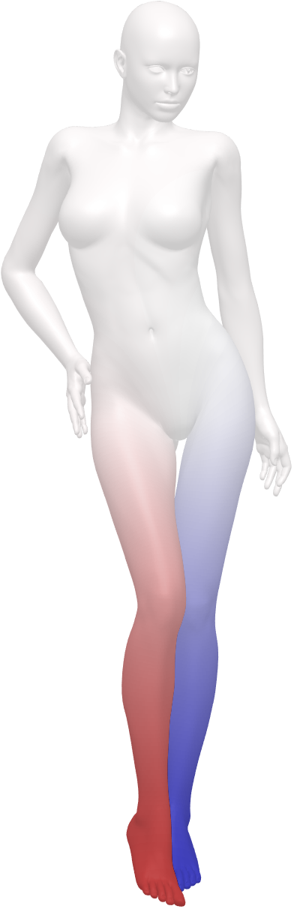}}\hfill
        \subfloat[$\bm{\phi}_4^\mathcal{V}$]{\includegraphics[width=0.07\linewidth]{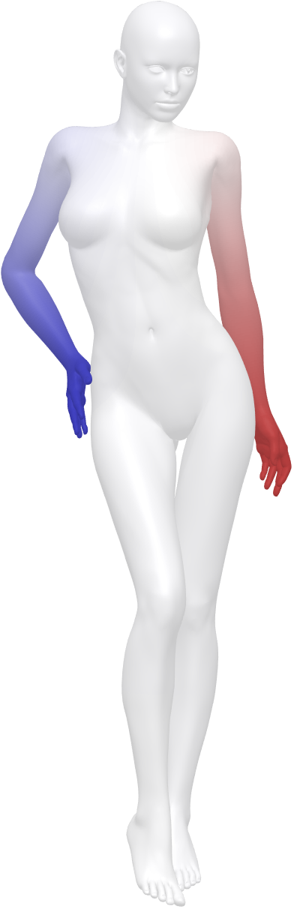}}\hfill
        \subfloat[$\bm{\phi}_5^\mathcal{V}$]{\includegraphics[width=0.07\linewidth]{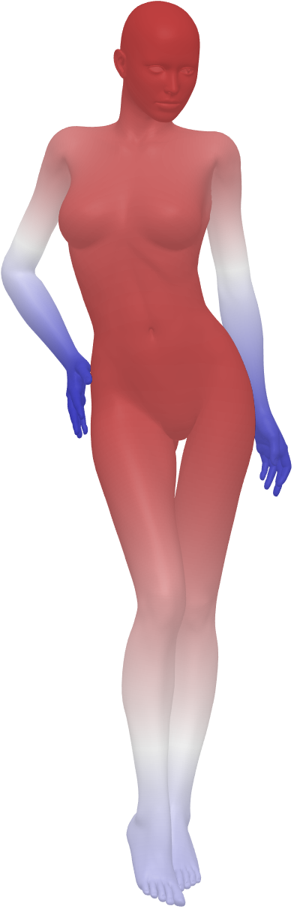}}\hfill
        \subfloat[$\bm{\phi}_6^\mathcal{V}$]{\includegraphics[width=0.07\linewidth]{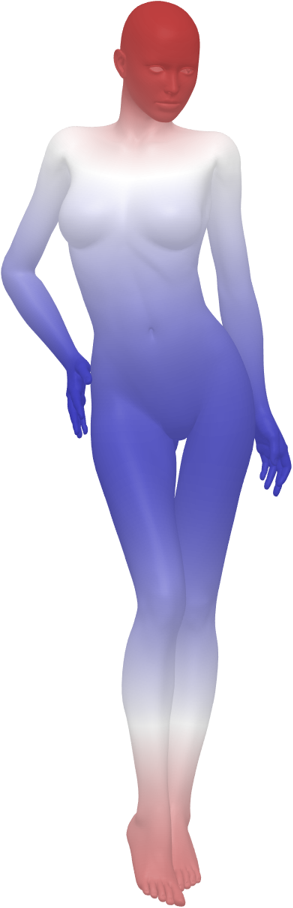}}\hfill
        \subfloat[$\bm{\phi}_7^\mathcal{V}$]{\includegraphics[width=0.07\linewidth]{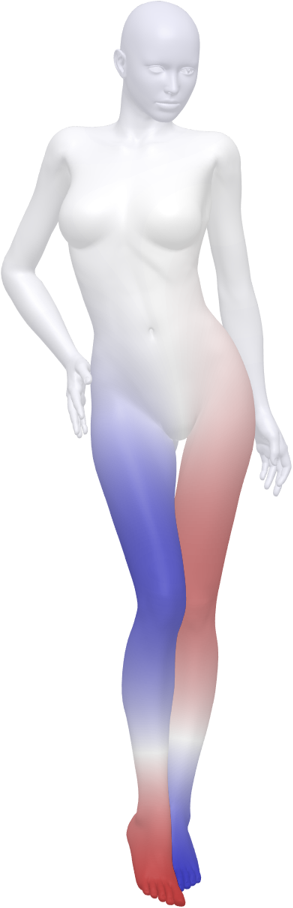}}\hfill
        \subfloat[$\bm{\phi}_8^\mathcal{V}$]{\includegraphics[width=0.07\linewidth]{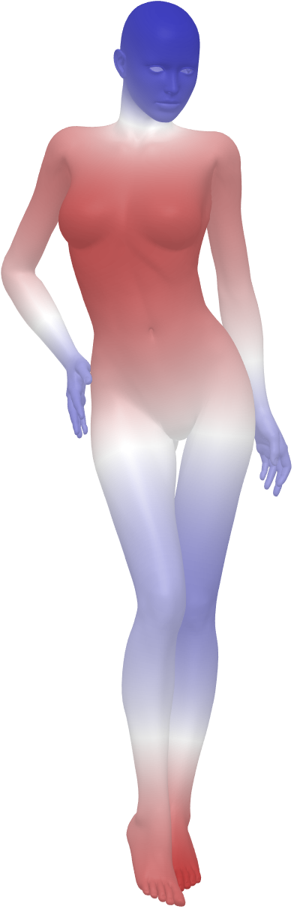}}\hfill
        \subfloat[$\bm{\phi}_9^\mathcal{V}$]{\includegraphics[width=0.07\linewidth]{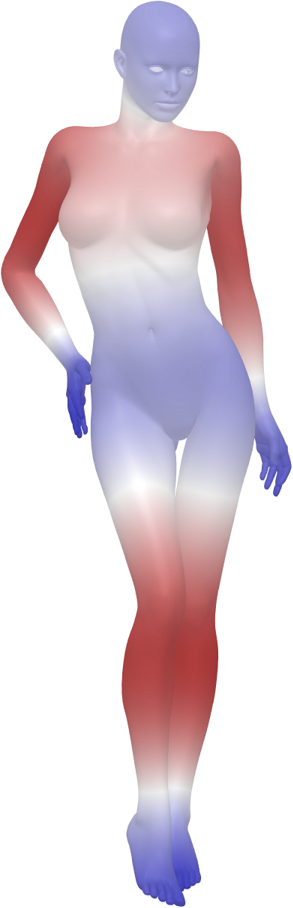}}\hfill\\
        \caption{The first eight eigenvectors on shape $\mathcal{V}$ in the ascending order of their corresponding eigenvalues.}
        \label{fig:full-eigs-example-2}
    \end{figure*}

	\subsubsection{Spectral/Functional mapping methods}
    
    \paragraph*{Functional mapping fundamentals}
    
    Rather than directly computing a point-to-point map, the shapes are abstracted by functions defined upon each surface. The functions are computed from the Laplace--Beltrami operator. A common discretization is the cotangent Laplacian~\cite{meyer2002discrete} $\mathbf{L}=\mathbf{D}^{-1}\mathbf{M}$, which is composed of $\mathbf{D}^{n \times n}$ a diagonal mass matrix equal to $\frac{1}{3}$ of the total area of the faces within the 1-ring neighborhood of each vertex, and $\mathbf{M}^{n \times n}$ is a matrix of cotangent weights between connected vertices:
    \[m_{ij} = \begin{cases}
        \frac{1}{2}(\cot{\alpha_{ij}} + \cot{\beta_{ij}}) & i,j \in E \\
        0 & \text{otherwise},
    \end{cases}
    \]
    where $\alpha_{ij}$ and $\beta_{ij}$ are the internal angles at the vertices opposite edge $ij$.

    The eigenbasis of the intrinsic Laplace--Beltrami operator---denoted by $\bm{\Phi}$, with corresponding eigenvalues $\bm{\Lambda}$---is often selected as it is invariant under isometric deformation. 
    \begin{align}\label{eq:eig-decomp}
    \mathbf{M}\bm{\Phi} = \mathbf{D}\bm{\Phi}\bm{\Lambda}.
    \end{align}
    Fig.~\ref{fig:full-eigs-example} illustrates the characteristics of the eigenvectors. It can be seen that each eigenfunction $\bm{\phi}_i$ varies smoothly over the surface. As the eigenvalue increases, the oscillations observed in the eigenfunction are similar between the isometrically deforming shapes, up to their sign (\ie $\bm{\phi}_i^\mathcal{X} \approx \pm\bm{\phi}_i^\mathcal{Y}$). For the motivations and specific implementation details, please refer to the seminal paper on functional maps~\cite{ovsjanikov2012functional}.

    Given this functional representation, it is possible to compute a functional correspondence matrix $\mathbf{C}$ that maps corresponding eigenvectors between surfaces, \ie $\bm{\Phi}_\mathcal{Y} \mathbf{C} = \bm{\Phi}_\mathcal{X}$, where $\bm{\Phi}$ is a matrix of column eigenvectors. N.B. $\bm{\Phi}_\mathcal{Y}$ and $\bm{\Phi}_\mathcal{X}$ are truncated to $k$ eigenvalues. In the example in Fig.~\ref{fig:full-eigs-example}, this would induce a map that would approximately be an identity matrix $\mathbf{I}$, where $\bm{\phi}_i^\mathcal{Y}$ maps to $\bm{\phi}_i^\mathcal{X}$.

    Since the Laplacian operator is only invariant under isometric deformation, the strong diagonal of $\mathbf{C}$ gradually weakens to greater non-isometries. This effect is most significant in eigenvectors with larger eigenvalues, while smaller eigenvalues are progressively less affected. This leads to a pre-dominantly diagonal structure, with the strongest diagonal found between smaller eigenvalues. $\bm{\phi}_{3,7,8}^\mathcal{V}$ in Fig.~\ref{fig:full-eigs-example-2} are negative equivalents of $\bm{\phi}_{3,7,8}^\mathcal{Y}$ and $\bm{\phi}_{3,7,8}^\mathcal{X}$ in Fig.~\ref{fig:full-eigs-example}, and $\bm{\phi}_{9}^\mathcal{V}$ presents a completely different eigenvector that is not present in any $\bm{\phi}_i^\mathcal{Y}$ and $\bm{\phi}_i^\mathcal{X}$ ($i\leq9$).
    
    An optimal $\mathbf{C}$ can generally be computed by solving the following linear system:
    \begin{align}\label{eq:fmap}
    \argmin_{\mathbf{C}} \| \bm{\Phi}_\mathcal{Y} - \mathbf{C}\bm{\Phi}_\mathcal{X} \|^2 + E_\text{corr}.
    \end{align}
    Where $E_\text{corr}$ imposes some regularization on $\mathbf{C}$, \eg promoting Laplacian commutativity by penalizing matches of dissimilar eigenvalues $E_\text{corr} = \alpha \left \| \bm{\Lambda}_{\mathcal{X}}\mathbf{C} - \mathbf{C}\bm{\Lambda}_{\mathcal{Y}} \right \|^2$.

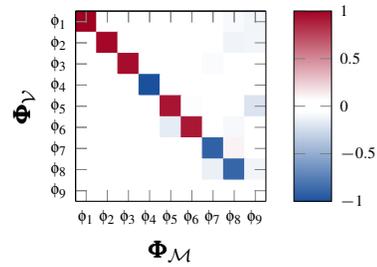
\begin{figure}[tb]
        \centering
        \begin{tikzpicture}

%\begin{scope}[local bounding box=plots]
\pgfplotsset{
%fprintf('rgb255=(%.0f,%.0f,%.0f)\n',interp1([27,81,156;255,255,255;161,0,33],[linspace(1,2,9),linspace(2,3,9)])')
/pgfplots/colormap={temp}{rgb255=(27,81,156)
rgb255=(56,103,168)
rgb255=(84,124,181)
rgb255=(112,146,193)
rgb255=(141,168,206)
rgb255=(170,190,218)
rgb255=(198,212,230)
rgb255=(226,233,243)
rgb255=(255,255,255)
rgb255=(255,255,255)
rgb255=(243,223,227)
rgb255=(232,191,200)
rgb255=(220,159,172)
rgb255=(208,128,144)
rgb255=(196,96,116)
rgb255=(184,64,88)
rgb255=(173,32,61)
rgb255=(161,0,33)
}}
\usepgfplotslibrary{colorbrewer}

\begin{axis}[
width=0.3\linewidth,%4.521in
height=0.3\linewidth,%3.566in
at={(0.758in,0.481in)},
scale only axis,
mesh/ordering=x varies,
unbounded coords=jump,
colorbar,
ticklabel style = {font=\scriptsize},
xtick={0.5,1.5,2.5,3.5,4.5,5.5,6.5,7.5,8.5},
xticklabels={$\phi_1$,$\phi_2$,$\phi_3$,$\phi_4$,$\phi_5$,$\phi_6$,$\phi_7$,$\phi_8$,$\phi_9$},
ytick={0.5,1.5,2.5,3.5,4.5,5.5,6.5,7.5,8.5},
yticklabels={$\phi_1$,$\phi_2$,$\phi_3$,$\phi_4$,$\phi_5$,$\phi_6$,$\phi_7$,$\phi_8$,$\phi_9$},
view={0}{90},
y dir=reverse,
xlabel={$\bm{\Phi}_\mathcal{M}$},
ylabel={$\bm{\Phi}_\mathcal{V}$},
xlabel near ticks,
ylabel near ticks,
%colormap/jet,
colorbar style={xticklabel pos=upper,
                xlabel style={yshift=.22cm},
                ytick={-1,-0.5,0,0.5,1},
                point meta min=-1,
                point meta max=1,
                },
]
%\addplot3[surf,shader=flat corner,draw=\pgfkeysvalueof{/pgfplots/faceted color},mesh/rows=10,mesh/cols=10] coordinates {
\addplot3[surf,shader=flat corner,mesh/rows=10,mesh/cols=10] coordinates {
(0,0,1.071732) (1,0,0.052000) (2,0,-0.000264) (3,0,-0.007937) (4,0,-0.022233) (5,0,0.005757) (6,0,-0.010197) (7,0,-0.071480) (8,0,-0.098926) (9,0,0.053266) 
(0,1,-0.000000) (1,1,1.044258) (2,1,-0.021911) (3,1,-0.014093) (4,1,-0.002628) (5,1,-0.009411) (6,1,-0.005468) (7,1,-0.100006) (8,1,-0.094250) (9,1,0.048039) 
(0,2,0.000000) (1,2,0.023526) (2,2,1.021338) (3,2,-0.006859) (4,2,-0.008508) (5,2,0.009414) (6,2,-0.058393) (7,2,0.002297) (8,2,0.017175) (9,2,0.001819) 
(0,3,0.000000) (1,3,-0.013021) (2,3,-0.007434) (3,3,-1.024504) (4,3,-0.030958) (5,3,-0.013686) (6,3,-0.003871) (7,3,-0.015384) (8,3,0.000293) (9,3,0.381729) 
(0,4,0.000000) (1,4,-0.003930) (2,4,0.004788) (3,4,-0.017292) (4,4,0.994364) (5,4,0.091620) (6,4,0.007448) (7,4,-0.019953) (8,4,-0.196738) (9,4,0.023871) 
(0,5,0.000000) (1,5,-0.022037) (2,5,-0.007382) (3,5,-0.012875) (4,5,-0.144920) (5,5,0.984886) (6,5,-0.030132) (7,5,-0.076413) (8,5,-0.004701) (9,5,0.005667) 
(0,6,-0.000000) (1,6,0.005878) (2,6,0.003844) (3,6,0.002610) (4,6,-0.002707) (5,6,-0.005226) (6,6,-0.933772) (7,6,0.134120) (8,6,0.001220) (9,6,0.013982) 
(0,7,0.000000) (1,7,-0.000278) (2,7,0.004933) (3,7,0.002093) (4,7,-0.004322) (5,7,-0.006714) (6,7,-0.121829) (7,7,-0.907124) (8,7,-0.099062) (9,7,0.048620) 
(0,8,-0.000000) (1,8,-0.007757) (2,8,-0.001343) (3,8,-0.007110) (4,8,-0.012784) (5,8,0.002729) (6,8,0.016216) (7,8,0.005624) (8,8,0.032600) (9,8,0.814174) 
(0,9,-0.000000) (1,9,0.007657) (2,9,-0.001352) (3,9,0.008103) (4,9,0.046267) (5,9,-0.021004) (6,9,0.017772) (7,9,0.169420) (8,9,-0.845716) (9,9,0.025395) 
};
\end{axis}

%\end{scope}
\end{tikzpicture}
        \caption{A functional map $\mathbf{C}$ between $\mathcal{X}$ in Fig.~\ref{fig:full-eigs-example} \& $\mathcal{V}$ in Fig.~\ref{fig:full-eigs-example-2}.}
        \label{fig:fmap-example}
    \end{figure}

    Considering the example of $\mathcal{X}$ and $\mathcal{V}$, once $\mathbf{C}$ is obtained (such as in Fig.~\ref{fig:fmap-example}), we see that negated corresponding eigenvectors cause negative entries along the diagonal, while eigenvectors that do not correspond have values close to zero.
    
    Much like the permutation matrix $\mathbf{P}$, for computing bijective correspondence, a key property of $\mathbf{C}$ is that it should be an orthogonal matrix. Since $\mathbf{C}$ is a soft map, it is a member of the larger Stiefel manifold set $\mathcal{S}_{k_1 \times k_2} = \{ \mathbf{C} \in \mathbb{R}^{k_1 \times k_2} \mid \mathbf{C}^{\top}\mathbf{C} = \mathbf{I} \}$, where $\mathcal{P} \subset \mathcal{S}$. This can either be incorporated explicitly by limiting the solution space to $\mathcal{S}$ or added as a regularization term~\cite{rodola2017partial} (\eg $\off(\mathbf{C}^{\top}\mathbf{C}) = \sum_{i \neq j}^k (\mathbf{C}^{\top}\mathbf{C})_{ij}^2$~\cite{cardoso1996jacobi}).

    For shape correspondence, it is necessary to recover a point-to-point mapping. \cite{ovsjanikov2012functional} propose the following energy term:
    \begin{align}\label{eq:pmap}
    \argmin_{\mathbf{P} \in \mathcal{P}} \| \bm{\Phi}_\mathcal{Y} - \mathbf{P}(\mathbf{C}\bm{\Phi}_\mathcal{X}) \|^2,
    \end{align}
    where $\mathbf{C}$ is fixed. This may be considered to be a form of point alignment in a high-dimensional space where other alignment schemes can be employed~\cite{rodola2015pointwise,rodola2017regularized}.
    
    \paragraph*{Partial matching}
    For partial matching, the requirements of~\eqref{eq:bijective-perms} must be relaxed. \cite{rodola2017partial} propose to optimize a correspondence map by adopting suitable correspondence and part regularization terms, \ie
    \[
    E = E_\text{data} + \underbrace{E_\text{corr} + E_\text{part}}_\text{regularization},
    \]
    where the $E_\text{corr}$ and $E_\text{part}$ are solved in alternating steps to optimize the correspondence and the part separately. In the first step, a correspondence is estimated between a part of the full shape $\mathcal{X}' \subseteq \mathcal{X}$ and a partial shape $\mathcal{Y}$. Given a deformation that is near-isometric, $E_\text{corr}$ helps constrain the optimizer to guide it towards a sensible solution.
    The part regularizer seeks to match the area between $\mathcal{Y}$ and $\mathcal{X}'$, while also minimizing the boundary length $\ell$ of $\mathcal{X}'$,
    \[
    \argmin_{\mathcal{X}'} E_\text{data} + E_\text{part},
    \]
    where
    \[
    E_\text{part} = \alpha \left | \areafunc \left ( \mathcal{Y} \right ) - \areafunc \left ( \mathcal{X}' \right ) \right |^2 + \beta \ell(\partial \mathcal{X}').
    \]
    The $E_\text{data}$ term is modified to optimize the functional map for the subset of eigenvectors $\bm{\Phi}_{\mathcal{X}'} \subseteq \bm{\Phi}_{\mathcal{X}}$ for $\mathcal{X}'$,
    \[
    \argmin_{\mathbf{C}} E_\text{data} + \alpha E_\text{corr},
    \]
    where
    \[
    E_\text{data} = \| \bm{\Phi}_\mathcal{Y} - \mathbf{C}\bm{\Phi}_{\mathcal{X}'} \|^2.
    \]
    
    In practice, this is implemented using a fuzzy approach. This is done by giving eigenvectors that are not in $\mathcal{X}'$ a low weight, rather than completely removing them.
    
    \begin{figure*}[htp]
        \centering
        \captionsetup[subfigure]{labelformat=empty} % Default: parens
        \subfloat[$\mathcal{Y}$]{\includegraphics[width=0.1\linewidth]{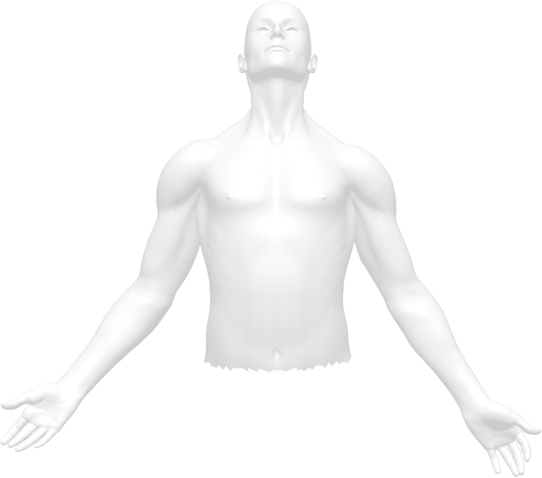}}\hfill
        \subfloat[$\bm{\phi}_2^\mathcal{Y}$]{\includegraphics[width=0.1\linewidth]{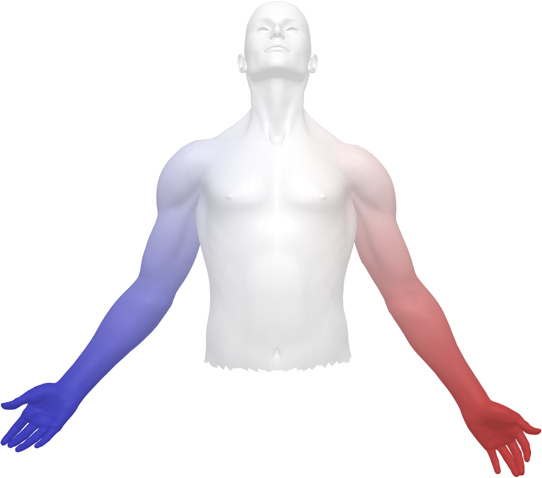}}\hfill
        \subfloat[$\bm{\phi}_3^\mathcal{Y}$]{\includegraphics[width=0.1\linewidth]{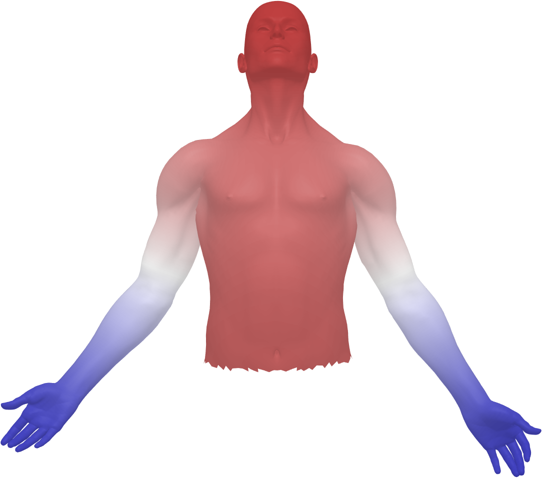}}\hfill
        \subfloat[$\bm{\phi}_4^\mathcal{Y}$]{\includegraphics[width=0.1\linewidth]{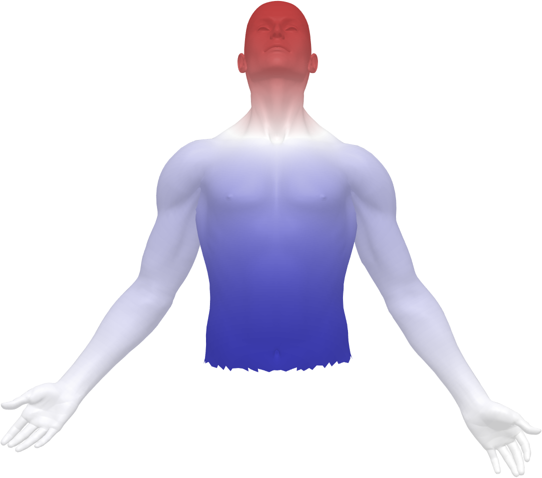}}\hfill
        \subfloat[$\bm{\phi}_5^\mathcal{Y}$]{\includegraphics[width=0.1\linewidth]{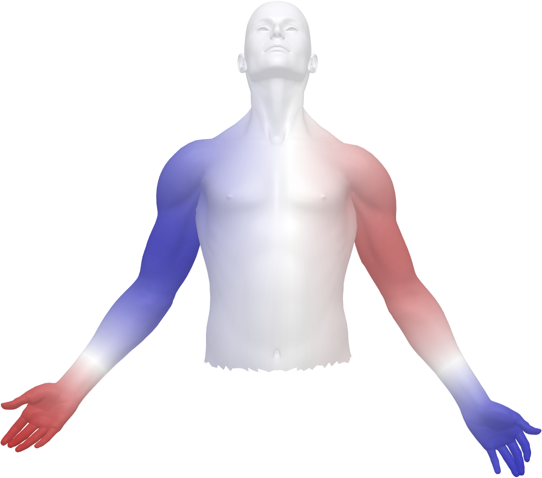}}\hfill
        \subfloat[$\bm{\phi}_6^\mathcal{Y}$]{\includegraphics[width=0.1\linewidth]{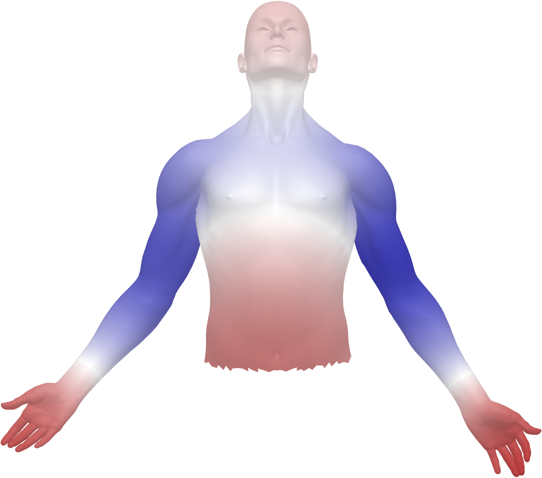}}\hfill
        \subfloat[$\bm{\phi}_7^\mathcal{Y}$]{\includegraphics[width=0.1\linewidth]{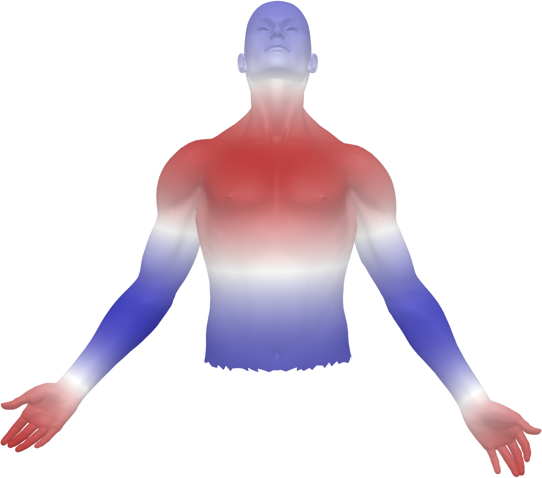}}\hfill
        \subfloat[$\bm{\phi}_8^\mathcal{Y}$]{\includegraphics[width=0.1\linewidth]{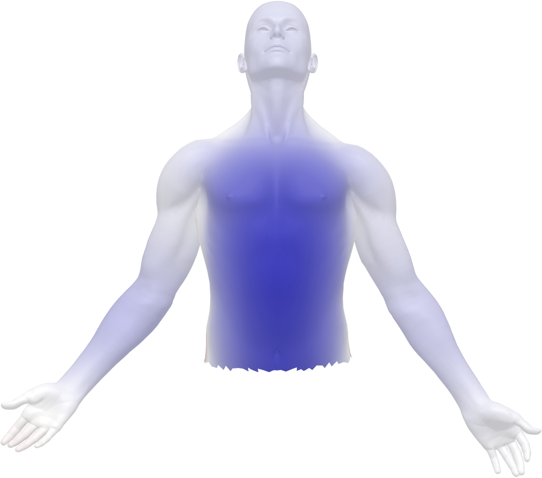}}\hfill
        \subfloat[$\bm{\phi}_9^\mathcal{Y}$]{\includegraphics[width=0.1\linewidth]{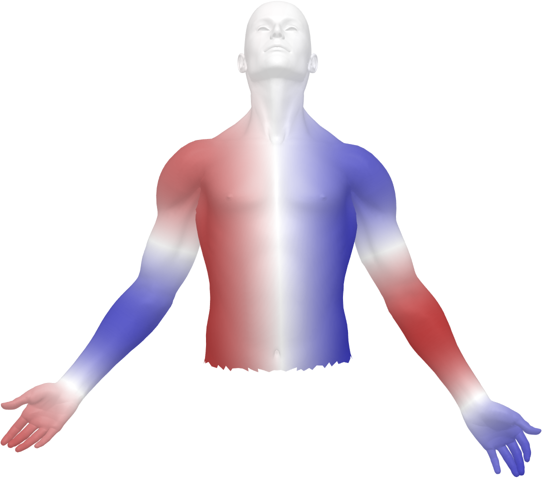}}\\
        \subfloat[$\mathcal{X}$]{\includegraphics[width=0.075\linewidth]{figs/partial-eigs/michael15_1.png}}\hfill
        \subfloat[$\bm{\phi}_2^\mathcal{X}$]{\includegraphics[width=0.075\linewidth]{figs/partial-eigs/michael15_2.png}}\hfill
        \subfloat[$\bm{\phi}_3^\mathcal{X}$]{\includegraphics[width=0.075\linewidth]{figs/partial-eigs/michael15_3.png}}\hfill
        \subfloat[$\bm{\phi}_4^\mathcal{X}$]{\includegraphics[width=0.075\linewidth]{figs/partial-eigs/michael15_4.png}}\hfill
        \subfloat[$\bm{\phi}_5^\mathcal{X}$]{\includegraphics[width=0.075\linewidth]{figs/partial-eigs/michael15_5.png}}\hfill
        \subfloat[$\bm{\phi}_6^\mathcal{X}$]{\includegraphics[width=0.075\linewidth]{figs/partial-eigs/michael15_6.png}}\hfill
        \subfloat[$\bm{\phi}_7^\mathcal{X}$]{\includegraphics[width=0.075\linewidth]{figs/partial-eigs/michael15_7.png}}\hfill
        \subfloat[$\bm{\phi}_8^\mathcal{X}$]{\includegraphics[width=0.075\linewidth]{figs/partial-eigs/michael15_8.png}}\hfill
        \subfloat[$\bm{\phi}_9^\mathcal{X}$]{\includegraphics[width=0.075\linewidth]{figs/partial-eigs/michael15_9.png}}\\
        \caption{The first eight eigenvectors on $\mathcal{X}$ and $\mathcal{Y}$ ordered by their corresponding eigenvalues.}
        \label{fig:partial-eigs-example}
    \end{figure*}
    % [PFM] characterises the sloped diagonal as a ratio of the areas between $\mathcal{X}'$ and $\mathcal{Y}$.
    
    A fundamental observation by~\cite{rodola2017partial} is that for a given eigenvector $\bm{\phi}_i^\mathcal{Y}$ on $\mathcal{Y}$, there exists a corresponding $\bm{\phi}_j^\mathcal{X}$ where $i \leq j$. This is illustrated in Fig.~\ref{fig:partial-eigs-example}, where the following pairs of eigenvectors (partially) match: ($\bm{\phi}_2^\mathcal{Y}$, $\bm{\phi}_4^\mathcal{X}$); ($\bm{\phi}_3^\mathcal{Y}$, $\bm{\phi}_5^\mathcal{X}$); and ($\bm{\phi}_5^\mathcal{Y}$, $\bm{\phi}_9^\mathcal{X}$).
    
    The proposed formulation by~\cite{rodola2017partial} is not suitable for part-to-part matching as it relies on an estimate of overlap that assumes the mapping to be injective. \cite{cosmo2016matching} extend~\cite{rodola2017partial} for part-to-part matching by attempting to optimize for both the part $\mathcal{Y}'$ and $\mathcal{X}'$. To do this, three optimization steps are used: first $\mathbf{C}$ is computed, then $\mathcal{X}'$ is estimated, and finally $\mathcal{Y}'$ is estimated, before repeating all three steps until a convergence criterion is met. Similarly, \cite{litany2016nonrigid} adapt~\cite{rodola2017partial} for matching multiple parts ($\mathcal{Y}'_i$ and $\mathcal{X}'_i$) between a full shape ($\mathcal{X}'_i \subseteq \mathcal{X}$) and multiple partial shapes ($\mathcal{Y}'_i \subseteq \mathcal{Y}_i$). Unfortunately, the proposed formulations do not scale well, and require optimization in both the spectral domain and the spatial domain.
    
    \cite{litany2017fully} propose an approach that is conducted entirely in the spectral domain for computing part-based correspondence. The method builds upon the full-to-full matching method by~\cite{kovnatsky2013coupled}. The authors align the eigenbases $\bm{\Phi}_\mathcal{X}$ and $\bm{\Phi}_\mathcal{Y}$ jointly by formulating an optimization in the form of the joint approximate diagonalization problem,
    \[
    \argmin_{\mathbf{A},\mathbf{B} \in \mathcal{S}_{n \times n}} \underbrace{\off(\mathbf{A}^\top\bm{\Lambda}_\mathcal{X}\mathbf{A}) + \off(\mathbf{B}^\top\bm{\Lambda}_\mathcal{Y}\mathbf{B})}_\text{diagonalization} + \alpha \underbrace{\left \| \mathbf{P} \bm{\Phi}_\mathcal{X}\mathbf{A} - \mathbf{Q} \bm{\Phi}_\mathcal{Y} \mathbf{B} \right \|^2_F}_\text{coupling}.
    \]
    $\mathbf{P},\mathbf{Q} \in \left \{ 0,1 \right\}^{l \times n}$ are matrices that map between $l$ initial corresponding vertices and rows from the eigenbases. In practice, a discrete point $p$ is a row in $\bm{\Phi}$, \ie $\bm{\phi}_i(p) \in \mathbb{R}$ for $i=1,\dots,k$. Effectively, $\mathbf{P}\bm{\Phi}_\mathcal{X}$ and $\mathbf{Q}\bm{\Phi}_\mathcal{Y}$, represent a subset of corresponding rows from each eigenbasis. The diagonalization terms promote $\mathbf{A}$ and $\mathbf{B}$ to be orthogonal matrices. These are used to perform an orthogonal transform, which rotates and flips $\mathbf{P}\bm{\Phi}_\mathcal{X}$ and $\mathbf{Q}\bm{\Phi}_\mathcal{Y}$. Importantly, the inner products are preserved. The coupling term then measures the similarity of corresponding rows between $\mathbf{P}\bm{\Phi}_\mathcal{X}$ and $\mathbf{Q}\bm{\Phi}_\mathcal{Y}$. The aligned eigenbases can now be computed as:
    \[
    \bm{\hat{\Phi}}_\mathcal{X} = \bm{\Phi}_\mathcal{X}\mathbf{A} \quad \text{and} \quad \bm{\hat{\Phi}}_\mathcal{Y} = \bm{\Phi}_\mathcal{Y}\mathbf{B}.
    \]
    This formulation is not suitable for partial matching problems, as all rows in $\mathbf{P}\bm{\Phi}_\mathcal{X}$ and $\mathbf{Q}\bm{\Phi}_\mathcal{Y}$ must be aligned.
    
    For partial correspondence~\cite{litany2017fully} modify the coupling term to the following:
    \begin{equation}\label{eq:partial-spectral}
    \begin{split}
    \argmin_{\mathbf{A},\mathbf{B} \in \mathcal{S}_{k \times r}} \off(\mathbf{A}^\top\bm{\Lambda}_\mathcal{X}\mathbf{A}) + \off(\mathbf{B}^\top\bm{\Lambda}_\mathcal{Y}\mathbf{B}) \\ + \alpha \left \| \mathbf{W}_r \left ( \bm{\Phi}_\mathcal{X}\mathbf{A} - \bm{\Phi}_\mathcal{Y} \mathbf{B} \right ) \right \|_{2,1}.
    \end{split}
    \end{equation}
    The solution space is constrained to a set of rectangular orthogonal matrices  $\mathcal{S}_{k \times r}$. The rank $r = \rank(\mathbf{C})$ is based on an estimate of the overlapping area between $\mathcal{X}$ and $\mathcal{Y}$, where $r \ll k$. A mask $\mathbf{W}_r = \begin{pmatrix}\mathbf{I}_{r \times r} & \mathbf{0}_{r \times k-r}\end{pmatrix}^\top$ is applied. The $\ell_{2,1}$-norm ($\left \| \,\cdot\, \right \|_{2,1}$) promotes sparsity in the columns of the masked matrix. The proposed method demonstrates a strictly spectral approach to the partial functional map problem. How to effectively determine a suitable estimate of $r$ for part-to-part matching is unclear. The authors suggest that an optimal $r$ can be estimated by solving~\eqref{eq:partial-spectral} for multiple $r$'s; however, this is an expensive procedure.
    
    \cite{melzi2019zoomout} propose a fully-spectral method that uses a coarse-to-fine approach in which the size of $\mathbf{C}$ is increased incrementally. $\mathbf{C}$ initially only maps low-frequency eigenvectors, as the size of $\mathbf{C}$ increases the map is refined by high-frequency eigenvectors. The refinement process comprises of two steps in which the method transforms correspondence data between a functional map $\mathbf{C}$ and a pointwise map $\mathbf{P}$, \ie \eqref{eq:pmap}. Initially, $\mathbf{C} = \mathbf{I}_{2 \times 2}$ (or larger), and a point-to-point map $\mathbf{P}$ is recovered, using $\bm{\Phi}^\mathcal{Y}_{n \times k_1}$ and $\bm{\Phi}^\mathcal{X}_{n \times k_2}$ pre-computed $n \times k$ eigenbases that are truncated to the first $k_1$ and $k_2$ columns respectively (\ie $\bm{\Phi}^\mathcal{Y}_{n \times k_j} = \{ \bm{\phi}^\mathcal{Y}_{i} \mid i=1,\dots,k_j \}, j=1,2$). For full-to-full matching, the value of $k_1$ and $k_2$ are then increased by one (or more) before recomputing the functional map,
    \begin{equation}\label{eq:zoomout-2}
    \mathbf{C}_{k_1 \times k_2} = \left (\bm{\Phi}^\mathcal{Y}_{n \times k_1} \right )^{+}\mathbf{P}~\bm{\Phi}^\mathcal{X}_{n \times k_2}.
    \end{equation}
    where $(\cdot)^+$ denotes the Moore-Penrose inverse. Solving for $\mathbf{P}$ and~\eqref{eq:zoomout-2} are alternated between until $k$ is reached by either $k_1$ or $k_2$.
    
    To handle partial matching, \cite{melzi2019zoomout} propose to compute a rectangular functional map $\mathbf{C}_{k_1 \times k_2}$. An estimate of rank $r$, based on~\cite{rodola2017partial}, is used to determine the shape of $\mathbf{C}$ for each iteration,
    \[
    k_1 = k_1 + 1,
    \quad 
    k_2 = k_2 + \left \lceil \frac{k_2}{k}(k-r) \right \rceil.
    \]
    This simple approach achieves excellent results for full-to-full matching. However, the method does not fare so well with partial matching. This is because the method still relies on stable eigenvectors, especially in the initial iterations where only low-frequencies are mapped. As illustrated in Fig.~\ref{fig:partial-eigs-example}, even the order of low-frequency eigenvectors that have been independently decomposed can prove to be highly inconsistent for partial matching.
    
    \cite{rampini2019correspondence} demonstrate that replacing the typical Laplacian operator with the Hamiltonian operator facilitates the computation of consistent eigenvectors between partial shapes. For the Hamiltonian operator, only the left-hand side of~\eqref{eq:eig-decomp} is modified to solve the following generalized eigenvalue problem for the cotangent Laplacian,
    \[
    (\mathbf{M} + \mathbf{D} \diagmat(\mathbf{v}))\bm{\Psi} = \mathbf{D}\bm{\Psi}\bm{\lambda}.
    \]
    $\mathbf{v} : \mathcal{X} \mapsto \left \{ 0,\tau \right \}^n$ is a potential function that indicates whether a point $p \in \mathcal{X}$ is inside ($\mathbf{v}(p)=0$) or outside ($\mathbf{v}(p)=\tau$) of the overlapping region $\mathcal{X}' = \mathcal{X} \cap \mathcal{Y}$ in $\mathcal{X}$. The effect of adding $\mathbf{D}\diagmat(\mathbf{v})$ is that rows in $\bm{\Psi}$ that correspond to a point assigned a value of $\tau$ (\ie $\mathbf{v}(p) = \tau$) become $\bm{\Psi}(p) \approx \mathbf{0}^\top$. \cite{rampini2019correspondence} set $\tau = 10 \Lambda_k$ experimentally, $\Lambda_k$ is the largest eigenvalue of $\bm{\Lambda}$.
    
\begin{figure}[tb]
        \centering
        \hfill\hfill\hfill\hfill
        \subfloat[$\mathbf{v}$]{\includegraphics[width=0.165\linewidth]{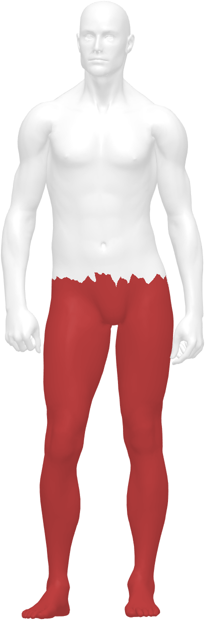}}\hfill
        \subfloat[$\bm{\phi}_7^{\mathcal{X}'}$]{\includegraphics[width=0.165\linewidth]{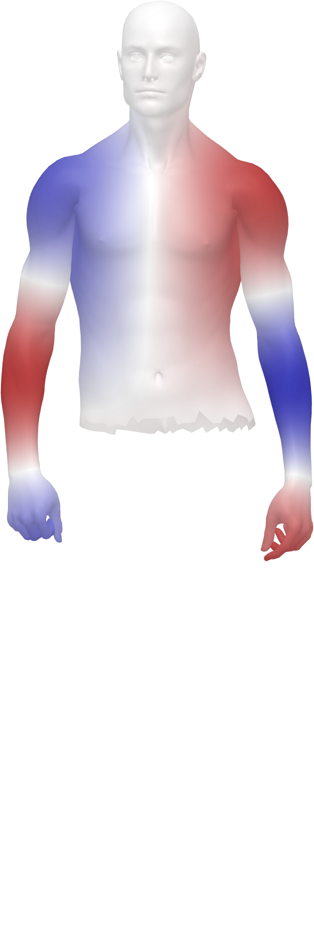}}\hfill
        \subfloat[$\bm{\psi}_7^\mathcal{X}$]{\includegraphics[width=0.165\linewidth]{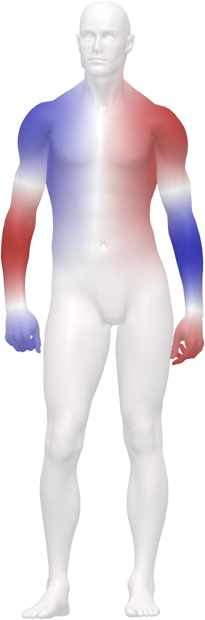}}
        \hfill\hfill\hfill\hfill
        \caption{Given two shapes where the overlapping region is known, the seventh eigenvector ($\bm{\phi}_7^{\mathcal{X}'}$ and $\bm{\psi}_7^{\mathcal{X}}$) computed using (b) the Laplacian operator and (c) the Hamiltonian operator $\tau\approx0.2313$.}
        \label{fig:hamiltonian-example}
    \end{figure}

    The advantage of this operator is that given a potential function $\mathbf{v}$ that characterizes $\mathcal{X}' \subseteq \mathcal{X}$, the eigenvectors on $\mathcal{X}$ are localized to $\mathcal{X}'$. This corresponds with the Laplcian operator---up to the sign ($\bm{\phi}_i^{\mathcal{X}'}=\pm\bm{\psi}_i^\mathcal{X}$)---when computed on a disconnected component $\mathcal{X}'$ (see Fig.~\ref{fig:hamiltonian-example}).
    
    To determine $\mathcal{X}'$, simply $\mathbf{v}$ must be found. An optimization function is formulated to align the Hamiltonian eigenvalues $\bm{\lambda}_\mathcal{X}$ with the Laplacian eigenvalues $\bm{\Lambda}_\mathcal{Y}$,
    \begin{align}\label{eq:hamitonian-opt}
    \min_{\mathbf{v} \in \mathbb{R}^n} \left \| \bm{\lambda}_\mathcal{X} ( \mathbf{D}^{-1}_\mathcal{X}\mathbf{M}_\mathcal{X} + \diagmat(\mathbf{v})) - \bm{\Lambda}_\mathcal{Y} \right \|^2_w.
    \end{align}
    To prevent large eigenvalues disproportionately influencing the result, a weighted $\ell_2$-norm is used,
    \[
    \left \| \bm{\lambda} - \bm{\Lambda} \right \|^2_w = \sum_{i=1}^k \frac{1}{\Lambda_i^2} (\lambda_i - \Lambda_i)^2.
    \]
    Once~\eqref{eq:hamitonian-opt} is solved, the eigenvectors should be sufficiently aligned to compute $\mathbf{C}$, as per~\eqref{eq:fmap}. The described approach is suitable for isometric partial-to-full matching, additionally the method is capable of handling multiple disconnected non-rigid components. Unfortunately, the non-convex nature of~\eqref{eq:hamitonian-opt} makes this implementation moderately slow. The method also fails when the number of eigenvectors $k$ is too low.

	\subsubsection{Spectral/Functional shape recovery methods}
    Given a mapping, by examining the intrinsic differences between two shapes it is possible to reconstruct a deformed source surface in the pose of the target shape. This section focuses on the problem of reconstructing an extrinsic representation from spectral representations and its application to non-rigid registration.
    
    \paragraph*{Shape difference operator reconstruction}
    Given a functional mapping between a source shape and a target shape, \cite{rustamov2013map} search for a mesh in a shape collection that has the minimal conformal- and equiareal-based shape difference. The method is limited by the collection available, and the complexity of the optimization increases when searching in large shape collections. \cite{boscaini2015shape} consider the problem of finding an embedding of the computed metric that minimizes an energy based on the shape difference operator. The method alternates between minimizing an energy term (based on~\cite{rustamov2013map}) to the Riemannian metric, and generating the extrinsic shape based on the discrete metric. The intrinsic metric does not suitably capture extrinsic curvature, causing ambiguity with possibly multiple isometric solutions. \cite{corman2017functional} address this problem by incorporating extrinsic information into an intrinsic operator. The edge lengths of an offset surface from a dual mesh are used to guide the reconstruction to a better solution. The authors demonstrate how the accuracy of the reconstruction deteriorates as the eigenbasis is truncated. \cite{huang2019operatornet} propose a deep learning framework to recover a shape using shape difference operators that also possess both extrinsic and intrinsic information. A set of training shapes is used to help constrain the solution space. The advantage of learning from shape operators is that the matrices may be treated as a grid structure---a convenient and well-studied domain for deep learning architectures. The method demonstrates superior results compared to~\cite{boscaini2015shape,corman2017functional}.
	
	\paragraph*{Template-less reconstruction}
	Unlike the previously discussed shape recovery works, which rely on shape difference operators, \cite{cosmo2019isospectralization} do not require a mapping to recover a shape as the method deforms a source mesh such that the Laplacian spectra better align. Similarly, \cite{marin2020instant,marin2021spectral} propose a learning-based approach to recover a shape from a given mesh Laplacian. By training a neural network to constrain the solution space to seen examples, the approach efficiently aligns the given spectra. To facilitate partial correspondence within the same class of shapes, \cite{moschella2021spectral} develop a shape union operator, which they incorporate into a reconstruction pipeline~\cite{marin2020instant}. With this method, reconstructed shapes are in a canonical pose.
	
	\paragraph*{Spectral operators}
	The problem of reconstructing a shape based on its spectral representation is not a new problem~\cite{gotsman2000spectral}; however, its use for registration has been limited. Given two shapes with an unknown correspondence, \cite{eisenberger2020smooth} align intrinsic and extrinsic shape representations to recover a deformed mesh in a coarse-to-fine manner. A novel operator is proposed that helps constrain the smoothness of the recovered mesh when increasing the number of dimensions of the embedding. In order to align the source and target shapes, the spectral representation is combined with Cartesian coordinates and surface normals. The correspondence and alignment are optimized in alternating steps.

	\subsection{Learning-Based Methods}
	
	Over the past decade, an increasing amount of effort has been placed into developing learning-based techniques that improve or enhance existing intrinsic pipelines. Many techniques seek to develop robust shape descriptors. As input to artificial neural networks, mesh geometry has been adapted to a variety of domains ($n$-dimensional descriptors~\cite{litany2017deep}, images~\cite{wang2018learning}, voxels~\cite{zeng2017match}, spectra~\cite{litman2014learning}, geodesics~\cite{masci2015geodesic}, and others~\cite{cao2020comprehensive,bronstein2021geometric}). Intrinsic information is also incorporated into the loss function (or equivalent energy) of these techniques (\eg labels~\cite{rodola2014kids,monti2017geometric}, geodesic distance~\cite{litman2014learning}, functional map quality~\cite{litany2017deep,roufosse2019unsupervised}).
	
	\paragraph*{Learning-based shape descriptors} A recent survey of related literature on data-driven descriptors was undertaken recently~\cite{rostami2019survey}, so the discussion here is focused on intrinsic methods specifically. A key series of works investigated learning-based methods to develop effective spectral descriptors~\cite{litman2014learning,windheuser2014optimal,boscaini2015learning,boscaini2016anisotropic}. \cite{litman2014learning} learn low-dimensional class-specific descriptors from the eigenvectors of the Laplacian. The authors observe that the \emph{heat kernel signature} (HKS)~\cite{sun2009concise} and \emph{wave kernel signature} (WKS)~\cite{aubry2011wave} may be considered as low-pass and band-pass filters in the spectral domain. A descriptor model that generalizes HKS and WKS is formulated. The training phase attempts to compute coefficients of a cubic spline that optimize the model to produce similar descriptors between corresponding points using a linear discriminant analysis-based approach. For the task of partial matching, the optimized descriptor is found to out-perform HKS \& WKS. \cite{windheuser2014optimal} propose a similar approach to metric learning to construct an infinite-dimensional descriptor. \cite{boscaini2015learning} exploit a localized spectral CNN. \cite{boscaini2016anisotropic} propose the use of multiple anisotropic kernels to form a directionally-sensitive descriptor. \cite{boscaini2016learning} incorporate the anisotropic technique as a weighting scheme into a CNN framework. \cite{masci2015geodesic} propose a CNN that uses a geodesic-based local operator based on~\cite{kokkinos2012intrinsic}. \cite{cosmo2016matching} learn an embedding function to construct descriptors that are robust to cluttered scenes. \cite{monti2017geometric} present a CNN framework that utilizes an intrinsic local operator that is shown to generalize operators from previous works~\cite{masci2015geodesic,boscaini2016learning} as mixtures of Gaussian kernels.
	
	A deficiency of previous chart-based methods is that they require a canonical orientation---well-structured data like digital images naturally have this property, but this is ill-defined on meshes. Previous methods use the maximal response over a set of rotations~\cite{masci2015geodesic} and the principal curvature directions~\cite{boscaini2016learning}, which can become inconsistent. \cite{poulenard2018multidirectional,wiersma2020cnns} propose to address this problem by establishing a directional function, based on parallel transport, over the surface. \cite{sun2020zernet} replace discrete bins used in~\cite{poulenard2018multidirectional} with a 2D rotation through the use of Zernike radial polynomials.
	
\begin{table*}[!t]
        \centering
        \caption{A summary of optimization-based intrinsic methods.}
        \label{tab:intrinsic-optimization-methods}
        \setlength{\tabcolsep}{3pt}
        \tablefont{
\begin{tabular}{c|c|cc|cccccc|cc|c}
\Xhline{1pt}
\multirow{3}{*}{~} & \multirow{3}{*}{\tabincell{c}{Sampling strategy}} & \multicolumn{2}{c|}{Data term} & \multicolumn{6}{c|}{Regularization constraints} & \multicolumn{3}{c}{Output} \\ \cline{3-5} \cline{6-8} \cline{8-13}
% 15
& & \multirow{2}{*}{\tabincell{c}{Point-to-point}} & \multirow{2}{*}{Spectral} & \multirow{2}{*}{Smooth} & \multirow{2}{*}{\tabincell{c}{Point\headingbreak{}constraints}} & \multirow{2}{*}{\tabincell{c}{Near-\headingbreak{}isometry}} & \multirow{2}{*}{\tabincell{c}{Non-\headingbreak{}isometry}} & \multirow{2}{*}{\tabincell{c}{Bijective or\headingbreak{}part-to-full injective}} & \multirow{2}{*}{Partial} & \multicolumn{2}{c|}{Precision} & \multirow{2}{*}{Transform} \\ \cline{11-12}
&&&&&&&&&& Point-to-point & Fuzzy & \\ \hline
\cite{Anguelov2004} & \checkmark & \checkmark &&&& \checkmark &&&&&& \checkmark \\
\cite{bronstein2006generalized} & \checkmark & \checkmark &&&&  \checkmark && \checkmark & \checkmark & \checkmark && \\
\cite{huang2008nonrigid} & \checkmark & \checkmark &&&& \checkmark &&&& \checkmark && \checkmark \\
\cite{devir2009reconstruction} && \checkmark &&&& \checkmark &&&& \checkmark && \checkmark \\
\cite{lipman2009mobius} && \checkmark && \checkmark & \checkmark &&&&& \checkmark && \\
\cite{ovsjanikov2010one} & \checkmark & \checkmark &&&& \checkmark &&&& \checkmark && \\
\cite{kim2011blended} && \checkmark && \checkmark &&& \checkmark &&& \checkmark && \\
\cite{sahillioglu2011coarse} & \checkmark &&&&& \checkmark &&&& \checkmark && \\
\cite{tevs2011intrinsic} & \checkmark & \checkmark &&&& \checkmark &&&& \checkmark && \\
\cite{ovsjanikov2012functional} && \checkmark & \checkmark & \checkmark & \checkmark & \checkmark && \checkmark && \checkmark && \\
\cite{rodola2012game} && \checkmark &&&& \checkmark && \checkmark && \checkmark && \\
\cite{sahillioglu2012minimum} & \checkmark &&&&& \checkmark &&&& \checkmark && \\
\cite{sahillioglu2012scale} & \checkmark & \checkmark &&&& \checkmark &&&& \checkmark && \\
\cite{solomon2012soft} & \checkmark &&&&&& \checkmark & \checkmark &&& \checkmark & \\
\cite{kovnatsky2013coupled} && \checkmark & \checkmark & \checkmark &&& \checkmark & \checkmark & \checkmark & \checkmark && \\
\cite{rustamov2013map} &&& \checkmark & \checkmark && \checkmark &&&&& \checkmark & \checkmark \\
\cite{vankaick2013bilateral} & \checkmark & \checkmark &&&&&&&& \checkmark && \\
\cite{tam2014diffusion} && \checkmark &&&& \checkmark &&&& \checkmark && \\
\cite{tam2014efficient} && \checkmark &&&& \checkmark &&&& \checkmark && \\
\cite{rodola2015pointwise} &&& \checkmark\textsuperscript{a} & \checkmark &&& \checkmark &&& \checkmark && \\
\cite{solomon2015convolutional} &&&&&&& \checkmark & \checkmark &&& \checkmark & \\
\cite{aigerman2016hyperbolic} && \checkmark && \checkmark & \checkmark &&&&& \checkmark && \\
\cite{litany2016nonrigid} && \checkmark & \checkmark & \checkmark && \checkmark && \checkmark & \checkmark & \checkmark && \\
\cite{solomon2016entropic} &&&& \checkmark &&&& \checkmark &&& \checkmark & \\
\cite{litany2017fully} && \checkmark & \checkmark & \checkmark && \checkmark && \checkmark & \checkmark & \checkmark && \\
\cite{mandad2017varianceminimizing} & \checkmark &&&&&& \checkmark & \checkmark && \checkmark && \\
\cite{rodola2017partial} && \checkmark & \checkmark & \checkmark & \checkmark & \checkmark && \checkmark & \checkmark & \checkmark && \\
\cite{rodola2017regularized} &&& \checkmark\textsuperscript{a} & \checkmark &&& \checkmark && \checkmark & \checkmark && \\
\cite{vestner2017efficient} & \checkmark & \checkmark & \checkmark &&&& \checkmark &&& \checkmark && \\
\cite{baden2018mobius} & \checkmark & \checkmark && \checkmark\textsuperscript{b} &&&&&& \checkmark && \\
\cite{sahillioglu2018genetic} & \checkmark & \checkmark &&&& \checkmark &&&& \checkmark && \\
\cite{cosmo2019isospectralization} & \checkmark && \checkmark\textsuperscript{a} & \checkmark &&& \checkmark &&&&& \checkmark \\
\cite{dyke2019non} && \checkmark &&&&& \checkmark &&& \checkmark && \checkmark \\
\cite{liu2019spectral} & \checkmark & \checkmark &&&&& \checkmark &&& \checkmark && \\
\cite{melzi2019zoomout} &&& \checkmark\textsuperscript{a} & \checkmark &&& \checkmark && \checkmark & \checkmark && \\
\cite{rampini2019correspondence} && \checkmark & \checkmark & \checkmark &&& \checkmark && \checkmark & \checkmark\textsuperscript{c} && \checkmark \\
\cite{zhao2019automatic} && \checkmark\textsuperscript{d} && \checkmark\textsuperscript{b} &&&&&& \checkmark && \\
\cite{eisenberger2020smooth} & \checkmark & \checkmark & \checkmark & \checkmark &&&&&& \checkmark && \checkmark \\
\cite{pai2021fast} &&& \checkmark & \checkmark &&& \checkmark & \checkmark && \checkmark && \\
\Xhline{1pt}
\multicolumn{13}{l}{\makecell[l]{\textsuperscript{a}Refines an initial functional map (\eg~\cite{nogneng2017informative}); \textsuperscript{b}Near-conformal; \textsuperscript{c}Part-to-part correspondence; \textsuperscript{d}Boundary-to-boundary.}}
        \end{tabular}
}
\end{table*}

\begin{table*}[!t]
	\caption{A summary of learning-based intrinsic methods.}
    \label{tab:intrinsic-learning-methods}
	\setlength{\tabcolsep}{2pt}
	\centering
	\tablefont{
		\begin{tabular}{ c | c c | c c c | c c c c c c }
			\Xhline{1pt}
			\multicolumn{1}{ c | }{\multirow{2}{*}{~}}
            &\multicolumn{2}{ c |}{Objective}
            &\multicolumn{3}{ c |}{Training type}
            &\multicolumn{5}{ c }{Loss} \\\cline{2-11}
			\multicolumn{1}{ c| }{}
			&\multicolumn{1}{c }{\tabincell{c}{Learnable \headingbreak{}correspondences}}
            &\multicolumn{1}{ c| }{\tabincell{c}{Learnable \headingbreak{}deformation}}
            &\multicolumn{1}{ c }{Supervised}
            &\multicolumn{1}{ c }{\tabincell{c}{Semi-\headingbreak{}supervised}}
            &\multicolumn{1}{ c| }{Unsupervised}
            &\multicolumn{1}{ c }{\tabincell{c}{Ground-truth\headingbreak{} correspondences}}
            &\multicolumn{1}{ c }{\tabincell{c}{Ground-truth\headingbreak{} spectra}}
            &\multicolumn{1}{ c }{Alignment}
            &\multicolumn{1}{ c }{\tabincell{c}{Functional \headingbreak{}map heuristics}}
            &\multicolumn{1}{ c }{\tabincell{c}{Non-\headingbreak{}isometry}}\\ \hline
\cite{litman2014learning} & \checkmark & & \checkmark & & & \checkmark & & & & \\
\cite{rodola2014kids} & \checkmark & & \checkmark & & & \checkmark & & & & \checkmark \\
\cite{masci2015geodesic,boscaini2016learning,monti2017geometric} & \checkmark & & \checkmark & & & \checkmark & & & & \\ % also uses a descriptor loss
\cite{litany2017deep} & \checkmark & & \checkmark & & & \checkmark & & & & \\
\cite{wang2018learning} & \checkmark & & \checkmark & & & \checkmark & & & & \\
\cite{halimi2019unsupervised} & \checkmark & & & & \checkmark & & & & \checkmark & \\
\cite{roufosse2019unsupervised} & \checkmark & & & & \checkmark & & & & \checkmark & \\
\cite{donati2020deep} & \checkmark & & \checkmark & & & & \checkmark & & & \\
\cite{marin2020instant,marin2021spectral} & & \checkmark & & \checkmark & & & \checkmark & \checkmark & & \\
\cite{wang2020mgcn} & \checkmark & & \checkmark & & & \checkmark & & & & \\ % also uses a descriptor loss
\cite{moschella2021spectral} & & \checkmark & & \checkmark & & & \checkmark & \checkmark & & \\
\Xhline{1pt}
\end{tabular}
    }
\end{table*}	
	
	\paragraph*{Learning-based functional maps}
	A collection of works have investigated task-driven approaches to learning enhanced descriptors for functional maps~\cite{corman2015supervised,litany2017deep,halimi2019unsupervised,roufosse2019unsupervised,donati2020deep}. The earliest method~\cite{corman2015supervised} proposes a shallow learning framework in which spectral descriptors are re-weighted. These re-weighted descriptors are then used to initialize a functional map~\cite{ovsjanikov2012functional}, where the difference to a known ground-truth map for training is minimized. \cite{litany2017deep} consider a similar approach in a deep learning framework. During the training phase, weights are learned to enhance feature descriptors. The loss function then measures the quality of the subsequent spatial map with geodesic distances to a ground-truth map. The method requires training data with pointwise correspondence, which are often expensive and difficult to acquire. Follow-up works have sought to facilitate unsupervised learning through penalizing geodesic distortion~\cite{halimi2019unsupervised}, and functional map heuristics (\eg orthogonality, bijectivity, and commutativity)~\cite{roufosse2019unsupervised}. \cite{donati2020deep} propose a similar framework in which the descriptors are learned from point clouds. While some of these methods do demonstrate promising evaluative results, they rely on simple functional mapping techniques that are known to not be suitable for partial correspondence leading to sub-optimal performance in this scenario. Recently, to address the problem of partial matching, \cite{attaiki2021dpfm} propose to incorporate a cross-attention block between sets of learned features for two input shapes. This block enables the communication of features between shapes and causes non-overlapping features to be down-weighted, making it possible to efficiently predict the overlapping region.
	
    \paragraph*{Learning-based reconstruction}
    \cite{eisenberger2020deep} incorporate the scheme from~\cite{eisenberger2020smooth} to produce a spectral-based learning framework, which---with sufficiently high-dimensional eigenvectors---can approximately reconstruct a target surface. \cite{cosmo2020limp} incorporate geodesic regularization into an auto-encoder framework, leading to low-distortion reconstructions. \cite{eisenberger2021neuromorph} adopts a deep neural network architecture to match intrinsic feature descriptors and interpolate extrinsic vertex locations simultaneously unsupervised. The method is not fully intrinsic as it also relies on an as-rigid-as-possible error term. Given a shape's spatial coordinates as input, \cite{bouritsas2019neural,gong2019spiralnet} implement auto-encoder architectures that at their core comprise of intrinsic convolutional operators. The operator is composed of neighboring vertices that are traversed in a spiral pattern, which is inherently sensitive to variations in connectivity.
	
	\subsection{Summary}
	The key optimization-based and learning-based works discussed in this section are summarized in Tables~\ref{tab:intrinsic-optimization-methods} and \ref{tab:intrinsic-learning-methods}.

		\begin{table*}[!t]
    \centering
    \caption{Datasets for benchmarking correspondence and registration methods.}
    \label{tab:dataset-taxonomy}
    \setlength{\tabcolsep}{2.3pt}
    \tablefont{
    \begin{tabular}{c|ccc|c|ccc|c|ccc|c|c|c|c}
        \Xhline{1pt}
        \multirow{3}{*}{~} & \multicolumn{3}{c|}{Deformation type} & \multirow{3}{*}{\tabincell{c}{Partial\headingbreak{}correspondence}} & \multicolumn{3}{c|}{Scan type} & \multicolumn{5}{c|}{Ground-truths} &  \multirow{3}{*}{Anthropometric} & \multirow{3}{*}{\tabincell{c}{No.~of \headingbreak{} scans}} & \multirow{3}{*}{$>$ 25K vertices}\\ \cline{2-4} \cline{6-10} \cline{11-13}
        & \multirow{2}{*}{\tabincell{c}{Near-\headingbreak{}isometry}} & \multirow{2}{*}{\tabincell{c}{Non-\headingbreak{}isometry}} & \multirow{2}{*}{\tabincell{c}{Topological\headingbreak{}change}} && \multirow{2}{*}{Real} & \multirow{2}{*}{Synthetic} & \multirow{2}{*}{\tabincell{c}{Deformed \headingbreak{} template}} & \multirow{2}{*}{Dense} & \multicolumn{3}{c|}{Acquisition} & \multirow{2}{*}{\tabincell{c}{Training \headingbreak{} facility}} & & & \\ \cline{10-12}
        &&&&&&&&& Manual & Automatic & Intrinsic &&&&\\ \hline
        \cite{robinette1999caesar} & & \checkmark & \checkmark & \checkmark & \checkmark & & & & \checkmark & & & & \checkmark & 4,431\textsuperscript{a} & \checkmark \\ %
        \cite{sumner2004deformation} & \checkmark & & & & & \checkmark & \checkmark & \checkmark & & & \checkmark & & partially & 551 & \\ %
        \cite{anguelov2005scape} & \checkmark & & & & \checkmark & & \checkmark & &&&\checkmark & & \checkmark & 71 & \checkmark \\ %
        \cite{adobe2008mixamo} & \checkmark & & & & & \checkmark & \checkmark & & \checkmark & & & & \checkmark & 121\textsuperscript{b} & \checkmark \\ %
        \cite{bronstein2008numerical} & \checkmark & & & & & \checkmark & \checkmark & \checkmark & & & \checkmark & & partially & 80 & \checkmark \\ %
        \cite{vlasic2008articulated} & \checkmark & & & & \checkmark & & \checkmark & \checkmark & & & \checkmark & & partially & 1,500 & \\ %
        \cite{hasler2009mpii} & & \checkmark & & & \checkmark & & \checkmark & \checkmark & \checkmark & \checkmark & & & \checkmark & 550 & \checkmark \\ %
        \cite{bronstein2010shrec} & \checkmark\textsuperscript{c} & & & \checkmark & & \checkmark & \checkmark & \checkmark & & & \checkmark & & & 56 & \checkmark \\ %
        \cite{kim2011blended} & & \checkmark & & & \checkmark & \checkmark & \checkmark &  & \checkmark & & & & partially & 551 & \\ %
        \cite{bogo2014faust} & & \checkmark & \checkmark & \checkmark & \checkmark & & \checkmark\textsuperscript{d} & \checkmark & & \checkmark & & \checkmark & \checkmark & 300 & \checkmark \\ %
        \cite{cao2014facewarehouse} & & \checkmark & & & \checkmark & & \checkmark & \checkmark & & \checkmark & & & \checkmark & 3,000 & \checkmark \\
        \cite{rodola2014kids} & \checkmark & & & & & \checkmark & \checkmark & \checkmark & & & \checkmark & & & 32 & \checkmark \\ %
        \cite{yang2014spring} & & \checkmark & & & \checkmark & & \checkmark & \checkmark & & \checkmark & & & \checkmark & 3,000 & \\%
        % topological kids
        \cite{lahner2016shreckids} & \checkmark & & \checkmark & \checkmark & & \checkmark & \checkmark & \checkmark & & & \checkmark & & \checkmark & 26 & \checkmark \\ %
        \cite{bogo2017dfaust} & & \checkmark & \checkmark & \checkmark & \checkmark & & \checkmark\textsuperscript{d} & \checkmark & & \checkmark & & \checkmark & \checkmark & 40,000 & \checkmark \\ %
        \cite{pishchulin2017building} & & \checkmark & & & \checkmark & & \checkmark & \checkmark & & \checkmark & & & \checkmark & 4,309 & \\ %
        \cite{rodola2017partial}
        & \checkmark & & & \checkmark & & \checkmark & \checkmark & \checkmark & & & \checkmark & & partially & 1,216 & \checkmark \\ %
        \cite{rodola2017range} & \checkmark & & & \checkmark & & \checkmark & \checkmark & \checkmark  & & & \checkmark & \checkmark & partially & 4,258 & \checkmark \\ %
        \cite{cheng2018fab} & & \checkmark & & & \checkmark & & \checkmark & \checkmark & & \checkmark & & & \checkmark & 1,800,000 & \checkmark\\
        \cite{saint2018bodytex} & & \checkmark & & & \checkmark & & \checkmark & \checkmark & & \checkmark & & \checkmark & \checkmark & 400 & \checkmark \\
        \cite{dyke2019shrec} & & \checkmark & \checkmark & & \checkmark & & & & \checkmark & & & & & 50 & \checkmark \\ %
        \cite{melzi2019matching} & & \checkmark & \checkmark & & \checkmark & \checkmark & \checkmark & \checkmark & & \checkmark & & & \checkmark & 44 & \checkmark \\ %
        \cite{dyke2020shrecb} & & \checkmark & \checkmark & \checkmark & \checkmark & \checkmark & & & \checkmark & & & & & 14 & \checkmark \\ %
        \cite{dyke2020shreca} & & \checkmark & \checkmark & & \checkmark & & &  & \checkmark & & & & & 12 & \checkmark \\ %
        \cite{chen2021capturing} & & \checkmark & & \checkmark & \checkmark & & & \checkmark & & \checkmark & & & \checkmark & 10,200 & \\
        \Xhline{1pt}
        \multicolumn{16}{l}{\tabincell{l}{\textsuperscript{a}Dataset contains some shapes that are isometric up to scaling; \textsuperscript{b}Only a subset of ground-truths are publicly available; \textsuperscript{c}110 of the scans do not have landmarks;
        \textsuperscript{d}Base models.}}
    \end{tabular}
    }
\end{table*}

	%-------------------------------------------------------------------------
	\section{Datasets and Benchmarks}
	\label{sec:benchmarks}
	There are many publicly available datasets that can be used for the effective evaluation of non-rigid registration methods. Such datasets may be used to measure the quality of a surface alignment quantitatively (\eg fitting error~\cite{cignoni1998metro,rusinkiewicz2001efficient}, geodesic error\cite{kim2011blended}, coverage error~\cite{dyke2020shrecb}, area ratio~\cite{zeng2010dense,zeng2016high}) and qualitatively (\eg model rendering, distortion~\cite{ovsjanikov2013analysis}, information transport~\cite{melzi2019visual}). Alternatively, applications that may rely on registration (\eg shape retrieval~\cite{funkhouser2005shape,escolano2010bypass}) can be used to indirectly measure the performance of registration methods. The reader is directed to~\cite{vankaick2011survey} for further discussion of other relevant evaluation methods.
	
	State-of-the-art datasets for benchmarking often include ground-truth correspondences, which can be used to measure the root-mean-square error and geodesic error~\cite{kim2011blended} between a predicted ground-truth location and the known ground-truth location on a target surface. This is an effective and commonly used technique to summarize and compare the accuracy of registration methods (\eg RMSE~\cite{shimada2019dispvoxnets,yao2020quasi,li2020robust}, geodesics~\cite{chen2015robust,corring2016resonant,maron2016point,groueix2018coded,eisenberger2020hamiltonian}). \cite{rostami2019survey} discuss benchmarks for the purposes of evaluating 3D learning-based shape descriptors, many of which can be---and are---used for evaluating registration techniques.
	
	A common trend has been the use of anthropomorphic shapes, such as human bodies and human faces. An extended overview of public 3D facial datasets is discussed by~\cite{zhou2018face}. Most contain sparse fiducial landmarks (\eg \cite{phillips2005overview,savran2008bosphorus,gupta2010texas}) that can be used for evaluating registration methods~\cite{gilani2015shape,santa2016face,fan2018dense}. If RGB texture data is acquired, it is possible to establish reliable landmarks automatically~\cite{cao2014facewarehouse}. Beyond facial datasets, texture data is commonly used for computing ground-truth correspondences reliably~\cite{bogo2014faust,bogo2017dfaust,dyke2019shrec,dyke2020shreca}.
	
	Over the past decade, many notable benchmarks have been developed for Shape Retrieval Contest~(SHREC) tracks~\cite{bronstein2010shrec,lahner2016shreckids,rodola2017range,dyke2019shrec,melzi2019matching,dyke2020shreca,dyke2020shrecb}. \cite{lahner2016shreckids,rodola2017range} introduce challenges such as topological changes and partial correspondence by applying synthetic modifications to models.
	
	% RMD could discuss sparse GTs just before this.
	Dense ground-truth correspondences for real meshes are often automatically acquired by non-rigidly aligning a template to a given scan using methods such as non-rigid registration~\cite{zuffi2017menagerie,melzi2019matching}, linear blend skinning~\cite{vlasic2008articulated,ponsmoll2015dyna,chen2021capturing} or parametric modeling~\cite{anguelov2005scape,yang2014spring,xu2018multilevel}. The advantage of aligning a template mesh to a set of target scans is that it is possible to compute dense correspondences; however, these correspondences may contain local alignment errors. The alignment can be improved by incorporating texture information~\cite{bogo2014faust,cao2014facewarehouse,bogo2017dfaust,cheng2018fab,saint2018bodytex} or a sparse set of reliable landmarks~\cite{hasler2009mpii}.
	
	A key deficiency in existing datasets is the lack of a training facility for learning-based methods. As the popularity of machine learning continues to expand, the necessity for benchmarks that enable fair and consistent comparisons between techniques has increased. Presently, only a few benchmarks address this problem~\cite{bogo2014faust,bogo2017dfaust,rodola2017range,saint2018bodytex}. For the benchmark dataset~\cite{dyke2019shrec}, which does not have a training facility, the learning-based method~\cite{groueix2018coded} is trained on~\cite{varol2017learning}. Also, \cite{marin2020instant,marin2021spectral} utilize a generative face model~\cite{ranjan2018generating} to train a learning-based method for evaluative purposes.
	
	Table~\ref{tab:dataset-taxonomy} provides an overview of datasets used for evaluative purposes in the literature.

	\section{Directions for Future Works}
	\label{sec:future}
	
	\paragraph*{Input Data Type}
	One important application for non-rigid registration is the acquisition of 3D shapes from scanned data. For existing works, the input data type is mainly 3D geometry such as point clouds or depth maps. Although such data can be captured with depth sensors, currently for the general public it is still easier to capture RGB videos with portable devices.
	For convenient 3D content acquisition, it is desirable to efficiently reconstruct 3D geometry from monocular RGB video on mobile devices.
	Recent developments in differentiable rendering~\cite{Kato2020DifferentiableRendering} and neural rendering~\cite{TewariFTSLSMSSN20} have enabled 3D reconstruction from monocular videos~\cite{YarivKMGABL20,niemeyer2020differentiable,yariv2021volume,zhang2021learning,Oechsle2021ICCV}, but existing methods are mainly designed for rigid shapes. Their high computational costs also prevent them from being run in real-time on mobile devices. 
	Efficient reconstruction of dynamic 3D shapes from monocular videos will be an interesting research avenue with a significant impact.

	\paragraph*{Geometric Shape Representation}
	The input 3D data of current registration methods is usually represented as discrete signals. Recently, neural implicit representations~\cite{Park2019DeepSDF,sitzmann2020implicit,chibane2020neural}, which are based on parameterizing a continuous differentiable signal with a neural network, have become a promising alternative to conventional 3D representations such as point clouds and meshes. They are not coupled to the spatial resolution, and have been successfully applied to various tasks like novel view synthesis~\cite{MildenhallSTBRN20} and multi-view 3D reconstruction~\cite{YarivKMGABL20}. It must be noted that neural implicit representations are different from the implicit surface representations discussed in Section~\ref{sec:representation}. The neural implicit function is defined over the whole continuous space and is represented in the form of function composition rather than the explicit grid and explicit function values.
	Adapting existing registration methods to this new representation, as well as designing new methods that can make use of its representational power, is a promising research direction.

	\paragraph*{Extrinsic methods}
	For registration problems that involve large deformation, optimization-based approaches often require proper initialization to achieve good results. Although they can be initialized using shape descriptors such as FPFH or SHOT, such handcrafted features are inevitably ambiguous when the local shapes are similar. Moreover, such descriptors tend to be sensitive to noise, and may lead to poor registration results on noisy models. Therefore, further research is needed for optimization-based approaches that can handle large deformation between noisy models. In recent years, learning-based approaches have shown promising results for such problems, either as a standalone registration technique or as a way to determine initial correspondence for an optimization method. However, the performance of machine learning models is highly dependent on the quality of the data. Currently, public datasets with accurate correspondence labels are still limited in terms of both their quantities and the types of shapes they cover. As a result, the generalizability of existing machine learning models is still not satisfactory. It will be highly beneficial to construct more labeled datasets with better coverage of shapes, and/or to develop machine learning approaches with better generalization performance.   
	
	The use of adaptive weights and robust norms have helped optimization-based methods align shapes that overlap partially. However, their performance is still not satisfactory when the overlap area is small, especially when the alignment involves
	large deformation and/or the data is noisy. A key issue is that due to the non-rigid nature of the problem, an extrinsic formulation may not effectively distinguish whether a source point has a corresponding point on the target surface. 
	This is a challenging case that can occur in practical applications (\eg dynamic reconstruction from a small number of scans), and will require further investigation.
	
	Existing shape regularizations in optimization-based methods assume the deformation to be either near-isometric, non-isometric but with certain characteristics such as conformality, or a combination of them. While such priors can handle deformations with a certain degree of regularity, they are less effective for some complex physically-based deformations such as the ones shown in~\cite{dyke2020shreca}. New strategies for handling such complex deformations, such as data-driven approaches to properly model the deformation behavior, will be an interesting research direction.

	\paragraph*{Intrinsic methods}
	A substantial amount of work has focused on the development of correspondence techniques between humans. Consequentially, we observe the strong performance in anthropometric applications.
	
	Many registration methods are sensitive to their initialization, or depend upon invariant point-wise features. While intrinsic methods can also be susceptible to these problems, the use of pairwise measures can alleviate these in some scenarios. Unfortunately, pairwise measures introduce other challenges: for example, geodesic distances are only invariant under isometric deformation. Furthermore, the convenient assumptions of many intrinsic methods can quickly deteriorate when confronted with partial geometry, where certain bijectivity conditions can no longer be satisfied.
	
	For real-time applications, the performance of hand-crafted intrinsic partial matching techniques is inadequate for larger models. Learning-based methods have the potential to achieve superior performance, as they may avoid expensive optimization problems.
	
	Accurately estimating the overlap between the source and target surfaces is instrumental to the accuracy of these methods. The development of overlap estimation techniques (\eg \cite{attaiki2021dpfm}) is a promising avenue of research that may lead to intrinsic methods that handle both partial and non-isometric shapes.
	
	For 3D reconstruction using devices such as hand-held scanners, where the subject is captured in small patches, the inconsistency of standard Laplacian eigenfunctions means that current functional mapping methods are not suitable.

	Existing literature on intrinsic shape correspondence tends to focus on a subset of related challenges. The past ten years has seen a focus on relevant problems such as non-isometry of smooth deformations, mesh connectivity and part-to-full correspondence. Topological changes caused by fused geometry during scanning continue to be a challenge for intrinsic optimization methods, while data-driven approaches may help with such scenarios. Furthermore, geometry with extensive and numerous surface discontinuities are understudied (such as that of a tea towel that has been wound up). This type of problem challenges the key underlying assumption about the local smoothness of a mapping. (A variety of challenging deformations are considered in~\cite{schmidt2019visual,dyke2020shreca}.)

	\paragraph*{Acknowledgements}
	This work was partially supported by the Swiss National Science Foundation (SNSF) under project number 200021-188577, the Guangdong International Science and Technology Cooperation Project (2021A0505030009), the National Natural Science Foundation of China (No. 62122071), the Youth Innovation Promotion Association CAS (No. 2018495), and ``the Fundamental Research Funds for the Central Universities'' (No. WK3470000021).

	% bibtex
	\bibliographystyle{eg-alpha-doi}
	\bibliography{NonRigidRegistration}   
	
\end{document}